 \documentclass[final,3p,times,twocolumn]{elsarticle}

\usepackage{times}  
\usepackage{helvet}  
\usepackage{courier}  
\usepackage[hyphens]{url}  
\usepackage{graphicx} 
\usepackage{natbib}  
\usepackage{caption} 
\usepackage{algorithm}
\usepackage{algorithmic}
\usepackage{amsmath} 
\usepackage{booktabs} 
\usepackage{multirow} 
\usepackage{amsthm}
\theoremstyle{plain} 
\newtheorem{proposition}{Proposition}
\usepackage{threeparttable}
\usepackage{siunitx}
\usepackage[table]{xcolor}
\usepackage{tabularx}

\usepackage{subcaption}
\usepackage{multirow} 
\usepackage{xcolor}
\usepackage{listings}
\usepackage{booktabs}
\definecolor{codegreen}{rgb}{0,0.6,0}
\definecolor{codegray}{rgb}{0.5,0.5,0.5}
\definecolor{codepurple}{rgb}{0.58,0,0.82}
\definecolor{backcolour}{rgb}{0.95,0.95,0.92}

\lstdefinestyle{mystyle}{
    backgroundcolor=\color{backcolour},
    commentstyle=\color{codegreen},
    keywordstyle=\color{magenta},
    numberstyle=\tiny\color{codegray},
    stringstyle=\color{codepurple},
    breakatwhitespace=false,
    breaklines=true,
    captionpos=b,
    keepspaces=true,
    numbers=none,
    numbersep=5pt,
    showspaces=false,
    showstringspaces=false,
    showtabs=false,
    tabsize=2,
    basicstyle=\small\ttfamily,
    frame=single,
    moredelim=[is][\bfseries]{[b]}{[/b]},
}

\usepackage{chngcntr} 
\usepackage{apptools} 

\AtAppendix{
    \counterwithin{equation}{section}   
    \counterwithin{table}{section}      
    \counterwithin{figure}{section}     
}

\usepackage{newfloat}
\usepackage{listings}
\DeclareCaptionStyle{ruled}{labelfont=normalfont,labelsep=colon,strut=off} 
\floatstyle{ruled}
\newfloat{listing}{tb}{lst}{}
\floatname{listing}{Listing}

\usepackage{amssymb}
\usepackage{amsmath}

\journal{Journal of Manufacturing Systems}

\begin{document}

\begin{frontmatter}



\title{ReflecSched: Solving Dynamic Flexible Job-Shop Scheduling via LLM-Powered Hierarchical Reflection}




\author[buaa]{Shijie Cao}
\ead{cls1277@buaa.edu.cn}

\author[buaa,buaaqd,buaahz,zgclab]{Yuan Yuan\corref{cor}}
\ead{yuan21@buaa.edu.cn}

\cortext[cor]{Corresponding author}

\address[buaa]{School of Computer Science and Engineering,\\
	Beihang University, Beijing 100191, China}

\address[buaaqd]{Qingdao Research Institute}

\address[buaahz]{Hangzhou Innovation Institute}

\address[zgclab]{Zhongguancun Laboratory} 

\begin{abstract}
The NP-hard Dynamic Flexible Job-Shop Scheduling (DFJSP) problem involves real-time events and complex routing. While traditional rules are efficient but rigid, deep learning is opaque and requires feature engineering. Large Language Models (LLMs) promise adaptive reasoning without this engineering overhead, yet we find their direct application is suboptimal. Baseline LLMs suffer from three key pitfalls: the long-context paradox, where crucial data is underutilized; an underutilization of expert heuristics; and myopic decision-making. To address this, we propose ReflecSched, a framework that empowers the LLM beyond a direct scheduler by equipping it with a strategic analysis capability. ReflecSched tasks the LLM to analyze heuristic-driven simulations across multiple planning horizons and distill them into a concise, natural-language summary termed ``Strategic Experience''. This summary is then integrated into the prompt of a final decision-making module, guiding it to produce non-myopic actions. Experiments demonstrate ReflecSched achieves superior performance, with its best variants attaining an average RPD of \textcolor{blue}{6.09\%} and rank of \textcolor{blue}{4.39 on GEN-Bench}, significantly outperforming strong traditional and learning-based methods \textcolor{blue}{including HMPSAC and IDDQN}. \textcolor{blue}{Extensive cross-benchmark evaluations on MK-Bench and the semiconductor-focused JMS-Bench further confirm the framework's robustness, yielding minimum average RPDs of 6.83\% and 6.18\%, respectively.} It also statistically and decisively surpasses direct LLM baselines, securing a 71.35\% Win Rate while being, on average, 15.1\% more token-efficient on Normal-scale problems. \textcolor{blue}{Furthermore, cumulative runtime analysis reveals that ReflecSched's zero-shot nature eliminates the training bottleneck, providing a decisive efficiency advantage in high-variability manufacturing environments.} Ablation studies attribute this performance to a robust reflection mechanism that leverages high-quality, contrastive experience. This mechanism mitigates key LLM pitfalls like myopic greed, enabling ReflecSched to outperform all evaluated heuristics. Ultimately, the framework's performance is statistically on par with an oracle-like strategy, showcasing its effectiveness and robustness.
\end{abstract}







\end{frontmatter}



\section{Introduction}
\label{sec:introduction}

Dynamic Flexible Job-Shop Scheduling (DFJSP) is a long-standing challenge in operations research and a key component of modern smart manufacturing~\cite{li2025learning}.
As an NP-hard problem, it involves the complex task of continuously allocating operations to machines in a real-time environment characterized by stochastic events such as new job arrivals or machine breakdowns.
The ability to efficiently schedule these operations critically impacts the agility, resilience, and profitability of production systems.
For decades, the field has been dominated by methods ranging from classical heuristics to meta-heuristics~\cite{cao2023inverse}.
While foundational, these approaches often rely on handcrafted rules that can struggle to generalize across diverse and unforeseen dynamic scenarios~\cite{ferreira2022effective}.

In recent years, deep learning (DL), particularly reinforcement learning (RL), has emerged as a promising paradigm for this problem~\cite{xu2025learn}.
By training agents on simulated data, DL-based methods can learn sophisticated, state-aware scheduling policies.
However, this capability comes at a significant cost.
A key limitation lies in the intricate process of state and action space engineering: translating the rich state of a dynamic workshop into a fixed-size numerical vector that a neural network can process is a complex task requiring dual expertise in scheduling and machine learning~\cite{zhang2020learning}.
Furthermore, the resulting models are often ``black boxes'' with opaque decision-making logic that is difficult for engineers to interpret or trust in high-stakes environments~\cite{ngwu2025reinforcement}.

The advent of Large Language Models (LLMs) offers a different approach, potentially circumventing these challenges by leveraging natural language as a more accessible interface.
An LLM-based scheduler could, in principle, reason directly over a textual description of the factory state, reducing the need for complex numerical encoding~\cite{abgaryan2024llms}.
This suggests the possibility of developing and adapting scheduling strategies with greater flexibility.

However, our work finds that a direct application of LLMs to DFJSP leads to consistently suboptimal performance.
We argue that the standard autoregressive generation process is not inherently well-suited for the strategic, lookahead-dependent reasoning that scheduling demands.
Through a motivational analysis, we identify and empirically validate three key pitfalls of this direct approach:

\begin{itemize}
\item \textbf{The Long-Context Paradox}: Foundational static information, such as machine processing times and job structures, is often underutilized by the model as the prompt length increases.
\item \textbf{Underutilization of Heuristics}: LLMs exhibit difficulty in reliably applying expert-provided procedural knowledge, such as Priority Dispatching Rules (PDRs), often allowing their generalized pre-trained behaviors to override explicit guidance.
\item \textbf{Myopic Greed}: The lack of a structured planning mechanism creates a strong tendency towards locally optimal but globally inefficient decisions, which can lead to downstream bottlenecks.
\end{itemize}

To address these shortcomings, we introduce ReflecSched, a framework that fundamentally restructures the LLM's role in the scheduling process.
Rather than acting as a purely reactive decision-maker, the LLM also serves as a strategic analyst within a planning loop that decouples long-horizon reasoning from immediate execution.
At its core, ReflecSched's Hierarchical Reflection module performs multi-level, heuristic-driven simulations of future trajectories.
It then tasks the LLM to reflect upon these simulations and distill them into a concise, actionable ``Strategic Experience.''
This experience, generated only at critical, event-driven moments, subsequently guides a lean, fast Experience-Guided Decision-Making module to produce a high-quality, non-myopic action.


Our contributions are threefold:
\begin{itemize}
    \item \textbf{A rigorous empirical diagnosis of LLM-based schedulers.} We are the first to systematically identify and validate three fundamental failure modes of LLMs in the complex DFJSP domain: the long-context paradox, heuristic underutilization, and myopic greed. This analysis provides crucial insights for the broader field of LLM-based planning.

    \item \textbf{A novel LLM-based scheduling paradigm, ReflecSched.} We propose a framework that recasts the LLM from a reactive decision-maker into a strategic analyst. By decoupling long-horizon planning from immediate execution via a hierarchical reflection mechanism, ReflecSched is designed to \textcolor{blue}{mitigate} the identified pitfalls.

    \item \textcolor{blue}{\textbf{Multi-tiered benchmark suites and open-source ecosystem.} We introduce a comprehensive benchmarking ecosystem comprising GEN-Bench, MK-Bench, and JMS-Bench. These suites span from general-purpose scenarios to standardized flexible job-shop instances and high-fidelity semiconductor manufacturing environments. By open-sourcing these benchmarks alongside the ReflecSched framework, we provide a rigorous, multi-dimensional evaluation protocol to facilitate future research in LLM-based industrial planning.}

\end{itemize}

\section{Related Work}
\label{sec:related_work}

\textbf{Heuristic and Metaheuristic Approaches:}
Heuristic-based methods are a foundational approach for the Dynamic Job-Shop Scheduling Problem (DJSP) due to their computational tractability.
Research has advanced from simple dispatching rules to complex metaheuristics designed to handle diverse operational constraints.
For instance, some works have applied \textcolor{blue}{greedy randomized adaptive search} algorithms to manage a wide array of dynamic events~\cite{baykasouglu2020greedy}, while others have used hybrid Particle Swarm Optimization for joint production and transportation scheduling~\cite{ren2022joint}.
Further research has focused on specific uncertainties, such as variable processing times, using techniques like Artificial Bee Colony algorithms~\cite{shahgholi2019heuristic}.
\textcolor{blue}{To align with the human-centric paradigm of Industry 5.0, recent research has formalized the HDDFJSP by integrating worker-related dynamic factors into a multi-objective optimization framework, leveraging a Q-learning-enhanced differential evolution algorithm to balance production efficiency with operator well-being~\cite{li2025q}.}
A key limitation of these metaheuristics, however, is their reliance on a pre-defined search logic that is inherently less adaptive and struggles to generalize to new, unseen problem scenarios~\cite{chen2025optimizing}.

\textbf{Deep Reinforcement Learning Approaches:}
To address the generalization limitations of heuristics, Deep Reinforcement Learning (DRL) has emerged as an alternative that learns adaptive policies, typically by formulating the DJSP as a Markov Decision Process.
Architectural innovations have been central to this area.
For example, Pointer Networks have been utilized for end-to-end policy learning that avoids reliance on predefined rules~\cite{yang2024deep}.
To enhance scalability, a multi-agent perspective has been adopted through decentralized frameworks~\cite{liu2023deep}.
Other approaches decompose the decision-making process using hierarchical structures~\cite{luo2021real}.
\textcolor{blue}{Recently, to enhance the adaptability of dynamic job-shop scheduling, Yu et al. introduced a DRL-based framework that reformulates scheduling states as multi-channel images, leveraging a Spatial Pyramid Pooling Fast module to achieve scale-independent feature extraction and a region-based dense reward function to facilitate fine-grained policy optimization~\cite{yu2026adeepreinforcement}.}
Nevertheless, DRL presents its own significant challenges: performance is highly sensitive to the complex engineering of state and reward functions; the resulting policies are often opaque ``black boxes'' hindering trust; and they demand accurate simulation environments for training, which are often unavailable for complex real-world systems.

\textbf{Large Language Models Paradigms:}
Seeking to address the challenges of both handcrafted logic and the black-box nature of DRL, LLMs have recently introduced a new paradigm for scheduling by leveraging their vast pre-existing knowledge and reasoning capabilities.
One direction explores LLMs as end-to-end solvers by fine-tuning them on large-scale datasets of problems paired with their solutions~\cite{abgaryan2025starjob, abgaryan2024llms}.
Another approach employs LLMs as automated problem modelers that translate natural language into executable code for traditional solvers~\cite{amarasinghe2023ai, ahmaditeshnizi2024optimus}.
A third direction positions the LLM as a hyper-heuristic generator that iteratively reflects upon and improves heuristic rule-code~\cite{huang2024automatic}.
Despite their potential, early works reveal critical challenges: the risk of generating infeasible solutions (``hallucinations''), high computational costs, and a primary focus on static problems, leaving their application to complex, dynamic scheduling environments largely unexploredc.

\textbf{Hierarchical Memory and Reasoning Paradigms:}
A significant body of work explores hierarchical paradigms to structure and access information beyond the fixed context window.
A prominent direction involves creating hierarchical memory systems, such as MemGPT's OS-inspired information paging~\cite{packer2024memgptllmsoperatingsystems} or the semantic-level memory structures in MemoryBank and H-MEM that enable targeted retrieval~\cite{zhong2023memorybankenhancinglargelanguage, sun2025hierarchicalmemoryhighefficiencylongterm}.
A related approach, seen in HIAGENT, employs hierarchy for task decomposition to manage an agent's working memory during long-horizon tasks~\cite{hu2024hiagenthierarchicalworkingmemory}.
Diverging from these methods, which primarily focus on organizing and retrieving an agent's past, our proposed ReflecSched framework introduces a hierarchical reflection mechanism that functions as a dynamic, planning-time operator.
Instead of retrieving historical data, it prospectively explores the future by leveraging multi-level, heuristic-driven simulations to analyze potential decision pathways.
The hierarchy in ReflecSched is thus one of planning abstraction distilling numerical simulations into a transient, textual Strategic Experience to guide the next action, rather than creating a persistent memory store.

\section{Preliminaries}

\subsection{Dynamic Flexible Job-Shop Scheduling}
\begin{figure}[!htbp]
\centering
\includegraphics[width=0.8\columnwidth]{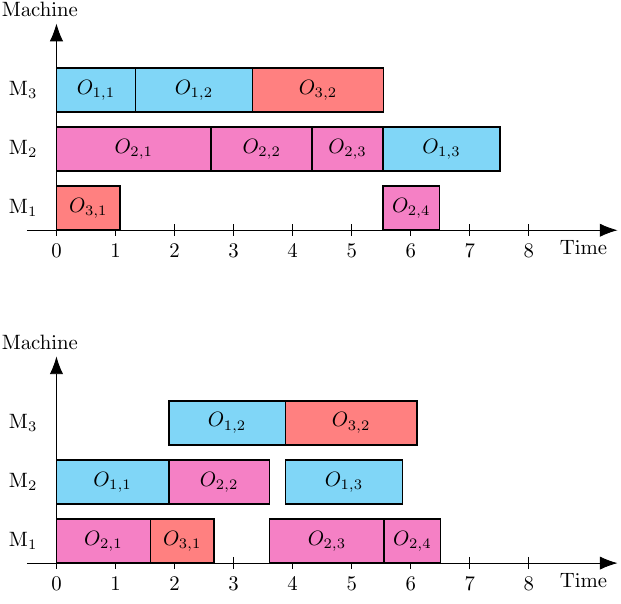} 
\caption{An example of two valid schedules for the same problem instance. The Gantt charts illustrate how a locally optimal choice at an early stage can lead to a suboptimal final makespan (top, 7.51), whereas a more globally-aware decision sequence yields a better outcome (bottom, 6.51).}
\label{fig:problem_gantt}
\end{figure}

DFJSP is the problem of assigning operations from multiple jobs to a set of machines, subject to real-time stochastic events.
It is characterized by two features: flexibility, which allows each operation to be processed on a subset of candidate machines, and dynamism, which refers to the occurrence of real-time stochastic events such as new job arrivals or machine breakdowns.
The objective is to generate a valid schedule that minimizes the makespan, defined as the maximum completion time among all jobs~\cite{cao2024novel}.
Figure~\ref{fig:problem_gantt} illustrates an example of such a schedule, visualized as a Gantt chart.
A valid schedule must satisfy both precedence constraints between operations of the same job and the resource constraints of the machines.

In this work, we model a rich set of dynamic events to capture the stochastic nature of modern manufacturing environments.
These include new job arrivals, machine breakdowns and subsequent repairs, job cancellations, and the introduction of high-priority emergency jobs.
This set of events is representative of the primary disruptions studied in the DFJSP literature~\cite{baykasoglu2020greedy, ngwu2025reinforcement, wang2022multi}.
More importantly, the core architecture of ReflecSched is designed to be event-agnostic and generalizable.
Its strategic Hierarchical Reflection module is invoked by an event-driven trigger, which responds to any significant state change rather than to a specific event type.
Consequently, the framework can be readily extended to handle other stochastic events, such as variable processing times~\cite{shahgholi2019heuristic} or rush order arrivals~\cite{ren2024dynamic}, without requiring modifications to its fundamental logic.

\begin{table}[H]
\centering
\caption{Notation for the DFJSP Mathematical Formulation}
\label{tab:notation}
\begin{tabular}{l p{0.8\columnwidth}}
\toprule
\multicolumn{1}{c}{\textbf{Symbol}} & \multicolumn{1}{c}{\textbf{Description}} \\
\midrule
\multicolumn{2}{l}{\textit{\textbf{Sets and Indices}}} \\
\textcolor{blue}{$t$} & \textcolor{blue}{Discrete decision step index, $t = 0, 1, 2, ...$} \\
\textcolor{blue}{$\tau$} & \textcolor{blue}{Continuous physical time index, $\tau \in [0, \infty)$.} \\
$\mathcal{J}$ & Set of active jobs currently in the system, $\{1, ..., n\}$. \\
$\mathcal{M}$ & Set of available machines, $\{1, ..., m\}$. \\
$\mathcal{O}$ & Set of all operations to be scheduled. \\
\textcolor{blue}{$\mathcal{O}_i$} & Set of operations for job $J_i \in \mathcal{J}$. \\
\textcolor{blue}{$\mathcal{M}_{ij}$} & \textcolor{blue}{Set of candidate machines for operation $O_{ij}$.} \\
\textcolor{blue}{$\mathcal{O}_{int}$} & Set of interrupted operations awaiting machine repair. \\
\midrule
\multicolumn{2}{l}{\textit{\textbf{Parameters and Constants}}} \\
$n_i$ & Number of operations for job $J_i$. \\
$p_{ijk}$ & Processing time of $O_{ij}$ on machine $k \in \mathcal{M}_{ij}$. \\
$a_i$ & Arrival time of job $J_i$. \\
\textcolor{blue}{$M_{big}$} & A sufficiently large positive number (a ``Big-M'' constant). \\
\midrule
\multicolumn{2}{l}{\textit{\textbf{Decision Variables}}} \\
$x_{ijk}$ & Binary: 1 if $O_{ij}$ is assigned to machine $k$. \\
$y_{ij,i'j'k}$ & Binary: 1 if operation $O_{ij}$ precedes $O_{i'j'}$ when both are assigned to machine $k$. \\
\midrule
\multicolumn{2}{l}{\textit{\textbf{Auxiliary and State Variables}}} \\
$s_{ij}$ & Start time of operation $O_{ij}$. \\
$c_{ij}$ & Completion time of operation $O_{ij}$. \\
\textcolor{blue}{$r_k$} & \textcolor{blue}{Earliest time machine $k$ becomes free.} \\
$C_{max}$ & The makespan, defined as \textcolor{blue}{$\max_{J_i \in \mathcal{J}} \{c_{i,n_i}\}$}. \\
\bottomrule
\end{tabular}
\end{table}

\subsection{Mathematical Formulation of the DFJSP}
\label{sec:mathematicalformulation}

The notation used throughout is summarized in Table~\ref{tab:notation}.

{\color{blue}
\subsubsection{Model Assumptions}
\label{subsec:modelassumptions}

To ensure clarity and specify the applicable scope of the proposed mathematical model, we explicitly state the following assumptions regarding the DFJSP environment~\cite{zhang2024dynamic}.
These assumptions define the boundaries of resource constraints, temporal dynamics, and event handling logic:

To define the operational scope of the proposed mathematical model, we stipulate the following assumptions regarding the DFJSP environment~\cite{zhang2026novel}. These assumptions delineate the parameters of resource constraints, temporal dynamics, and the logic governing dynamic events:

\textbf{Resource and Task Constraints:}
Each job follows a predefined sequence of operations, with each operation occupying a single machine at any given time.
Machines constitute the primary capacity constraints and are assumed to have infinite buffer capacity.
Auxiliary resources, such as operators and tools, are considered non-constraining and remain readily available.

\textbf{Temporal Specifications:}
Setup times are assumed to be sequence-independent and are incorporated into the operation processing times.
Similarly, inter-machine transfer times are considered negligible or are subsumed under the processing durations.
All processing times are deterministic and remain invariant throughout the execution of an operation.

\textbf{Dynamic Event Logic:}
The scheduler operates in an online environment where the timing and characteristics of stochastic events, such as job arrivals and machine breakdowns, are unknown a priori.
Operations are non-preemptive except when interrupted by equipment failures.
In such instances, we assume a preempt-resume logic: the accumulated processing progress is preserved, and the operation resumes as soon as the machine becomes available again.
}

\subsubsection{Objective Function and Constraints}
\label{subsec:objectivefunctionandconstraints}

The primary objective is to minimize the makespan ($C_{max}$), which is defined as the maximum completion time among all jobs.
\begin{equation}
\min C_{max}
\end{equation}
subject to:
\begin{equation}
C_{max} \ge c_{i,n_i} \quad \forall J_i \in \mathcal{J} \label{eq:makespan}
\end{equation}

\noindent The schedule must satisfy the following constraints:

\textbf{Processing Time:} The completion time of an operation is determined by its start time and the processing time on its assigned machine.
\begin{equation}
    c_{ij} - s_{ij} = \sum_{k \in M_{ij}} p_{ijk} \cdot x_{ijk} \quad \forall O_{ij} \in \mathcal{O} \label{eq:proc_time}
\end{equation}

\textbf{Assignment:} Each operation must be assigned to exactly one of its candidate machines.
\begin{equation}
    \sum_{k \in M_{ij}} x_{ijk} = 1 \quad \forall O_{ij} \in \mathcal{O} \label{eq:assignment}
\end{equation}

\textbf{Job Arrival:} An operation cannot start before its corresponding job has arrived.
\begin{equation}
    s_{ij} \ge a_i \quad \forall O_{ij} \in \mathcal{O} \label{eq:arrival}
\end{equation}

\textbf{Precedence:} Within each job, operations must be performed in their specified sequence.
\begin{equation}
    s_{i,j+1} \ge c_{ij} \quad \forall J_i \in \mathcal{J}, \forall j \in \{1, \dots, n_i-1\} \label{eq:precedence}
\end{equation}

\textbf{Machine Capacity:} For any two distinct operations, if they are assigned to the same machine, they cannot be processed simultaneously.
This is enforced by the following disjunctive constraints for any pair of operations $(O_{ij}, O_{i'j'})$ where $i \neq i'$ or $j \neq j'$, and for each machine $k \in M_{ij} \cap M_{i'j'}$:
\begin{align}
    s_{i'j'} &\ge c_{ij} - \textcolor{blue}{M_{big}} \cdot (3 - x_{ijk} - x_{i'j'k} - y_{ij,i'j'k}) \label{eq:machine_capacity1} \\
    s_{ij} &\ge c_{i'j'} - \textcolor{blue}{M_{big}} \cdot (2 + y_{ij,i'j'k} - x_{ijk} - x_{i'j'k}) \label{eq:machine_capacity2}
\end{align}

\subsubsection{Formalization of Dynamic Events}
\label{subsec:dynamicevents}

\textcolor{blue}{The state of the system at time $\tau$ is defined by the tuple $S_{\tau} = (\mathcal{J}_{\tau}, \{a_i\}, \{r_k\}, \mathcal{O}_{int})$.
A dynamic event occurring at time $\tau$ triggers an instantaneous state transition from \textcolor{blue}{$S_{\tau}$ to $S_{\tau^+}$}, where $\tau^+$ denotes the time immediately following the event execution.
The specific update logic for each event type is formalized below~\cite{baykasouglu2020greedy}.}

\textbf{Job Arrival:}
At \textcolor{blue}{physical} time \textcolor{blue}{$\tau = \tau_{new}$}, a new job $J_{new}$ with its own set of operations $\mathcal{O}_{new}$ is introduced into the system~\cite{shi2025deep}.
The state is updated \textcolor{blue}{instantaneously} as follows:
{\color{blue}
\begin{align}
    \mathcal{J}_{\tau^+} &\leftarrow \mathcal{J}_{\tau} \cup \{J_{new}\} \\
    \mathcal{O}_{\tau^+} &\leftarrow O_{\tau} \cup O_{new}
\end{align}
}
The emergency status of a job is a characteristic handled by the scheduling policy via prioritization rather than being a direct modification to the model's constraints.

\textbf{Job Cancellation:}
At \textcolor{blue}{physical} time \textcolor{blue}{$\tau = \tau_{cancel}$}, a job $J_{cancel}$ is removed from the system.
This action removes the job from the set of active jobs $\mathcal{J}$ and all of its corresponding uncompleted operations from the set of operations to be scheduled $\mathcal{O}_{cancel}$.
The historical record of any completed operations of \textcolor{blue}{$J_{cancel}$} is preserved, but they are no longer relevant for future scheduling decisions.

The state update is formalized as:
{\color{blue}
\begin{align}
    \mathcal{J}_{\tau^+} &\leftarrow \mathcal{J}_{\tau} \setminus \{J_{cancel}\} \\
    O_{\tau^+} &\leftarrow O_{\tau} \setminus O_{cancel}
\end{align}
}
If an operation from $\mathcal{O}_{cancel}$ was in process on a machine $k$ at \textcolor{blue}{$\tau$}, it is preempted. The machine is immediately freed, and its availability time \textcolor{blue}{$r_k$ is updated to $\tau$}.

\textbf{Machine Breakdown:}
At \textcolor{blue}{physical} time \textcolor{blue}{$\tau = \tau_{break}$}, a machine $k \in \mathcal{M}$ fails and becomes unavailable~\cite{li2025categorized}.
The immediate consequence is that the machine's earliest availability time is postponed until its projected repair time, \textcolor{blue}{$\tau_{repair}$}.
This is formalized as:
{\color{blue}
\begin{equation}
    r_{k} = \tau_{repair}
\end{equation}
}
This update affects all subsequent scheduling decisions, as no new operations can be assigned to machine $k$ until \textcolor{blue}{$\tau_{repair}$}.

If an operation $O_{ij}$ was being processed on machine $k$ at this time i.e., \textcolor{blue}{$s_{ij} < \tau_{break} < c_{ij}$}, it is immediately interrupted.
The operation is formally moved to a set of interrupted operations, $\mathcal{O}_{int}$, to signify its special status.
It retains its assignment to machine $k$ and awaits the machine's repair to resume processing.

\textbf{Machine Repair:}
At \textcolor{blue}{physical} time \textcolor{blue}{$\tau = \tau_{repair}$}, machine $k$ becomes operational again.
Any operation $O_{ij} \in \mathcal{O}_{int}$ that was interrupted on this machine can now resume.
Upon repair, $O_{ij}$ is removed from the set $\mathcal{O}_{int}$.

The resumption of its processing leads to a new, postponed completion time, $c'_{ij}$.
While the intrinsic processing time $p_{ijk}$ remains unchanged, the new completion time is calculated based on the remaining work:
{\color{blue}
\begin{equation}
    c'_{ij} = \tau + p_{ijk} - (\tau_{break} - s_{ij})
\end{equation}
}
where \textcolor{blue}{$\tau_{break} - s_{ij}$} represents the processing work completed before the interruption.
This new completion time $c'_{ij}$ subsequently serves as the basis for re-evaluating the precedence constraints for all downstream operations of job $J_i$.
The machine $k$ is also now available for scheduling other operations from time \textcolor{blue}{$\tau_{repair}$} onwards.

\begin{figure}[!htbp]
\centering
\includegraphics[width=0.8\columnwidth]{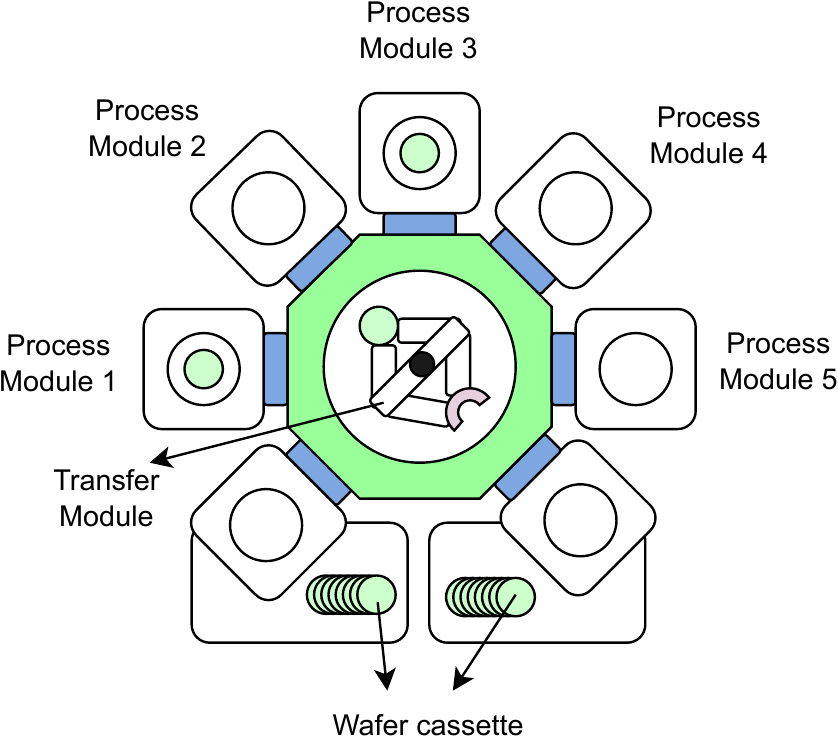} 
\caption{Architectural schematic of a semiconductor cluster tool utilized in the JMS-Bench dataset. The system integrates five heterogeneous process modules (PM1 to PM5) with a central transfer module and dual-arm robot. Wafers are loaded from cassettes and navigate through multi-stage recipes involving re-entrant flows and strict residency time constraints.}
\label{fig:cluster_tools}
\end{figure}

{\color{blue}
\subsection{Industrial Application of Semiconductor Cluster Tool Scheduling}

The DFJSP framework investigated in this study is motivated by the operational management of cluster tools within semiconductor wafer fabrication facilities.
As illustrated in Figure \ref{fig:cluster_tools}, a cluster tool is a representative manufacturing unit comprising several heterogeneous processing chambers, denoted as PM1 through PM5, which surround a central multi-link handling robot~\cite{pan2017scheduling}.
Wafers enter the system through load lock interfaces, after which the robot executes the sequence of transport steps required to move them between specific modules.
These chambers perform high-precision processes, such as chemical vapor deposition or reactive ion etching, where scheduling complexity is dictated by a set of technological constraints that transcend traditional job-shop models.

Machine eligibility is determined by wafer recipes, as only a specific subset of process modules is configured with the gas chemistry, RF power, or thermal settings required for a given technological step~\cite{xiong2020reducing}.
Furthermore, re-entrant flow patterns are prevalent in this environment; wafers must frequently revisit modules such as PM2 or PM3 to undergo multi-layer processing at various stages of the production cycle.
Yield management introduces a third critical constraint: residency time limits.
These constraints mandate that the handling robot extracts wafers from a chamber immediately upon processing completion, as prolonged exposure to the chamber environment renders the wafer surface vulnerable to oxidation or chemical contamination.
Consequently, the scheduler must precisely synchronize robot transport with chamber cycle times to mitigate defects induced by processing delays.

The operational dynamism of this fabrication environment is further characterized by frequent perturbations that necessitate the real-time reconfiguration of pre-planned schedules.
These disturbances encompass the stochastic arrival of high-priority ``hot lots'' requiring immediate prioritization to satisfy stringent delivery constraints, equipment failures resulting from parameter drifts, and job cancellations triggered by upstream quality excursions.
When a processing module becomes unavailable, the system must autonomously identify alternative routings through compatible modules to ensure production continuity.
Addressing these challenges requires a scheduling framework capable of non-myopic reasoning to optimize throughput while adhering to the physical constraints of the equipment.
Consequently, the coordination of semiconductor cluster tools provides a compelling case for the proposed hierarchical reflection mechanism, as it offers a structured paradigm for balancing transient efficiency with long-term schedule robustness~\cite{ahn2024novel}.

}

\subsection{Baseline LLM-based Scheduler}
To evaluate our proposed method, we define a direct baseline, LLM-Direct, which employs the LLM to perform the function of classical dispatching rules.
In this approach, a state-aware prompt is constructed at each decision point.
The shop-floor state is encoded into the prompt, including both static specifications, such as job structures and machine data, and dynamic information, such as the set of currently available actions and current machine statuses.
The action selected by the LLM is then executed in a shop-floor simulator, advancing the system state to the next decision point.
Further implementation details for LLM-Direct are presented in \ref{app:llm_direct_details}.

\section{Motivational Analysis}

\subsection{The Long-Context Paradox}
\label{sec:long-context}

While replacing state engineering with a single LLM prompt is appealing, it creates an information utility paradox: encoding the high-dimensional shop-floor state into a dense, verbose prompt leads to underutilization of the provided data~\cite{jiang2024longllmlinguaacceleratingenhancingllms}.
We observe this is not merely a context-length issue but a fundamental challenge in reasoning over such textual complexity~\cite{liu2023lostmiddlelanguagemodels}, as exemplified by our case study with smaller models in \ref{app:casestudy}.

\begin{figure}[!htbp]
    \centering
    \begin{subfigure}[t]{0.49\columnwidth}
        \centering
        \includegraphics[width=\columnwidth]{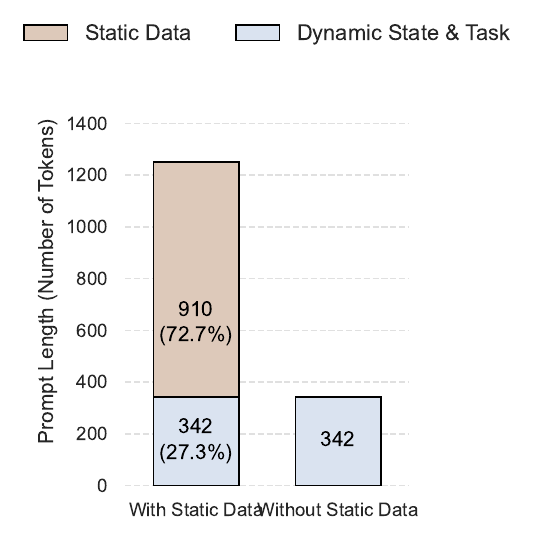}
        \caption{Prompt Composition}
        \label{fig:prompt_composition}
    \end{subfigure}
    \hfill
    \begin{subfigure}[t]{0.49\columnwidth}
        \centering
        \includegraphics[width=\columnwidth]{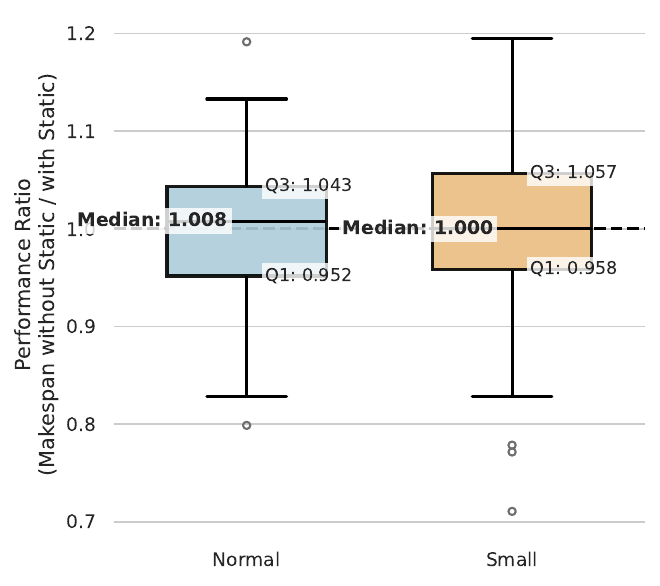}
        \caption{Performance Impact}
        \label{fig:performance_comparison}
    \end{subfigure}

    \vspace{10pt}

    \begin{subfigure}[t]{0.99\columnwidth}
        \centering
        \includegraphics[width=\columnwidth]{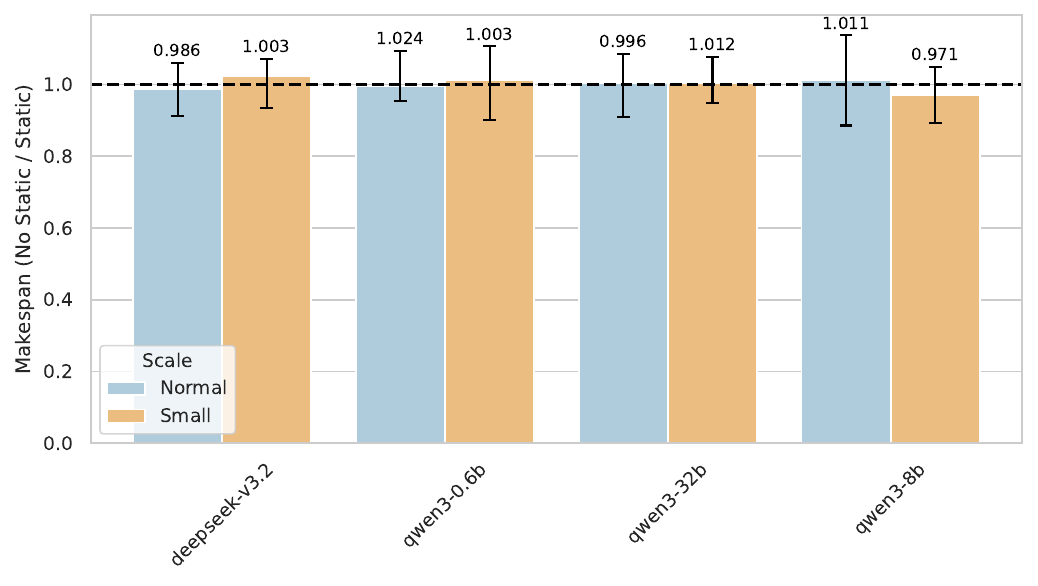}
        \caption{Cross-model Scaling Analysis}
        \label{fig:third_subfig}
    \end{subfigure}

    \caption{An empirical investigation of the Long-Context Paradox. (a) A breakdown of the prompt composition for the baseline model. (b) Box plots showing the ratio of makespans resulting from running the model with and without the static data portion of the prompt for the primary evaluation model. \textcolor{blue}{(c) Comparative scaling analysis of makespan ratios across the Qwen3 series and DeepSeek-v3.2. The consistent alignment of these ratios with unity across a parameter range of 0.6B to 685B indicates that the underutilization of dense static specifications is a systemic reasoning limitation that persists irrespective of model capacity.}}
    \label{fig:mainfig}
\end{figure}

To validate this hypothesis, we conducted a targeted experiment by omitting the static information block from the LLM-Direct baseline's prompt.
As shown in Figure~\ref{fig:mainfig}, this omission had a negligible impact on performance, even though the static block constitutes over 70\% of the prompt.
The median makespan ratio between the conditions (with vs. without static data) is nearly 1.0, providing strong evidence that the model largely ignores this information and fails to ground its decisions in the provided specifications.
\textcolor{blue}{To test this hypothesis, we conducted experiments across a broad spectrum of model scales, ranging from the Qwen series (0.6B, 8B, and 32B)~\cite{yang2025qwen3technicalreport} to the frontier-scale DeepSeek-V3.2 (685B)~\cite{deepseekai2025deepseekv32pushingfrontieropen}.
Specifically, we systematically excluded the static information block from the LLM-Direct baseline’s prompt to evaluate whether increasing model capacity can compensate for the absence of explicit structural constraints.}

As shown in Figure~\ref{fig:mainfig}, this omission had a negligible impact on performance \textcolor{blue}{across all tested scales}, even though the static block constitutes over 70\% of the prompt.
\textcolor{blue}{The experimental results reveal that the median makespan ratio between the two conditions (with vs. without static data) approaches unity, even for the 685B parameter model. This suggests that the models, irrespective of parameter scale, exhibit a notable insensitivity to this information and demonstrate a limited capacity to ground their decisions in the provided specifications.}

\textcolor{blue}{
The systemic neglect of context is further evidenced by the localized semantic errors analyzed in Appendix~\ref{app:case_study:rem_work=0}.
These instances reveal that the model often fails to incorporate critical state descriptors despite their explicit inclusion in the prompt.
Such findings indicate that the primary obstacle is not a lack of descriptive clarity but rather a fundamental deficiency in semantic grounding when processing high-density context.
Attempting to rectify these errors by further refining the prompt only serves to exacerbate the long-context paradox.
Specifically, increasing the granularity of descriptions to ensure information capture inevitably expands the total prompt volume.
As demonstrated by our scaling analysis, this expansion leads to the attenuation of information saliency and ultimately reinforces the model's reliance on parametric priors over the provided task-specific instructions.}

\subsection{Underutilization of Heuristics}
\label{sec:pdr_underutilization}

A key challenge in LLM-based scheduling is the effective application of PDRs.
While LLMs are generally proficient at following declarative instructions, their capacity to faithfully execute complex, state-dependent procedural rules remains unreliable~\cite{mu2024llmsfollowsimplerules}.
\textcolor{blue}{To quantify and isolate this competency, we introduced PDR-Bench, a diagnostic benchmark wherein each instance is systematically synthesized to possess a single, ground-truth optimal heuristic, and utilized it within a controlled experiment to evaluate the baseline LLM, the technical specifics of which are elaborated in Appendix \ref{app:pdr_experiment_details}.}

\begin{figure}[!htbp]
\centering
\includegraphics[width=0.9\columnwidth]{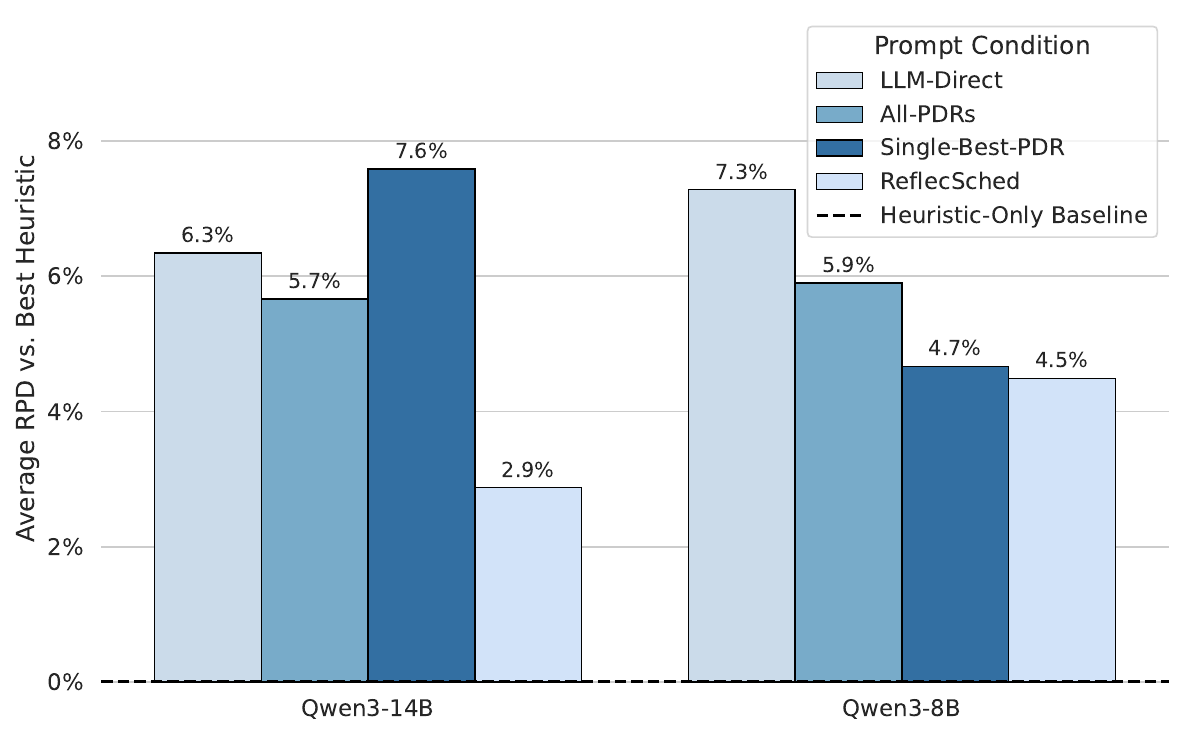} 
\caption{Analysis of heuristic utilization on the PDR-Bench dataset, with RPD measured against the known optimal heuristic for each instance (0\% baseline). The baseline LLM struggles to apply the optimal heuristic even when explicitly prompted, whereas ReflecSched substantially improves performance.}
\label{fig:pdr_utilization}
\end{figure}

The ``optimal heuristic'' for any given instance in PDR-Bench is known by design, as each problem instance is specifically constructed to ensure that one particular heuristic yields a superior outcome over all others.
The results \textcolor{blue}{presented in Figure~\ref{fig:pdr_utilization}} reveal the LLM's significant difficulty in effectively utilizing these heuristics~\cite{uzunoglu2024paradiseevaluatingimplicitplanning}.
This finding \textcolor{blue}{suggests} the model's \textcolor{blue}{limited capacity} to translate the strategic guidance of the provided rules into its decision-making.
Furthermore, when explicitly instructed to use the single optimal heuristic, the model still failed to match the performance of that heuristic applied in isolation, reverting instead to its generalized, pre-trained behaviors.

\subsection{The Pitfall of Myopic Greed}
\label{sec:myopic_greed}

The autoregressive, token-by-token generation process inherent in LLMs presents a structural mismatch with the strategic, non-local search required for complex scheduling problems.
This mismatch predisposes them to myopic decision-making, where locally optimal choices lead to globally suboptimal outcomes.

To quantify this myopic behavior, we introduce the \textbf{Greedy Decision Ratio (GDR)}.

We formally define the one step greedy action at any decision point $t$ as the one
that minimizes the local finish time.
\textcolor{blue}{Let $\mathcal{A}(S_\tau)$ denote the set of all currently feasible actions at state $S_\tau$, where each action $\mathbf{a} \in \mathcal{A}(S_\tau)$ represents a valid assignment of a ready operation to an available machine.
The greedy action is defined as:}

{\color{blue}
\begin{equation}
\mathbf{a}_t = \arg\min_{\mathbf{a} \in \mathcal{A}(S_\tau)} (\tau_{start}(\mathbf{a}) + p_\mathbf{a})
\end{equation}
}
where \textcolor{blue}{$\tau_{start}(\mathbf{a})$} is the earliest start time of an action and \textcolor{blue}{$p_\mathbf{a}$} is its processing time.
The GDR is then the fraction of decision points where the agent's chosen action matches \textcolor{blue}{$\mathbf{a}_t$}.

\begin{figure}[!htbp]
\centering
\includegraphics[width=0.95\columnwidth]{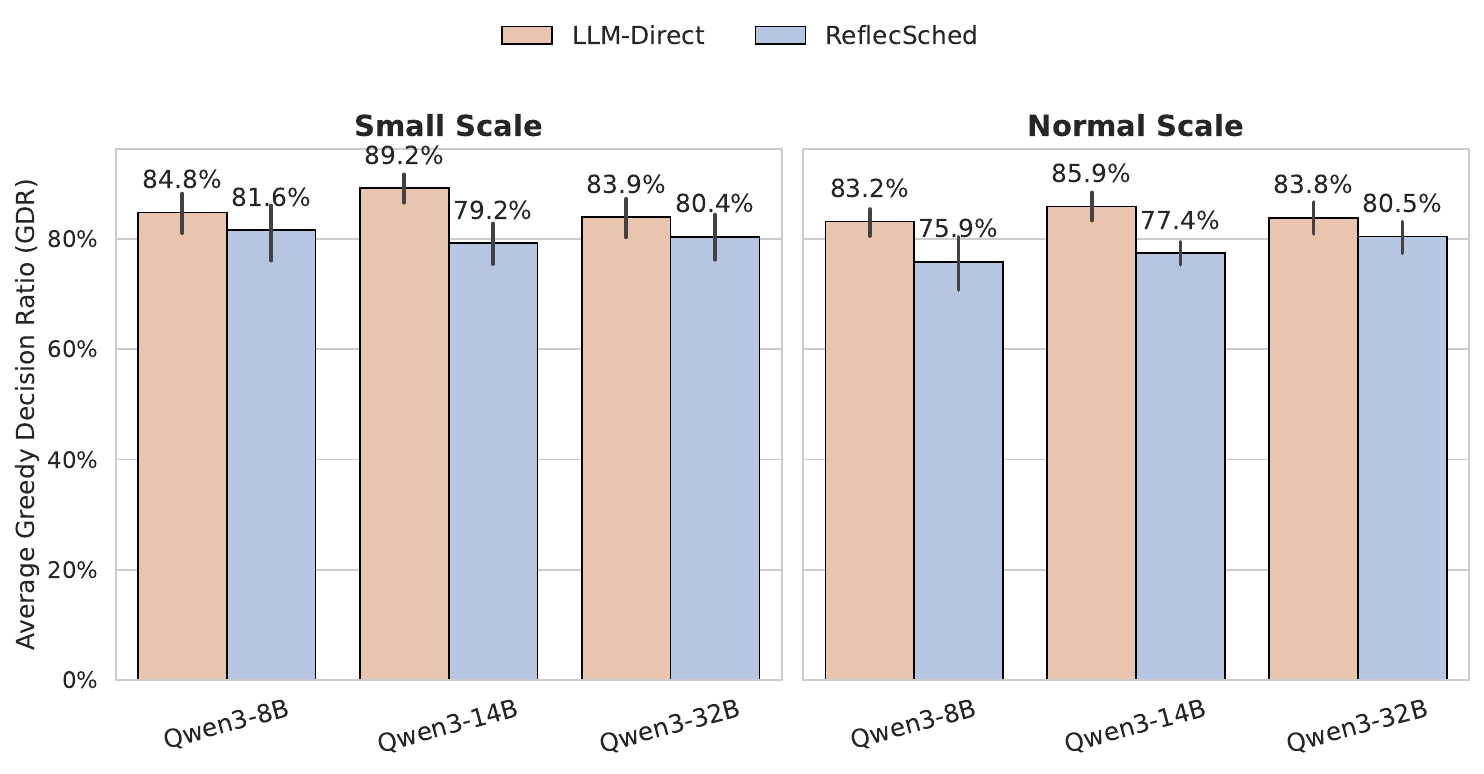} 
\caption{Comparison of the Average Greedy Decision Ratio (GDR) between the LLM-Direct baseline and the ReflecSched framework. Results are presented for different models and are segmented by the Normal and Small problem scales.}
\label{fig:greedyratio}
\end{figure}

We evaluated baseline LLMs using this metric on GEN-Bench, a benchmark specifically designed to be globally balanced and control for what we term rule-balance bias---a scenario where a dataset might implicitly favor a single, simple heuristic.
This curation is critical: on a rule-balanced benchmark, a persistently high GDR ($\approx$80-90\%) is a strong indicator of intrinsic myopic decision-making rather than an artifact of the model correctly identifying a simple optimal heuristic for the dataset.
The quantitative results \textcolor{blue}{presented in Figure~\ref{fig:greedyratio}} reveal that the GDR is indeed persistently high across all tested models, indicating a significant greedy bias~\cite{baeumel2025lookaheadlimitationmultioperandaddition}.

A qualitative case study \textcolor{blue}{presented in Figure~\ref{fig:problem_gantt}} further illustrates the consequences, showing how a single myopic choice can trigger a cascade of downstream inefficiencies.
\textcolor{blue}{The complete data for this instance is provided in \ref{app:figure1_data}.}

\textcolor{blue}{
Specifically, consider the resource contention at $t=0$, where operations $O_{2,1}$ and $O_{3,1}$ are both available for processing.
A myopic scheduling logic typically prioritizes $O_{3,1}$ because it exhibits zero machine flexibility, being restricted solely to $M_1$.
This immediate allocation occupies $M_1$ until $t=1.08$.
Consequently, when the greedy policy subsequently evaluates $O_{2,1}$, it identifies two alternatives: waiting for $M_1$ to become available at $t=2.67$ or utilizing the idle $M_2$ to complete the operation at $t=2.62$.
The greedy policy selects $M_2$ to achieve a marginal local gain of 0.05 time units.
This decision is suboptimal as it preemptively occupies $M_2$ for an extended duration, thereby obstructing the critical path for Job 1's final operation, $O_{1,3}$, which requires $M_2$ until $t=5.53$.
In contrast, the strategic policy accounts for this downstream bottleneck.
By executing a global trade-off, it reverses the assignment sequence and allocates $O_{2,1}$ to $M_1$ at $t=0$.
Although this choice introduces a slight delay for $O_{3,1}$, it reserves $M_2$ capacity for Job 1.
As a result, $O_{1,3}$ can commence at $t=3.88$, yielding a makespan of 6.51, which is a significant improvement over the 7.51 achieved by the greedy approach.}

\section{ReflecSched Framework}
\label{sec:the_framework}

\subsection{Framework Overview}
\label{sec:framework_overview}
\begin{figure*}[!t]
\centering
\includegraphics[width=\textwidth]{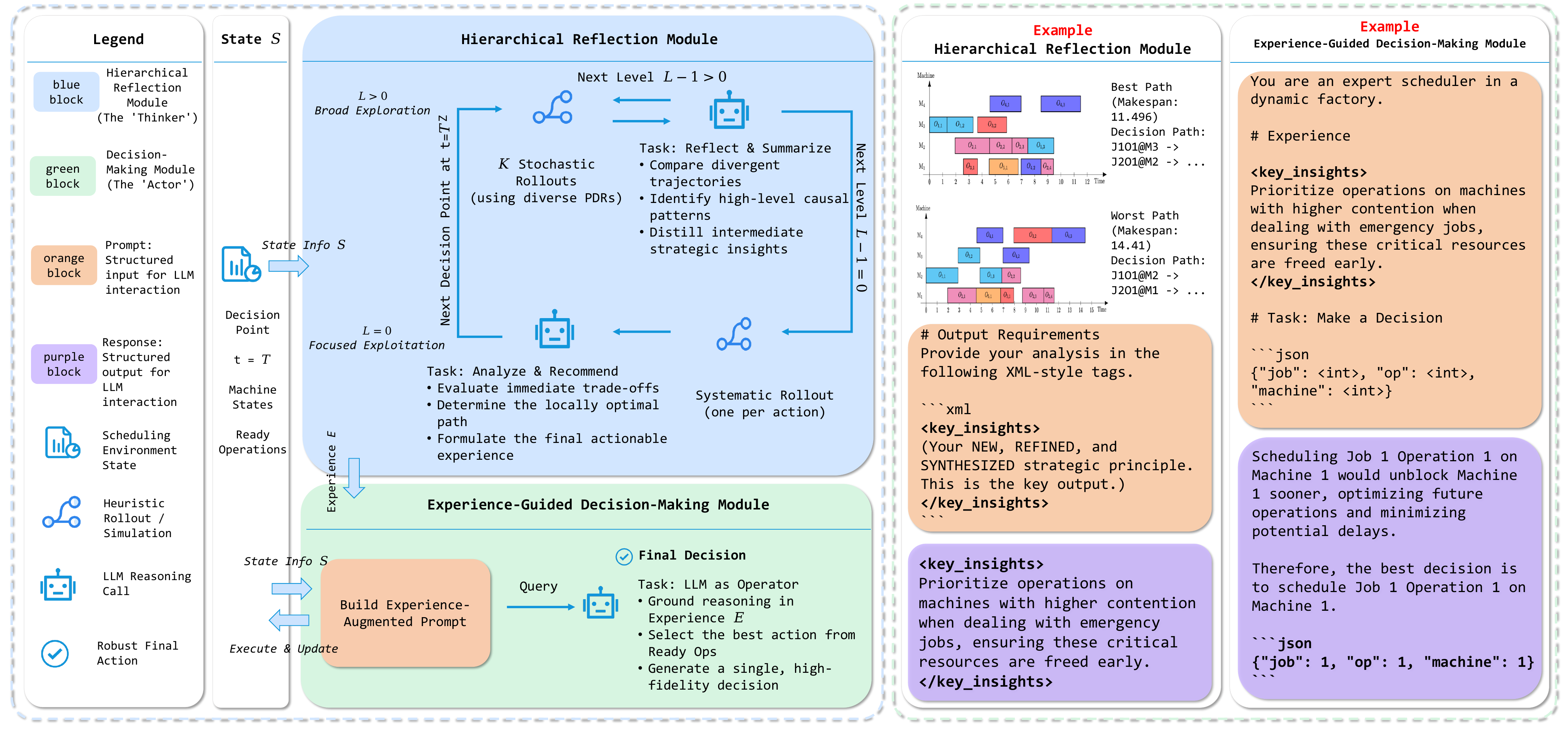}
\caption{The architecture of the ReflecSched framework, designed to decouple strategic planning from immediate execution. The main workflow is initiated at a decision point. The \textbf{Hierarchical Reflection Module} (blue) performs multi-level, heuristic-driven simulations to explore future state trajectories. It then leverages an LLM to analyze these outcomes and distill a concise Strategic Experience ($\mathcal{E}$). This experience is passed to the \textbf{Experience-Guided Decision-Making Module} (green), which constructs an experience-augmented prompt to guide the LLM in selecting a final, strategically-informed action. The right-hand panels provide concrete examples of the prompts and responses for both the reflection and decision-making stages, illustrating the flow of information from simulation data to strategic guidance, and finally to a specific action.}
\label{fig:framework_arch}
\end{figure*}

To address the identified limitations, we introduce ReflecSched, a novel framework that structurally decouples long-horizon, strategic reasoning from low-latency, online decision-making via two interconnected, sequential modules (Figure~\ref{fig:framework_arch})~\cite{yao2023react}.
The first, the Hierarchical Reflection Module, serves as the framework's strategic planning component.
It leverages multi-level, heuristic-driven simulations to explore future state trajectories~\cite{yao2023tree}.
It then employs the LLM to analyze these simulated outcomes, distilling them into a concise, actionable Strategic Experience $\mathcal{E}$.
This experience then informs the second component, the Experience-Guided Decision-Making Module, which handles real-time execution.
Guided by the strategic foresight in $\mathcal{E}$, this module makes a strategically-informed decision based on the immediate environment state.

This two-stage architecture is designed to address each of the identified challenges:
First, by explicitly simulating future consequences, the Hierarchical Reflection Module provides the necessary foresight to counteract myopic greed, shifting the model's focus from locally optimal choices to globally efficient strategies.
Second, the framework directly addresses the underutilization of heuristics by employing PDRs to guide its exploratory rollouts~\cite{zhang2024restmctsllmselftrainingprocess}.
This ensures that expert procedural knowledge is integrated into the reasoning process.
Finally, the framework mitigates the long-context paradox.
By having the Reflection Module distill its findings into a concise Strategic Experience $\mathcal{E}$, the final decision prompt supplied to the LLM contains only this high-level guidance and the immediate state~\cite{jiang2024longllmlinguaacceleratingenhancingllms}.
This precludes informational overload and concentrates the model's reasoning on the most salient data.

\textcolor{blue}{ReflecSched mitigates the inherent opacity of traditional models by formalizing reasoning processes into explicit natural language representations.
In contrast to DRL policies, where decision logic is implicitly encoded within high-dimensional parameter spaces, this framework generates intermediate textual artifacts designated as Strategic Experience~\cite{wang2022multi, liu2024dynamic}.
These artifacts serve as an interpretable interface between complex simulation states and final scheduling executions.
By extracting discriminative features from successful and suboptimal trajectories into explicit heuristics, the framework enables human operators to verify the strategic rationale underlying each decision.
Furthermore, the integration of Chain-of-Thought prompting within the decision module facilitates the generation of traceable logical sequences for each assignment~\cite{wei2022chain}.
This transparency allows for the identification of potential reasoning failures and enhances operational verifiability, representing a distinct functional departure from purely numerical policies that lack explanatory mechanisms.}

\subsection{Hierarchical Reflection}
\label{sec:hierarchical_reflection}

Central to ReflecSched is the Hierarchical Reflection module, a mechanism designed to perform strategic evaluation, which is essential for the DFJSP.
\textcolor{blue}{From a structural perspective, the proposed hierarchy is characterized by an inverse relationship between planning granularity and temporal horizon.
Rather than employing a monolithic reasoning architecture, we decompose the reflective process into distinct levels where the analytical focus transitions from long-term strategic objectives to immediate tactical execution.}
It implements a recursive ``Simulate-Reflect-Refine'' loop~\cite{gui2025hypertreeplanningenhancingllm}.
This iterative process first explores the future through simulation, then reflects on the outcomes to identify key strategic principles, and finally refines these principles into a concise, actionable policy directive~\cite{shinn2023reflexion}.

The process is a top-down hierarchical simulation from $l_{max}$ to $l=0$, operating as a deterministic projection that assumes no new stochastic events during the lookahead.
Central to this simulation is a randomized base policy, $\pi_{base}$, which operates by sampling a PDR from a predefined pool at each decision point to select an action.
The simulation methodology differs by level to enable multi-timescale exploration.
At higher levels ($l>0$), the module performs sparse, long-range exploration \textcolor{blue}{with reduced granularity} by simulating trajectories where $\pi_{base}$ is applied at every step \textcolor{blue}{to identify global constraints such as future machine contention}.
In contrast, the base level ($l=0$) performs a systematic evaluation of immediate actions \textcolor{blue}{to determine the high-resolution impact of each candidate within a narrow temporal window}.
For each available action \textcolor{blue}{$\mathbf{a} \in \mathcal{A}(S_\tau)$}, a short-horizon rollout is initiated by first executing the action $a$, and then applying $\pi_{base}$ for all subsequent decisions.

The outcome of any such rollout is a trajectory \textcolor{blue}{$\zeta^{(l)}$} and its associated cost, defined as the hypothetical partial makespan:
{\color{blue}
\begin{equation}
\label{eq:truncated_cost_variable}
\hat{J}^{(l)}(S_\tau) = C_{\max}(S_{\text{end}}(\zeta^{(l)}))
\end{equation}
}
{\color{blue}where $S_{\text{end}}(\zeta^{(l)})$ denotes the terminal system state reached after simulating the trajectory $\zeta^{(l)}$ for a planning horizon of $H_l$ decision steps. Intuitively, $\hat{J}^{(l)}(S_\tau)$ represents the estimated completion time of the schedule if the current policy were to be followed for $H_l$ steps.}

Following the simulation at a given level $l$, the reflection phase distills actionable insights from the raw trajectory data.
This approach is designed to clearly delineate the attributes of successful versus unsuccessful strategies by focusing on extremal outcomes.
{\color{blue}Let $\mathcal{Z}(S_\tau)$ denote the set of all such valid simulated trajectories starting from the current state $S_\tau$.} 
Specifically, the module identifies the trajectories that resulted in the minimum and maximum cost estimates from the set of simulations performed at that level:

{\color{blue}
\begin{align}
\label{eq:best_worst_paths_variable}
    \zeta^{(l)}_{\mathrm{best}} &= \arg\min_{\zeta^{(l)} \in \mathcal{Z}(S_\tau)} \hat{J}^{(l)}(S_\tau) \\
    \zeta^{(l)}_{\mathrm{worst}} &= \arg\max_{\zeta^{(l)} \in \mathcal{Z}(S_\tau)} \hat{J}^{(l)}(S_\tau)
\end{align}
}

Finally, in the refinement phase, these two contrasting trajectories are provided to the LLM, which acts as a higher-level policy synthesizer~\cite{light2025strategistselfimprovementllmdecision}.
Its task is not merely to summarize but to identify the key distinguishing features between \textcolor{blue}{$\zeta^{(l)}_{\mathrm{best}}$} and \textcolor{blue}{$\zeta^{(l)}_{\mathrm{worst}}$} and distill these observations into a concise, textual strategic guideline, termed the Strategic Experience $\mathcal{E}$.
This synthesis, represented by the function $F_{LLM}$, effectively transforms high-dimensional, numerical simulation data into low-dimensional, human-interpretable strategic guidance:
{\color{blue}
\begin{equation}
\label{eq:experience_synthesis_final_variable}
\mathcal{E} = F_{LLM}\left( \zeta^{(l)}_{\mathrm{best}}, \zeta^{(l)}_{\mathrm{worst}} \right)
\end{equation}
}
To ensure operational efficiency, this computationally intensive reflection process is invoked selectively.
It is triggered only upon the occurrence of a dynamic event (e.g., a new job arrival or machine breakdown), as such an event may render the existing Strategic Experience $\mathcal{E}$ suboptimal or obsolete.

\subsection{Experience-Guided Decision-Making}
\label{sec:experience_guided_decision}

The strategic experience $\mathcal{E}$ distilled by the reflection module provides the long-horizon context that LLM-Direct schedulers lack.
This module translates this high-level guidance into a specific, executable action by constructing a concise, experience-augmented prompt.
This prompt, containing only the immediate dynamic state and the guidance from $\mathcal{E}$, is then used to query the LLM for a final decision.

Theoretically, this process connects to the principles of \textbf{Approximate Policy Iteration (API)}~\cite{Gordon1995StableFunction}.
The validity of this connection, and the resulting performance proposition, hinges on two key working assumptions regarding the reflection process: (1) the rollout cost \textcolor{blue}{$\hat{J}^{(l)}(S_\tau)$} serves as a reasonable approximation of the true cost-to-go function for the base policy $\pi_{\mathrm{base}}$; and (2) the LLM operates as a faithful synthesizer, meaning its generated experience $\mathcal{E}$ correctly reflects the superiority of the best simulated trajectory over the worst (Faithful Reflection).

The experience \textcolor{blue}{$\mathcal{E}$} generated under these conditions is then used to define the final policy \textcolor{blue}{$\pi_{\mathcal{E}}$}, which makes a greedy decision with respect to the guidance~\cite{schmied2025llmsgreedyagentseffects}:
{\color{blue}
\begin{equation}
\label{eq:greedy_policy_final}
\mathbf{a}^{\star} = \pi_{\mathcal{E}}(S_\tau)
\end{equation}
}
With the policy thus defined, our framework can be understood through the lens of classical rollout algorithms~\citep{bertsekas2012dynamic}, suggesting a conditional improvement over the base policy.

\begin{proposition}[Conditional Policy Improvement]
\label{prop:rollout_improve}
Let \textcolor{blue}{$\pi_{\mathcal{E}}$} be the policy defined above and \textcolor{blue}{$V^{\pi}(S_\tau)$} be the expected makespan from state \textcolor{blue}{$S_\tau$} under policy $\pi$.
If the assumptions of Cost Function Approximation and Faithful Reflection hold, then for all states \textcolor{blue}{$S_\tau$}, the expected performance of \textcolor{blue}{$\pi_{\mathcal{E}}$} is no worse than that of the base heuristic policy, $\pi_{\mathrm{base}}$.
This relationship holds over the expectation of stochastic events in the environment and the randomness of the base policy:
\begin{equation}
\color{blue}
\mathbb{E}[V^{\pi_{\mathcal{E}}}(S_\tau)] \;\le\; \mathbb{E}[V^{\pi_{\mathrm{base}}}(S_\tau)]
\end{equation}
\end{proposition}

Proposition \ref{prop:rollout_improve} establishes a conditional theoretical proposition, grounding ReflecSched in established reinforcement learning principles.
It suggests that, under plausible assumptions, the framework's performance is not expected to degrade below that of the underlying heuristics.
It is important to note that the simulated lookahead assumes no new dynamic events occur within its finite planning horizon.
While a formal proof of the underlying assumptions is intractable for black-box LLMs, our empirical study in Section~\ref{sec:experiments} shows a consistent and significant reduction in makespan.
This suggests that these assumptions hold sufficiently well in practice for performance improvements to be realized.
This combination of a conditional theoretical foundation and strong empirical performance suggests that ReflecSched is an effective and well-grounded framework for dynamic scheduling.

{\color{blue}
Proposition \ref{prop:rollout_improve} relies on the Faithful Reflection assumption, which posits that the LLM can identify key factors that differentiate successful from unsuccessful trajectories with reasonable consistency.
While formal guarantees for LLM reasoning remain limited, this assumption in ReflecSched is motivated by the contrastive design of the reflection prompt.
By presenting paired trajectories with high and low rollout evaluated performance ($\zeta_{best}^{(l)}$ and $\zeta_{worst}^{(l)}$), the framework reframes scheduling into a structured comparison that can be easier for LLMs than direct generation, as suggested by our empirical observations.
We nevertheless note practical limitations.
The usefulness of the reflection output depends on simulation complexity, since longer planning horizons or larger machine counts can dilute informative differences across trajectories and lead to suboptimal guidance.
Moreover, the Cost Function Approximation assumption requires that heuristic based rollouts provide a useful surrogate for the cost to go, which can break down when the randomized base policy $\pi_{base}$ explores a narrow set of trajectories.
In practice, the hierarchical refinement process is intended to partially alleviate these issues by integrating feedback across multiple temporal resolutions before the final decision is made.
}

{\color{blue}
\subsection{Illustrative Analysis of the Hierarchical Planning and Refinement Loop}
\label{subsec:illustrative_walkthrough}

To illustrate the hierarchical decision process, we analyze the initial dispatching decision at $t=0$ for the instance in Figure \ref{fig:problem_gantt}.
Operation $O_{1,1}$ is ready to be dispatched, and two eligible machines are available. Option A assigns $O_{1,1}$ to $M_3$ with a processing time of 1.34 time units, but this choice can create a downstream bottleneck because $M_3$ will later be needed by operation $O_{3,2}$.
Option B assigns $O_{1,1}$ to $M_2$ with a processing time of 1.90 time units, which keeps $M_3$ available for $O_{3,2}$ at a later stage.
This trade off between immediate processing time and future resource contention provides a test case for whether different planning levels can select the action that yields a lower overall objective.

\textbf{Long Horizon Reflection ($l=2$).}
At this level, the scheduler runs longer simulations to capture downstream dependencies, with a horizon that is long enough to include the release of future operations such as $O_{3,2}$.
Using multiple rollouts under a randomized base policy $\pi_{base}$, we observe that assigning $O_{1,1}$ to $M_3$ based solely on immediate processing time can induce substantial downstream delay, since $O_{3,2}$ later competes for $M_3$ when the machine is still occupied.
The resulting Strategic Experience $\mathcal{E}_2$ highlights $M_3$ as a potential future contention point and suggests that preserving its capacity for upcoming high priority operations can outweigh the small benefit of reducing the immediate processing time.

\textbf{Medium Horizon Refinement ($l=1$).}
In this stage, the scheduler shortens the planning horizon to focus on the interval between the current state and the downstream contention point highlighted by $\mathcal{E}_2$.
Using an additional set of rollouts with a shorter planning horizon, the results indicate that dispatching $O{1,1}$ to machine $M_2$ leads to lower downstream congestion, whereas dispatching it to $M_3$ increases contention for $M_3$ even within this shorter window.
The refined strategy $\mathcal{E}_1$ therefore recommends trading a small increase in immediate processing time on $M_2$ for improved availability of $M_3$ for the priority operations.

\textbf{Short Horizon Execution ($l=0$).}
At $l=0$, the scheduler performs per action evaluation using a short planning horizon via a forced rollout procedure.
For Option A, the rollout conditions on assigning $O_{1,1}$ to machine $M_3$ and computes a rollout estimated makespan, interpreted in light of the strategic guidance summarized at higher levels.
While this choice reduces the immediate processing time of $O_{1,1}$, the subsequent simulation captures the emergence of downstream contention on $M_3$, leading to a higher rollout cost.
For Option B, the rollout conditions on assigning $O_{1,1}$ to $M_2$, under which $M_3$ remains available when operation $O_{3,2}$ becomes relevant, yielding a lower rollout cost.
The scheduler therefore selects Option B, illustrating how the hierarchical procedure can expose longer range trade offs that a single horizon planner may miss.
 
To provide a step by step illustration of the cost computation in Eq.~\ref{eq:truncated_cost_variable}, we evaluate the candidate actions at $t=0$ using the deterministic processing times reported in Table~\ref{tab:figure1_instance}.
All rollouts are simulated in a disruption free environment, meaning that no exogenous events such as new job arrivals or machine breakdowns occur during the lookahead horizon.
In each rollout, the randomized base policy $\pi_{base}$ samples actions from a set of 24 priority dispatching rule combinations at every decision point, which promotes exploration and reduces dependence on any single heuristic.
For Option A, which fixes the assignment $O_{1,1} \rightarrow M_3$, if one sampled trajectory yields completion times $C_1 = 7.51$ and $C_2 = 6.80$, then the simulated makespan for that trajectory is $\hat{J}^{(0)} = \max(C_1, C_2) = 7.51$.
For Option B, the simulated makespan can be lower, for example 6.51, when preserving capacity on $M_3$ allows the downstream operation to complete earlier.
This contrast provides the numerical signal used by the Hierarchical Reflection module to refine its guidance.
Although higher levels focus on bottleneck identification, the base level operates with a shorter planning horizon over the same decision process; both levels use the same deterministic simulator and the same PDR sampling scheme to produce comparable rollout based guidance.
}

{\color{blue}
\subsection{Industrial Practicality and Human-Machine Collaboration}
Deploying ReflecSched on the shop floor requires addressing system integration constraints and operator facing requirements.
In modern wafer fabrication facilities, the framework is designed as a scheduling decision support layer that interoperates with existing Manufacturing Execution Systems (MES) and Enterprise Resource Planning (ERP) platforms~\cite{liu2002aps}.
Integration is implemented via a modular data pipeline.
The MES supplies the latest status snapshot of process chambers and wafer batches, and this structured data is converted into a textual prompt with a fixed schema for the LLM.
The resulting design is intended to support timely responses to operational events by using established interfaces to send scheduling recommendations back to the equipment control layer~\cite{huang2025leveraging}.

ReflecSched can offer practical advantages over common metaheuristic schedulers and DRL approaches in production settings.
Metaheuristic methods typically require problem specific encodings, neighborhood operators, and parameter tuning, which can make rapid adaptation to atypical disturbances challenging.
DRL based schedulers often provide limited interpretability for operators and may require additional training or fine tuning when chamber configurations or process conditions change~\cite{immordino2025explainable}.
In contrast, our framework leverages pretrained language representations to support transfer across diverse operating conditions.
A key component is the synthesis of Strategic Experience, which provides a longer horizon perspective on production flows with reduced reliance on task specific feature engineering.
This flexibility is relevant in semiconductor manufacturing, where product mixes, recipes, and tool configurations evolve over time.

Furthermore, the textual form of Strategic Experience can improve transparency and support communication of decision rationale.
Compared with priority score based dispatchers that output only numerical values, ReflecSched provides textual explanations, for example by highlighting a downstream bottleneck or recommending that a critical chamber remain available for a future operation.
Such explanations allow shop floor supervisors to inspect the strategic intent of the recommendation and assess its consistency with operational priorities.
Improved interpretability may reduce overrides that arise primarily from limited understanding, a challenge often reported in complex manufacturing settings.
In addition to makespan improvements observed in our experiments, ReflecSched can function as an interpretable decision support module that helps human schedulers understand trade offs in multi stage cluster tool operations.
}

\section{Experiments}
\label{sec:experiments}

\begin{figure*}[!t]
    \centering
    \includegraphics[width=\textwidth]{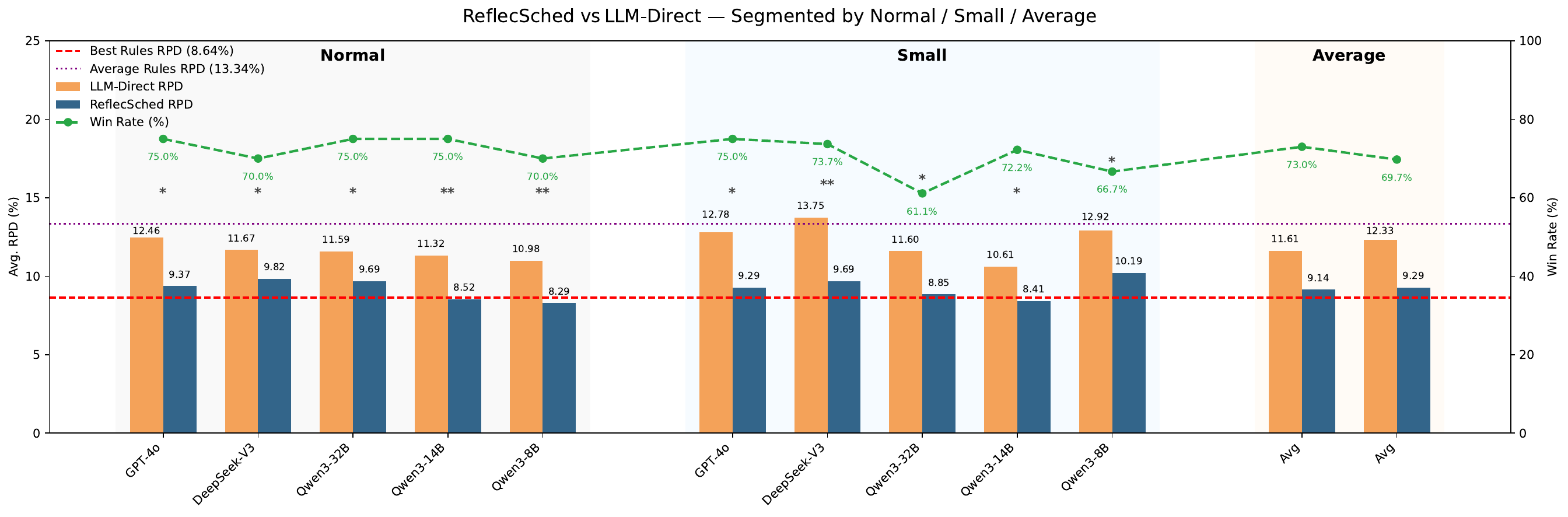}
    \caption{Comparative performance of ReflecSched and the LLM-Direct baseline on the GEN-Bench dataset. Bars (left y-axis) represent the average RPD, where lower is better. The green line (right y-axis) shows the Win Rate of ReflecSched against the baseline, indicating the percentage of instances where it finds a strictly better solution. For reference, the performance of the single best-performing PDR (i.e., the heuristic with the lowest average RPD across all instances) and the average of all PDRs are shown as red and purple dashed lines, respectively. Asterisks denote statistical significance from Wilcoxon signed-rank tests (* \textit{p} < 0.05, ** \textit{p} < 0.01, *** \textit{p} < 0.001).}
    \label{fig:main_results_barchart}
\end{figure*}

We conduct a series of experiments to rigorously evaluate our proposed ReflecSched framework.
In Section \ref{sec:exp_setup}, we detail the experimental setup, including the benchmark datasets, evaluation metrics, and models.
In Section \ref{sec:main_results_and_analysis}, we present a comparative performance analysis of ReflecSched against baseline methods and traditional heuristics.
In Section \ref{sec:ablation_and_sensitivity}, we conduct ablation studies to isolate the contribution of the framework's core components and analyze the sensitivity of its key hyperparameters.
Finally, in Section \ref{sec:efficiency_analysis}, we analyze the computational cost of the framework in terms of token consumption.

\subsection{Experimental Setup}
\label{sec:exp_setup}

\textbf{Benchmark Datasets.}
Our evaluation is performed on two benchmark suites developed for this study: GEN-Bench, for general-purpose evaluation, and PDR-Bench, a diagnostic benchmark to assess heuristic utilization. Both suites contain instances at two scales: Normal and Small. \textcolor{blue}{Specifically, GEN-Bench consists of 20 instances for the Normal scale and 18 instances for the Small scale, while PDR-Bench includes 111 instances for the Normal scale and 12 instances for the Small scale.} The detailed generation and curation methodology for these datasets is described in \ref{app:dataset_curation}.

\textcolor{blue}{To evaluate cross-benchmark generalizability, we additionally report results on MK-Bench and JMS-Bench. MK-Bench is derived from the Brandimarte MK01 to MK10 instances~\cite{brandimarte1993routing}, and JMS-Bench models a semiconductor cluster-tool manufacturing scenario; detailed descriptions are provided in Sections \ref{sec:mkbench} and \ref{sec:jmsbench}, respectively.
}

\textbf{Metrics and Models.}
To ensure a fair and scalable comparison across instances of varying difficulty, we use relative performance metrics. Our primary metrics are:
\begin{enumerate}
    \item \textbf{Relative Percent Deviation (RPD)}, which normalizes performance against the best-known solution for each instance, calculated as:
    \begin{equation}
        \color{blue}
        \text{RPD}(\%) = \frac{C_{\max} - C_{\max}^{\star}}{C_{\max}} \times 100
    \end{equation}
    where \textcolor{blue}{$C_{\max}$} is the makespan from the evaluated method and \textcolor{blue}{$C_{\max}^{\star}$} is the best makespan found for that instance across all methods and runs in this study. Lower RPD is better.
    \item \textbf{Average Rank}, which measures the consistency of a method's performance. For each instance, all competing methods are ranked based on their makespan. A method's average rank is its mean rank across all instances. A lower average rank indicates a consistently superior method.
\end{enumerate}
To assess statistical significance, we use the non-parametric Wilcoxon signed-rank test~\cite{Janez2006Statistical}. We evaluate a range of LLMs, including GPT-4o~\cite{openai2024gpt4ocard}, DeepSeek-V3~\cite{deepseekai2025deepseekv3technicalreport}, and the Qwen3 series (8B, 14B, 32B)~\cite{yang2025qwen3technicalreport}. All models are prompted in a zero-shot manner with Chain-of-Thought reasoning~\cite{wei2022chain}.

\textbf{Evaluation Protocol.}
For initial diagnostic analyses aimed at capturing intrinsic model behaviors, we employ a multi-sample voting scheme: actions are decided by a majority vote from 5 independent generations at a high temperature ($T=0.8$)~\cite{wang2022self}. For the final comparative experiments against the LLM-Direct baseline, all LLM inferences use a low temperature ($T=0.2$) for efficiency and stability.

\textcolor{blue}{To ensure the reliability of our findings and mitigate the stochastic nature of LLM generation, we conduct three independent runs for each problem instance and use the resulting mean values for our primary analysis.
The statistical significance of the performance improvements is assessed using the non-parametric Wilcoxon signed-rank test.
It is important to note that the sample size for this statistical test is defined by the total number of problem instances in the benchmark suites instead of the number of independent runs per instance.
This methodology ensures that the statistical claims are based on a diverse population of scheduling scenarios, while the repeated runs for each case provide a more stable and accurate estimate of the expected performance for each individual instance.}

\begin{table}[htbp]
    \centering
    \small
    \setlength{\tabcolsep}{4pt}
    
    \caption{Statistical comparison between ReflecSched and the best-performing heuristic on each instance across all problem cases.}
    \begin{tabular}{lcc}
        \toprule
        \textbf{Instance Set} & \textbf{RPD (\%)} & \textbf{\textit{p}-value} \\
        \midrule
        Normal & 0.0620 & 0.8107 \\
        Small  & 0.9331 & 0.1240 \\
        \bottomrule
    \end{tabular}
    \label{tab:performance_comparison_wide}
\end{table}


\begin{table*}[htbp]
\centering
\color{blue} 
\caption{Comprehensive performance evaluation of ReflecSched variants against baseline algorithms across diverse benchmarks. Performance is quantified via Average Relative Percentage Deviation (RPD, \%) and Average Rank, where lower values signify optimal outcomes. Bold values denote the best performance in each row.}
\label{tab:overall_results_subtables}

\begin{subtable}{\textwidth}
\centering
\caption{General Performance Evaluation on GEN-Bench}
\label{subtab:genbench}
\resizebox{\textwidth}{!}{
\begin{tabular}{l ccccc | ccccccc}
\toprule
\multirow{2}{*}{\textbf{Metric}} & \multicolumn{5}{c|}{\textbf{Baseline Methods}} & \multicolumn{7}{c}{\textbf{ReflecSched}} \\
\cmidrule(lr){2-6} \cmidrule(lr){7-13}
& GP & HMPSAC & IDDQN & DAN & PPO-OC & Q3-8b & Q3-14b & Q3-32b & DS-v3 & DS-v3.2 & GPT-4o & GPT-5n \\
\midrule
Avg. RPD (\%) & 54.85 & 11.00 & 10.45 & 10.74 & 13.47 & 6.94 & \textbf{6.09} & 6.87 & 7.49 & 7.74 & 7.14 & 8.03 \\
Avg. Rank     & 11.42 & 6.68 & 6.58 & 6.47 & 8.05 & 5.34 & \textbf{4.39} & 5.45 & 5.58 & 5.21 & 5.34 & 5.32 \\
\bottomrule
\end{tabular}
}
\end{subtable}

\vspace{1.5em} 

\begin{subtable}{\textwidth}
\centering
\caption{Cross-Benchmark Stability on MK-Bench and JMS-Bench}
\label{subtab:mk_jms_bench}
\resizebox{\textwidth}{!}{
\begin{tabular}{ll ccccc | ccccc}
\toprule
\multirow{2}{*}{\textbf{Dataset}} & \multirow{2}{*}{\textbf{Metric}} & \multicolumn{5}{c|}{\textbf{Baseline Methods}} & \multicolumn{5}{c}{\textbf{ReflecSched}} \\
\cmidrule(lr){3-7} \cmidrule(lr){8-12}
& & GP & HMPSAC & IDDQN & DAN & PPO-OC & Q3-8b & Q3-14b & Q3-32b & DS-v3.2 & GPT-5n \\
\midrule
\multirow{2}{*}{MK-Bench} 
& Avg. RPD (\%) & 55.80 & 14.32 & 11.85 & 17.81 & 15.91 & 10.30 & 14.30 & 13.83 & \textbf{6.83} & 13.19 \\
& Avg. Rank     & 9.00  & 5.40  & 4.70  & 6.60  & 5.20  & 4.00  & 5.70  & 5.20  & \textbf{3.90} & 4.90 \\
\midrule
\multirow{2}{*}{JMS-Bench} 
& Avg. RPD (\%) & 46.80 & 9.72  & 8.01  & 15.56 & 10.94 & 6.47 & \textbf{6.18} & 7.74 & 8.50 & 10.16 \\
& Avg. Rank     & 9.70  & 5.00  & 4.70  & 7.10  & 5.70  & 4.20 & \textbf{4.00} & 4.10 & 4.40 & 5.30 \\
\bottomrule
\end{tabular}
}
\end{subtable}

\vspace{5pt}
\begin{flushleft}
\footnotesize \textit{Note: Q3 denotes the Qwen3 series~\cite{yang2025qwen3technicalreport}, DS denotes the DeepSeek series~\cite{deepseekai2025deepseekv3technicalreport, deepseekai2025deepseekv32pushingfrontieropen}, and GPT-5n refers to GPT-5 nano~\cite{singh2025openaigpt5card}. RPD values are expressed as percentages to highlight subtle performance variations between competitive models.}
\end{flushleft}
\end{table*}

\subsection{Comparative Performance Analysis}
\label{sec:main_results_and_analysis}

\textcolor{blue}{In this section, we conduct a detailed quantitative analysis of ReflecSched using the GEN-Bench dataset to evaluate its general performance across varying problem scales.}

\subsubsection{Comparison with LLM-Direct and Heuristics}
Our primary comparison is against the LLM-Direct baseline to investigate the effectiveness of our hierarchical reflection architecture. As shown in Figure \ref{fig:main_results_barchart}, ReflecSched demonstrates a statistically significant performance improvement over LLM-Direct across all tested configurations. It achieves a 71.35\% average Win Rate and an average RPD reduction of 2.755\%.

Furthermore, the framework is capable of outperforming any single heuristic. We also compared its performance to an oracle baseline constructed by selecting the best-performing heuristic for each instance individually. As shown in Table \ref{tab:performance_comparison_wide}, a two-sided Wilcoxon signed-rank test confirms no statistically significant difference between ReflecSched and this strong oracle baseline ($p > 0.05$), highlighting the framework's effectiveness.

\subsubsection{Comparison with Baseline Methods}
To benchmark our ReflecSched framework, we selected and adapted \textcolor{blue}{five recent and competitive baselines spanning evolutionary and reinforcement learning approaches to scheduling.
The baselines are taken from recent peer reviewed literature and are chosen to cover diverse modeling and optimization styles.
Specifically, we include GP, an evolutionary method that uses genetic programming to evolve priority functions over a predefined feature set, reducing reliance on manually designed dispatching rules~\cite{mei2016feature}.
We also include DAN, an attention based model that jointly attends to operations and machines to learn instance specific representations for scheduling~\cite{wang2023flexible}.}

\textcolor{blue}{For benchmarking, we evaluate three distinct reinforcement learning baselines with diverse architectural characteristics.
IDDQN is a Double DQN variant optimized for dispatching rule selection under stochastic machine breakdowns, employing a dueling network architecture and a modified experience replay scheme~\cite{wu2025dynamic}.
PPO-OC implements a Proximal Policy Optimization policy that maps production states to dispatching rules through a discrete action space~\cite{yuan2025deep}.
HMPSAC adopts a hierarchical multi-policy Soft Actor-Critic framework, where a high-level controller determines intermediate objective signals and low-level policies execute specific dispatching rules~\cite{ding2025data}.
Since our problem formulation involves dynamic event types that deviate from the original assumptions of these algorithms, we adapted their state representations and action interfaces to ensure compatibility with our event-driven model.
The comprehensive training protocols and hyperparameter configurations for all evaluated methods are detailed in Table \ref{tab:hyperparameters}.
To ensure empirical fairness and a rigorous comparison, we strictly adhered to the canonical configurations and optimal settings recommended in the original literature for each baseline algorithm.}

The performance of ReflecSched, instantiated with \textcolor{blue}{seven different LLM backends}, was compared against these baselines. The results, summarized in Table~\ref{tab:overall_results_subtables}, unequivocally demonstrate the superior performance of our framework. \textcolor{blue}{To facilitate a more granular assessment of per-instance performance and variability, we provide a comprehensive record of raw makespan results in Appendix G. This supplement includes individual data for the LLM variants, specifically Qwen3-8B, Qwen3-14B, Qwen3-32B~\cite{yang2025qwen3technicalreport}, DeepSeek-V3~\cite{deepseekai2025deepseekv3technicalreport}, DeepSeek-V3.2~\cite{deepseekai2025deepseekv32pushingfrontieropen}, GPT-4o~\cite{openai2024gpt4ocard}, and GPT-5 Nano~\cite{singh2025openaigpt5card}, alongside the five specialized comparative algorithms.}

\textcolor{blue}{As detailed in Table \ref{subtab:genbench}, all ReflecSched configurations demonstrate a substantial performance advantage over the traditional reinforcement learning and heuristic baselines on the GEN-Bench dataset. While the most competitive baseline method IDDQN achieves an average RPD of 10.45 percent, every variant of the ReflecSched framework maintains a significantly lower deviation with values ranging from 6.09 percent to 8.03 percent. Notably, the configuration utilizing the Qwen3 14b model delivers the most superior outcomes by attaining the minimum average RPD of 6.09 percent and the leading average rank of 4.39. Other specialized reinforcement learning approaches such as HMPSAC and DAN exhibit higher deviations of 11.00 percent and 10.74 percent respectively, which further underscores the efficacy of the hierarchical reflection mechanism in optimizing complex scheduling trajectories. The consistency of the proposed framework is further validated by the rank metrics where the ReflecSched variants occupy the top seven positions while the GP heuristic trails significantly with an average rank of 11.42. These results indicate that ReflecSched possesses a robust capacity for high-quality solution generation that exceeds the capabilities of models relying on direct generation or conventional training paradigms.}

In summary, the ReflecSched framework not only delivers superior overall performance but also demonstrates remarkable robustness. Unlike the baseline methods, whose effectiveness varies considerably between problem scales, ReflecSched consistently delivers top-tier results, proving its strong generalization and balanced performance.

{\color{blue}

\subsection{Evaluation on Established Benchmarks: MK-Bench}
\label{sec:mkbench}
To validate the cross-configuration generalizability of ReflecSched, the framework is benchmarked against MK-Bench, a suite derived from the established Brandimarte MK01--MK10 Flexible Job-Shop Scheduling instances~\cite{brandimarte1993routing}.
The integration of these recognized benchmarks ensures consistency with standardized scheduling research.
Given that the original Brandimarte instances are inherently static, we implement a systematic transformation protocol to extend them into dynamic environments.
This adaptation introduces stochastic event arrivals while preserving the intrinsic shop topology, specifically the job and machine counts, as well as the nominal processing durations.

The adaptation protocol is structured into three sequential phases.
First, the scheduling horizon for each instance is determined by deriving a theoretical workload lower bound, defined as the maximum of the longest individual job processing time and the average machine-wise workload.
This bound is subsequently scaled by a factor of 1.2 to define the simulation horizon, thereby providing sufficient temporal slack for dynamic adjustments.
Second, the three primary bottleneck machines are identified based on their average workload share, establishing a systematic basis for the introduction of targeted stochastic disruptions.
Third, four categories of dynamic events are injected using a deterministic random seed to ensure experimental reproducibility.
In these scenarios, 60\% of the jobs are initialized at $t=0$ to simulate initial work-in-progress (WIP), while subsequent tasks follow an exponential arrival distribution restricted to the first half of the horizon.
Furthermore, each resource is subject to a 50\% breakdown probability, with mean time to repair sampled uniformly between 1.0 and 4.0 time units.
Furthermore, the transformation incorporates stochastic job cancellations with a 30\% probability and the insertion of high-priority urgent jobs, which arrive at timestamps sampled uniformly from 25\% to 75\% of the simulation horizon.
These urgent tasks are strategically assigned to the previously identified bottleneck machines to intensify resource contention and evaluate the framework’s robustness under adversarial conditions.

As detailed in the MK-Bench results within Table \ref{subtab:mk_jms_bench}, the ReflecSched framework demonstrates high stability when applied to flexible job-shop scheduling problems. While the most effective DRL baseline IDDQN achieves an average RPD of 11.85 percent, the ReflecSched variants maintain a clear performance advantage within this benchmarking suite. Specifically, the variant utilizing DeepSeek v3.2 attains the most favorable outcomes with an average RPD of 6.83 percent and a leading average rank of 3.90. This level of performance indicates that the hierarchical reflection mechanism successfully identifies fundamental coordination patterns that extend beyond the specific distributions encountered during the initial prompts. This finding validates the efficacy of the framework in addressing the intensive resource competition and machine flexibility constraints inherent in the standardized MK01 through MK10 instances. The overall results suggest that the strategic insights derived from the reflection module facilitate a robust approach to managing diverse workshop topologies without the necessity for specialized retraining or architecture modification.

\subsection{Validation on Manufacturing Scenarios: JMS-Bench}
\label{sec:jmsbench}
To evaluate the industrial applicability of ReflecSched, we benchmark the framework on the JMS-Bench dataset, which is tailored to replicate the operational dynamics of a semiconductor cluster tool.
In modern wafer fabrication, the cluster tool constitutes the primary production unit, comprising a central transfer robot integrated with five heterogeneous processing modules, designated as PM1 through PM5, as shown in Figure \ref{fig:cluster_tools}.
Within this manufacturing context, each job represents a Front Opening Unified Pod (FOUP) carrying a batch of silicon wafers that must undergo a sequence of high-precision chemical and physical operations. The production flow is comprised of four distinct processing stages: Rough Etching, Finishing, Surface Grinding, and Metrology Inspection.

The scheduling complexity of this system is exacerbated by the functional overlap of chamber capabilities across these processing stages. Specifically, the Rough Etching stage utilizes $PM_1$ and $PM_2$, while the Finishing stage is shared between $PM_2$ and $PM_3$. Surface Grinding is distributed across $PM_3$ and $PM_4$, whereas Metrology Inspection is primarily handled by $PM_5$ with occasional overflow support from $PM_4$. This configuration dictates that $PM_2$, $PM_3$, and $PM_4$ serve as dual-functional modules that interlink consecutive processing phases. Consequently, the scheduler must manage acute resource contention, where the completion of a finishing task in $PM_3$ may preclude the initiation of a grinding operation for a concurrent job. Unlike traditional job shops, the central robot acts as a critical shared transport resource that must be tightly synchronized with these overlapping chamber cycles to prevent deadlocks, requiring the scheduler to incorporate look-ahead reasoning regarding module availability to sustain high throughput.

In this case study, we characterize three representative categories of dynamic disturbances inherent to semiconductor manufacturing: the insertion of high-priority ``hot lots'' necessitating immediate preemption of nominal production, stochastic chamber failures induced by equipment drift, and job cancellations arising from upstream yield fluctuations. These events necessitate non-myopic reasoning to re-allocate resources while preserving the integrity of the global production schedule. The Strategic Experience produced by ReflecSched is designed to encode such high level coordination patterns, for example by deciding when a shared resource such as $PM_4$ should be kept available for metrology to reduce queue buildup at the inspection stage during high arrival periods.

The comparative performance metrics for the semiconductor manufacturing environment are detailed in Subtable b of Table \ref{subtab:mk_jms_bench}. According to the recorded data, the ReflecSched variants demonstrate significant advantages in managing the complex operational constraints of cluster tools when compared to both meta-heuristic and reinforcement learning baselines. Among the baseline algorithms, IDDQN emerges as the most competitive approach with an average RPD of 8.01 percent and an average rank of 4.70. However, the ReflecSched configuration powered by the Qwen3 14b model surpasses all baselines by achieving a minimum average RPD of 6.18 percent and an optimal average rank of 4.00. Other variants such as Qwen3 8b and Qwen3 32b also maintain high solution quality with RPD values of 6.47 percent and 7.74 percent respectively. This performance trend indicates that the hierarchical reflection mechanism effectively internalizes the intricate module to stage mappings and transport synchronization requirements inherent in semiconductor fabrication. Furthermore, the proposed framework consistently outperforms specialized reinforcement learning architectures such as HMPSAC and PPO-OC which yield higher RPD values of 9.72 percent and 10.94 percent respectively. These findings suggest that the strategic insights codified within the reflection module provide a robust foundation for mitigating resource contention even under the severe stochastic disruptions characteristic of modern wafer processing units.
}

\subsection{Ablation and Sensitivity Analysis}
\label{sec:ablation_and_sensitivity}
All experiments in this section were conducted using Qwen3-8B as the base model.

\subsubsection{Impact of Hierarchical Reflection}
\begin{figure}[!htbp]
    \centering
    \includegraphics[width=0.5\textwidth]{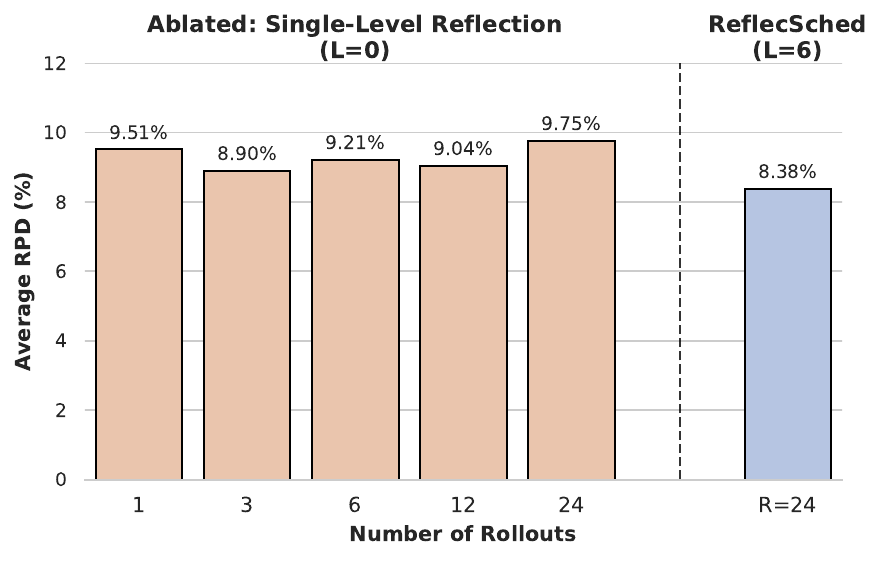}
    \caption{Ablation study on the hierarchical reflection mechanism. The chart compares the average RPD of the full ReflecSched framework with that of a single-level version. The performance of the single-level model is shown across a range of different values for the Number of Rollouts.}
    \label{fig:rollout_analysis}
\end{figure}

To isolate the contribution of our core architectural design, we conducted an ablation study comparing the full ReflecSched framework ($L=6, R=24$) against an ablated, single-level version ($L=0$). The results in Figure \ref{fig:rollout_analysis} confirm that the strategic foresight gained from multi-level reflection is a primary driver of the framework's effectiveness. More revealingly, simply increasing the search breadth for the single-level model fails to improve its performance, which plateaus and can even degrade. This suggests that a brute-force, ``flat'' search is ineffective without hierarchical guidance and confirms that the hierarchical reflection mechanism provides a qualitative benefit that simply increasing computation cannot replicate.

\begin{figure*}[!t]
    \centering
    \includegraphics[width=0.9\textwidth]{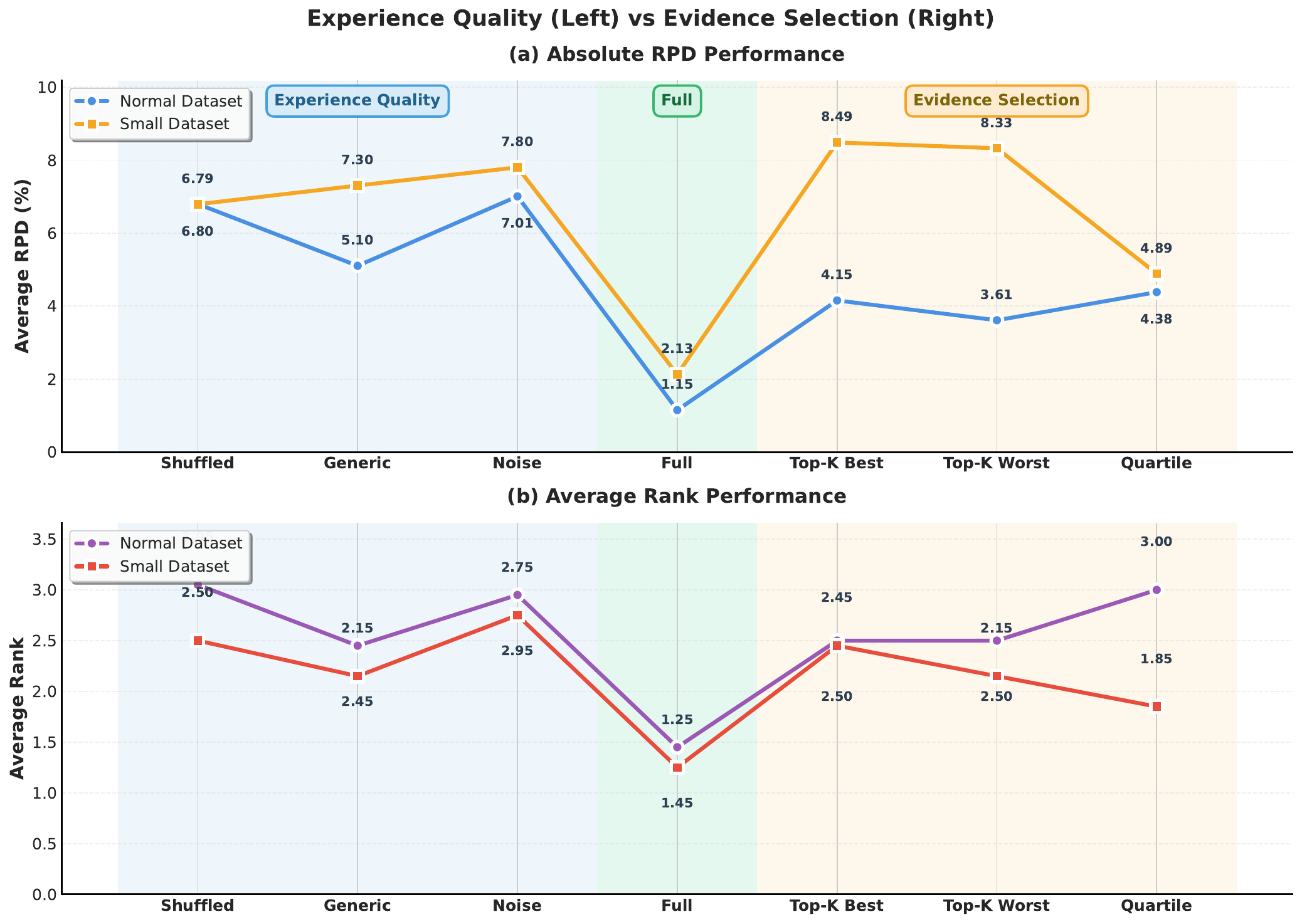}
    \caption{Ablation studies on Experience Quality (left) and Evidence Selection (right), using Qwen3-8B. (a) shows the Absolute RPD Performance, and (b) shows the Average Rank Performance. For both metrics, lower values are better.}
    \label{fig:choice_experence_merged}
\end{figure*}

\subsubsection{Impact of Experience Quality and Evidence Selection}
We conducted studies to assess the impact of experience quality and the strategy for selecting evidence. The results are presented in Figure \ref{fig:choice_experence_merged}.

\textbf{Experience Quality.} We compared four configurations: our complete Full Experience; Shuffled experience, which randomizes temporal order; Generic problem-agnostic experience; and Noise, where guidance from the best and worst historical experiences is swapped. As shown on the left of Figure \ref{fig:choice_experence_merged}, Full Experience significantly outperforms all alternatives. The poor performance with Shuffled and Generic experience highlights the importance of temporal coherence and domain-specific learning. The severe degradation with Noise demonstrates that misleading information is highly detrimental.

\textbf{Evidence Selection Strategy.} We investigated four strategies for selecting \textcolor{blue}{the trajectories that serve as the empirical basis for reflection}. \textcolor{blue}{Our Full approach provides contrastive insights by pairing the best and worst outcomes from the simulation pool. In contrast, Top-K Best focuses exclusively on successful trajectories to highlight optimal patterns, while Top-K Worst utilizes only the poorest-performing sequences to identify critical anti-patterns. The Quartile strategy employs a stratified sampling method to gather evidence from across the top, middle, and bottom performance tiers of the simulation results.} The results on the right of Figure~\ref{fig:choice_experence_merged} show that none of the alternatives match the comprehensive guidance of the Full experience. Interestingly, on the Normal Dataset, the Top-K Worst strategy outperforms Top-K Best, suggesting that learning from failures \textcolor{blue}{to identify actions that should be avoided} can be more valuable than learning exclusively from successes. This establishes that a comprehensive evidence base \textcolor{blue}{utilizing contrastive extremal outcomes} provides the most robust learning signal.

\subsubsection{Parameter Sensitivity Analysis under Budget Constraint}
\begin{figure}[!t]
    \centering
    \includegraphics[width=0.5\textwidth]{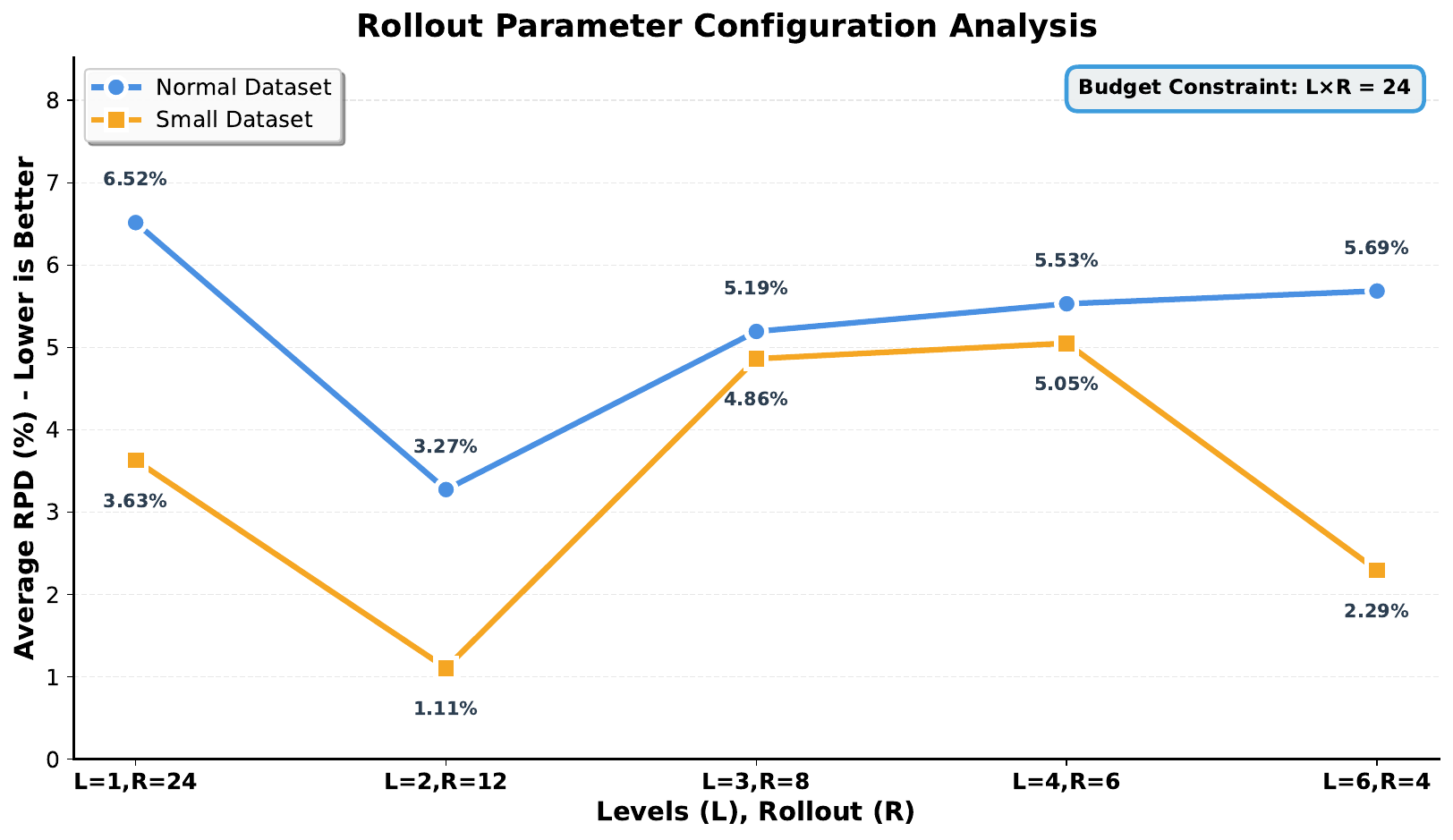}
    \caption{Analysis of Level (L) and Rollout (R) configurations under a fixed computational budget of $L \times R = 24$. The graph shows the Average RPD (\%) on both Normal and Small datasets for the Qwen3-8B model. Lower RPD indicates better performance.}
    \label{fig:level_rollout_analysis}
\end{figure}

To understand the trade-off between search depth (Levels, L) and search width (Rollouts, R), we conducted a sensitivity analysis under a fixed computational budget where $L \times R = 24$. As illustrated in Figure \ref{fig:level_rollout_analysis}, the results reveal a clear and consistent optimal balance.

\begin{figure*}[!t]
    \centering
    \includegraphics[width=0.98\textwidth]{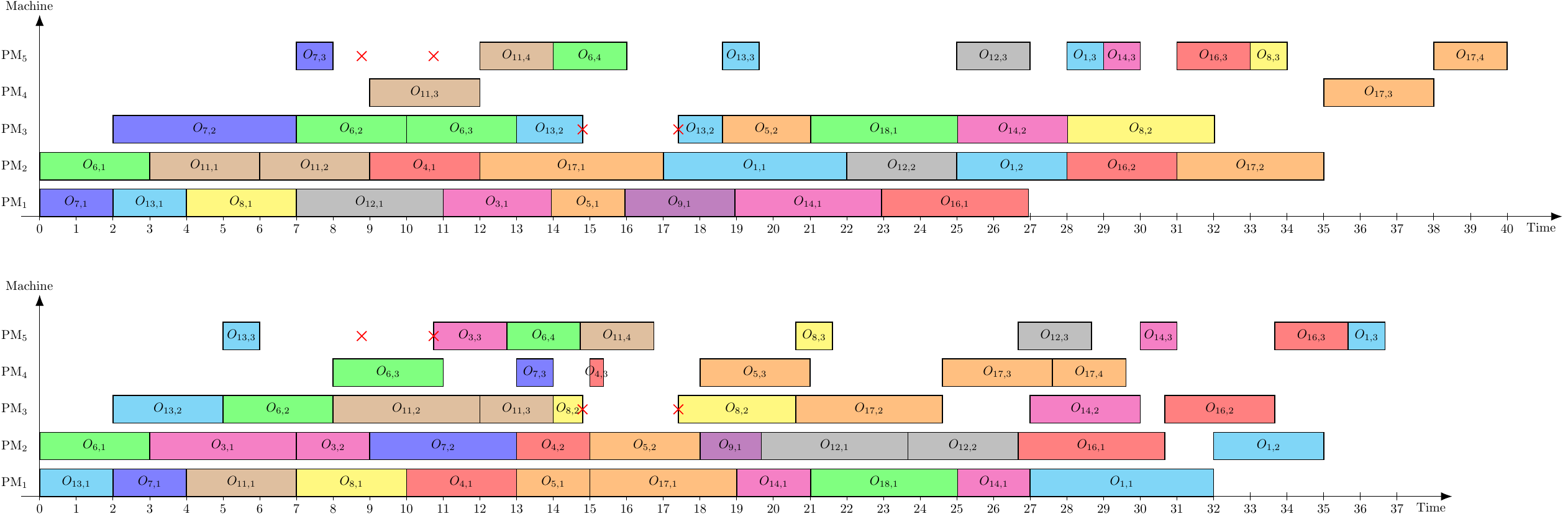}
    \caption{Comparative Gantt chart analysis of the semiconductor cluster tool scheduling instance. The top chart illustrates the schedule generated by the myopic baseline which achieves a makespan of 40.0. The bottom chart depicts the schedule produced by the ReflecSched framework yielding a superior makespan of 36.67. Red cross symbols denote the occurrence of dynamic machine breakdowns on modules PM3 and PM5. This comparison demonstrates how the proposed hierarchical reflection mechanism identifies global strategic trade-offs to mitigate the impact of stochastic disruptions.}
    \label{fig:cluster_gantt}
\end{figure*}

On both datasets, the configuration of \textbf{$L=2, R=12$} achieves the best performance. This strongly suggests that a moderate search depth combined with a substantial search width is the most effective strategy under a fixed budget. Both an extremely wide, shallow search ($L=1$) and an extremely deep, narrow search ($L=6$) lead to suboptimal performance. This analysis demonstrates that a balanced configuration that moderately favors search width over depth provides the most robust and effective performance across different problem scales.

{\color{blue}
\subsection{Case Study: Dynamic Scheduling under Stochastic Disruptions}
To provide a detailed elucidation of the decision-making mechanisms within ReflecSched, we conduct a comparative analysis of two distinct scheduling trajectories for a representative production instance. This scenario comprises five processing modules ($PM_1$ to $PM_5$) and 15 wafer batches with heterogeneous processing requirements. The primary objective is to evaluate the framework's resilience to stochastic machine failures and its ability to circumvent sub-optimal local states typical of reactive heuristics. As evidenced by the trajectory comparison, the ReflecSched-guided policy (Case A) yields an optimized makespan of 36.67, whereas the reactive baseline (Case B) results in an extended completion time of 40.0.

The dynamic scenario in this case study is characterized by two discrete equipment malfunctions: $PM_5$ undergoes a failure at $t = 8.78$ and is restored to operational status at $t = 10.74$, followed by a secondary failure of $PM_3$ at $t = 14.80$, with recovery at $t = 17.41$. Upon the initial failure of $PM_5$, the Strategic Experience codified by the reflection module designates Job 3 and Job 6 as critical-path tasks for the Metrology stage. Consequently, the scheduler preempts non-critical tasks, specifically Job 14 on $PM_1$ and Job 9 on $PM_2$, to expedite the resumption of Job 8 and Job 5 as resources become available. By tolerating localized delays for standard batches, ReflecSched prevents downstream delay propagation for Job 16 and Job 1, facilitating the conclusion of the final metrology operation for Job 1 at $t = 36.67$

In contrast, Case B exemplifies the inherent limitations of a reactive, greedy baseline. While this approach maximizes chamber utilization during the initial 10 time units, it lacks sufficient look-ahead capability to mitigate long-term resource contention. When the identical disruptions affect $PM_3$ and $PM_5$, the baseline model fails to reconfigure the global schedule optimally. Specifically, the myopic assignment of Job 13 and Job 11 to $PM_3$ early in the horizon induces severe congestion at the finishing stage. This inflexible allocation logic necessitates that Job 17 remains idle until the final phases of the simulation horizon before accessing its terminal processing chamber. Consequently, the absence of strategic coordination causes Job 17 to become the primary system bottleneck, yielding an extended makespan of 40.0. This comparison provides empirical evidence that the integration of hierarchical reflection enables the scheduler to anticipate downstream bottlenecks and sustain high throughput in the presence of intensive stochastic disturbances.

}

\subsection{Efficiency and Runtime Analysis}
\label{sec:efficiency_analysis}

\color{blue}
To evaluate the practical utility of ReflecSched, we conduct a multi-dimensional efficiency analysis encompassing wall-clock time, hardware overhead, and token consumption. All experiments were executed on a workstation featuring an Intel Xeon Platinum 8468V CPU and an NVIDIA H100 GPU (80 GB VRAM). LLM-based models were deployed via vLLM to ensure high-throughput inference~\cite{kwon2023efficient}.

In dynamic manufacturing environments, the selection of a scheduling framework requires a careful trade-off between deployment readiness and per-instance execution speed. While DRL models exhibit extremely low inference latency once weights are fully trained, their high initial training overhead serves as a significant bottleneck when factory configurations or optimization objectives shift. Conversely, ReflecSched utilizes a zero-shot architecture that enables immediate deployment without any prior training phase. 

\begin{figure}[H]
    \centering
    \includegraphics[width=0.5\textwidth]{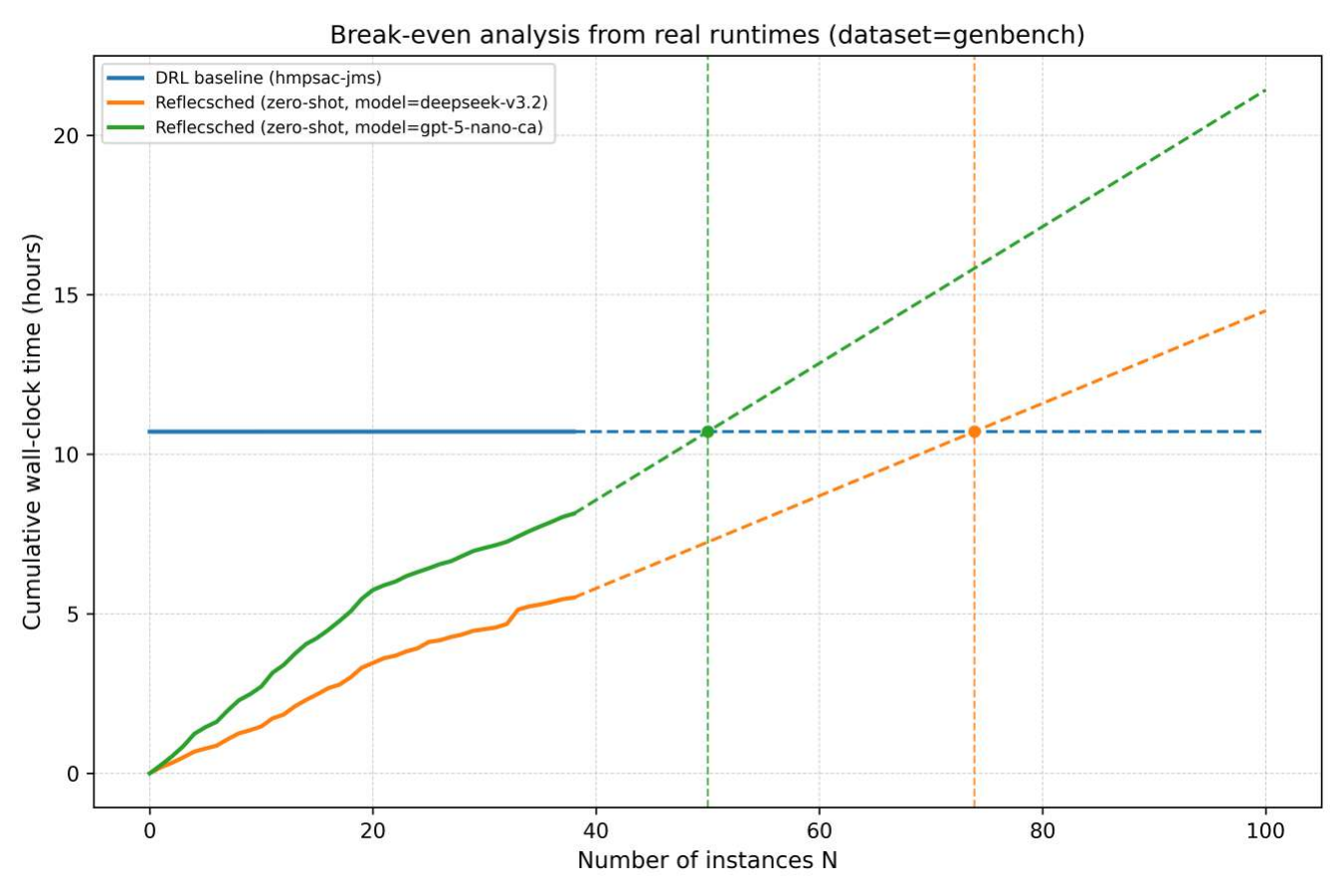}
    \caption{Break-even analysis of cumulative wall-clock time. Intercepts at $N=0$ represent initial training overhead. Solid and dashed lines denote recorded and extrapolated runtimes, respectively, with markers highlighting break-even points against the HMPSAC baseline.}
    \label{fig:run_time}
\end{figure}

We analyze this trade-off using Cumulative Wall-clock Time as defined by the following formulation:

\begin{equation}
    T_{cum}(N) = T_{train} + \sum_{i=1}^{N} T_{infer, i}
\end{equation}

where $T_{train}$ represents the offline training time and $T_{infer, i}$ denotes the wall-clock time required for each scheduling instance. As illustrated in Figure \ref{fig:run_time}, ReflecSched maintains a clear efficiency advantage in scenarios involving limited batch sizes or frequent environment re-configurations. Specifically, for the DeepSeek-V3.2 and GPT-5-nano backends, the cumulative time remains lower than that of the DRL baseline until the number of instances reaches a break-even point of approximately 74 and 50 respectively. 

However, we must acknowledge that due to the inherent computational complexity of large language model inference and iterative reflection, the marginal time cost per instance for ReflecSched is higher than that of specialized DRL models. In static environments characterized by massive, repetitive instance batches where $N$ significantly exceeds the break-even point, the initial training investment of DRL is effectively amortized, which allows its fast inference capability to become a dominant advantage. Therefore, ReflecSched is most suitably positioned as a robust solution for high-variability manufacturing systems where the cost of repeated model retraining would be prohibitive.

\begin{table}[htbp]
\centering
\caption{Average total token consumption (in thousands) per instance on the GEN-Bench dataset.}
\small
\setlength{\tabcolsep}{4pt}
\label{tab:token_consumption}
\resizebox{\columnwidth}{!}{%
    \begin{tabular}{lcccc}
    \toprule
    \multirow{2}{*}{\textbf{Model}} & \multicolumn{2}{c}{\textbf{Normal Scale}} & \multicolumn{2}{c}{\textbf{Small Scale}} \\
    \cmidrule(lr){2-3} \cmidrule(lr){4-5}
    & LLM-Direct & ReflecSched & LLM-Direct & ReflecSched \\
    \midrule
    GPT-4o & 137.4k & 115.4k & 21.3k & 45.2k \\
    DeepSeek-V3 & 144.4k & 126.3k & 23.7k & 49.7k \\
    Qwen3-32B & 153.3k & 144.6k & 23.6k & 52.5k \\
    Qwen3-14B & 146.4k & 120.9k & 21.7k & 46.6k \\
    Qwen3-8B & 148.8k & 112.8k & 22.8k & 45.5k \\
    \midrule
    \textbf{Average} & 146.1k & \textbf{124.0k} & \textbf{22.6k} & 47.9k \\
    \bottomrule
    \end{tabular}
}
\end{table}

The results in Table \ref{tab:token_consumption} reveal a trade-off dictated by problem scale. For Normal scale instances, which are longer and more complex, ReflecSched is significantly more token-efficient than the baseline. This is because the reflection module's generated \color{blue} Strategic Experience \color{black} allows subsequent decision-making prompts to be much leaner. Conversely, for Small scale instances, the upfront computational cost of the initial reflection outweighs these savings, resulting in higher token consumption. \color{blue} However, considering the significantly improved makespan ratio achieved by ReflecSched, \color{black} this highlights an intentional architectural trade-off: ReflecSched allocates a greater budget to its initial reflection phase, an investment that proves effective for larger, more complex problems by improving solution quality while simultaneously reducing the total \color{blue} computational and financial \color{black} tokens required.

\section{Conclusion}
\label{sec:conclusion}

This paper first provided an empirical diagnosis of applying LLMs to DFJSP, identifying three critical pitfalls: the long-context paradox, heuristic underutilization, and myopic decision-making.
To address these, our proposed ReflecSched framework recasts the LLM as a strategic analyst, distilling insights from multi-level simulations into a ``Strategic Experience'' to guide execution.
Experiments confirm that ReflecSched significantly outperforms direct LLM baselines, surpasses all traditional heuristics overall, and maintains performance on par with the instance-specific best heuristic across all problem instances.
We posit that the core principle of decoupling strategic reflection from execution offers a promising template for applying LLMs to a broader class of sequential decision-making problems.

\section*{Acknowledgements}
This work was partly supported by the National Natural Science Foundation of China.
{The code is available on \url{https://github.com/cls1277/ReflecSched} after being accepted.}

\appendix

\section{Implementation Details of LLM-Direct}
\label{app:llm_direct_details}

This appendix details the implementation of the LLM-Direct baseline scheduler.
The scheduler is implemented as a decision-making agent within an event-driven simulation framework.
It is invoked to make a scheduling decision at a discrete point in time, defined as a decision point.
A decision point occurs whenever the processing of system events (e.g., an operation completion or machine repair) leads to a state where at least one machine is idle while one or more operations are ready for processing.

\subsection{Event-Driven Simulation.}

The simulation framework is centered on an event loop that processes a series of time-stamped events (e.g., job arrivals, machine breakdowns, operation completions), which are managed in a priority queue.
The simulation progresses by advancing its internal clock to the timestamp of the next event in the queue and updating the system state according to the event's type.
A decision point is reached only after the event queue has been processed up to the current simulation time and the resulting state indicates that at least one operation is ready to be scheduled.
This event-driven architecture ensures that the LLM is invoked to make a decision only at these specific, necessary junctures, rather than on a continuous basis.

\subsection{State Representation and Prompt Construction}
At each decision point, a structured textual prompt is constructed to encode the current state of the workshop for the LLM.
This prompt serves as the sole source of information for the LLM's decision-making process.
The detailed structure, components, and the logic governing its construction are provided in \ref{app:promptdesign}.

\subsection{Core State Management}
The entire dynamic state of the simulation is encapsulated within a central state management class.
This class serves as the single source of truth for the workshop's status and is responsible for processing all dynamic events and calculating the set of currently feasible actions.
Its key functional components are as follows:

\paragraph{Initialization}
The state manager is initialized with the static problem definition, including job structures and machine processing times, along with a predefined queue of initial dynamic events and the starting simulation time.
    
\paragraph{State Variables}
Key variables maintained by the state manager include:
\begin{itemize}
    \item The availability time for each machine, corresponding to the parameter $r_k(t)$ in the formal model.
    \item The progress of each job, tracked by its last successfully completed operation.
    \item A set containing the identifiers of machines currently in a non-operational state.
    \item A set identifying jobs that have been designated with emergency status.
    \item A set of operations that have been interrupted (e.g., due to a machine breakdown) and are awaiting resumption. This corresponds to the set $\mathcal{O}_{int}$.
\end{itemize}
    
\paragraph{Event Handling}
This is the central method for state transition.
It processes a single event (e.g., a Job Arrival, Machine Breakdown, or Operation Completion) based on its type and timestamp.
This method is responsible for updating all relevant state variables.
For instance, a Machine Breakdown event adds the machine to the set of broken machines and moves any in-process operation to the set of interrupted operations.
A Job Emergency event dynamically loads the data for the new urgent job and adds its identifier to the set of emergency jobs.
    
\paragraph{Feasible Action Generation}
At each decision point, this method is invoked to determine the set of all feasible scheduling actions.
It iterates through all active jobs, verifies their arrival and precedence constraints, and identifies all operations that are ready to be scheduled.
For each such operation, it determines the subset of its candidate machines that are currently available (i.e., not broken and not busy).
The output is a set of feasible actions, which is used to generate the ``Ready Operations'' and ``Candidate Actions'' sections of the LLM prompt.
    
\paragraph{Action Execution}
Once the LLM selects an action, this method executes it.
It updates the availability time of the selected machine, records the assignment for performance tracking (e.g., for Gantt chart generation), and updates the machine's workload metrics.
Crucially, it generates a new Operation Completion event with a calculated end time and inserts it into the simulation's priority queue.

\section{Case Study: An Analysis of the Comprehension Limitations in Smaller LLMs}
\label{app:casestudy}
This appendix provides a qualitative analysis of the reasoning process exhibited by a smaller-scale model (Qwen3-4B) when functioning as the LLM-Direct scheduler~\cite{yang2025qwen3technicalreport}.
The subsequent examples, extracted directly from execution logs, are presented to illustrate specific failure modes.
These examples highlight the model's difficulties in consistently and accurately interpreting the complex, natural-language representation of the DFJSP state.
These case studies serve as empirical evidence for the central claim of our main paper: that the direct application of LLMs to this scheduling problem exhibits significant limitations.
This analysis, therefore, underscores the necessity for a more structured framework like ReflecSched.

\subsection{Over-reliance on a Simple Heuristic}

\paragraph{Prompt Context}
At the initial decision point in the simulation ($t=2.426$), the model is presented with a scenario involving a single ready operation, $O_{1,1}$, which has two candidate machines:
\begin{itemize}
    \item Machine 1, Processing Time: 1.78
    \item Machine 3, Processing Time: 2.66
\end{itemize}
The prompt also provides other strategic metrics for this operation.

\paragraph{Model's Reasoning and Behavior}
Across all five sampled responses, the model's reasoning was nearly identical.
A representative example is provided below, with the key reasoning statement highlighted in bold.
\par\vspace{\baselineskip}
\begin{lstlisting}[numbers=none, xleftmargin=\fill, xrightmargin=\fill]
Reasoning:
Since there are no emergency jobs, I need to choose the next operation...
The primary objective is to minimize the final makespan, which means [b]we should choose the operation that has the shortest processing time[/b]... 
the operation on Machine 1 has the shortest processing time of 1.78.
{"job": 1, "op": 1, "machine": 1}
\end{lstlisting}
\par\vspace{\baselineskip}

\paragraph{Analysis}
This case highlights the model's strong propensity to favor a simple, locally-optimal heuristic—the Shortest Processing Time (SPT) rule.
While the model correctly identifies the action corresponding to the SPT rule, its explicit reasoning reveals that it largely disregards other strategic information provided in the prompt, such as the remaining work and number of operations for the job.
The model's decision-making process reduces to a one-dimensional analysis, which serves as a concrete example of the ``myopic greed'' pitfall we identify in the main paper.
Although the model can parse individual data points, it fails to synthesize these multiple, potentially conflicting, pieces of information into a more robust strategic evaluation.

\subsection{Misinterpretation of State Descriptors}
\label{app:case_study:rem_work=0}

\paragraph{Prompt Context} 
At a later decision point ($t=7.466$), the model is presented with two ready operations, one of which is the final operation for its corresponding job.
The two candidate actions are:
\begin{itemize}
    \item Job 1, Operation 3 on Machine 1 (Time 2.99, \texttt{rem\_work=0.0}, \texttt{rem\_ops=0})
    \item Job 5, Operation 1 on Machine 1 (Time 1.85, \texttt{rem\_work=5.31}, \texttt{rem\_ops=3})
\end{itemize}

\paragraph{Model's Reasoning and Behavior}
Several of the model's responses exhibit a critical misinterpretation of the state descriptors.
A representative example follows:
\par\vspace{\baselineskip}
\begin{lstlisting}[numbers=none, xleftmargin=\fill, xrightmargin=\fill]
Reasoning:
Job 1 Operation 3 on Machine 1: This operation...[b]is already completed (rem_work = 0.0, rem_ops = 0)[/b]... 
Since Job 1 Operation 3 is already completed...the only operation that affects the makespan is Job 5 Operation 1.
Therefore, we should schedule Job 5 Operation 1...
{"job": 5, "op": 1, "machine": 1}
\end{lstlisting}
\par\vspace{\baselineskip}
\paragraph{Analysis}
This case provides a clear instance of a semantic interpretation failure.
The model's reasoning hinges on a critical error: it confuses a description of a potential future state with a historical fact.
Specifically, it incorrectly interprets the metrics \texttt{rem\_work=0.0} and \texttt{rem\_ops=0}---which describe the state of the job \textit{after} the proposed operation would be completed---as evidence that the operation is \textit{already} complete.
This logical fallacy leads the model to erroneously discard a feasible candidate action.
The example demonstrates the brittleness of the model's ability to map state descriptors to their correct temporal context, a foundational challenge for reliable decision-making.

\textcolor{blue}{
While redesigning state descriptors to distinguish between current status and post-action predictions might seem like a viable solution to prevent such conflations, the failure observed here is structural rather than merely linguistic.
The tendency of the model to misinterpret predicted metrics as historical facts reflects an inherent limitation in maintaining a consistent temporal context during one-step generation.
ReflecSched addresses this challenge by providing Strategic Experience to serve as a top-down interpretive framework.
This mechanism ensures that the model focuses on the strategic implications of a state, such as the potential to complete a critical job, instead of being misled by the semantic ambiguity of raw numerical descriptors.
}

\subsection{Inconsistent Adherence to Procedural Rules}

\paragraph{Prompt Context}
The prompt includes an explicit, high-priority rule for handling emergency jobs.
At timestamp $t=11.415$, the arrival of an emergency job creates a critical test of the model's ability to reliably apply this conditional instruction, particularly when presented with competing non-emergency options.
\begin{itemize}
    \item The ``Emergency Jobs'' list now contains ``[6]''.
    \item The ``Ready Operations'' list includes ``Job 6 Operation 1 [EMERGENCY]'' with three machine options.
    \item A non-emergency operation, ``Job 3 Operation 4'', is also available.
\end{itemize}

\paragraph{Model's Reasoning and Behavior}
The five sampled responses for this single decision point show a striking lack of consistency and, in some cases, outright confusion.
\begin{itemize}
    \item Correct Responses (3 out of 5): Three responses correctly identified the emergency job.
    \item Contradictory/Incorrect Responses (2 out of 5): Two responses exhibited confused reasoning. One's rationale is presented below.
\end{itemize}
\par\vspace{\baselineskip}
\begin{lstlisting}[numbers=none, xleftmargin=\fill, xrightmargin=\fill]
Reasoning:
First, I check for emergency jobs. Job 6 Operation 1 is marked as emergency... 
[b]However, the candidate actions list only includes non-emergency operations.[/b]
This is a contradiction...
\end{lstlisting}
\par\vspace{\baselineskip}
This assertion is factually incorrect, as the emergency job is explicitly listed in the ``Candidate Actions'' JSON block.
The model has thus failed to correctly parse the provided context, leading it to generate a spurious contradiction.

\paragraph{Analysis}
This case demonstrates a critical failure mode: the unreliable application of explicit, high-priority rules.
Even when presented with a straightforward directive, the model's reasoning process is shown to be inconsistent across multiple generation attempts.
The generation of a factually incorrect, self-contradictory rationale reveals a significant vulnerability.
While a majority voting mechanism might salvage the correct final decision in this instance, the flawed and unpredictable nature of the underlying reasoning process is exposed.
This unreliability is unacceptable in operational settings where consistent adherence to critical constraints is paramount.
This finding underscores that smaller-scale models cannot be reliably tasked with executing critical operational policies without a more robust, structured guiding framework.

\textcolor{blue}{The ReflecSched framework systematically mitigates the inconsistent rule adherence observed in this case study through its hierarchical architecture. While smaller models in the LLM-Direct baseline may generate contradictory rationales or overlook high-priority labels, ReflecSched implements a robust fallback and validation protocol. At each decision point, the candidate actions are filtered by the environment kernel to ensure they comply with technological constraints. If the model's output is found to be inconsistent with the set of feasible, high-priority actions, the framework utilizes a majority-voting scheme and a heuristic-driven tie-break mechanism to restore the correct decision path. Additionally, the periodic invocation of the reflection module upon the arrival of an emergency job forces the model to re-evaluate its strategy based on the most recent evidence, thereby reinforcing the high-priority rules in the Strategic Experience prompt.}

\section{Experimental Protocol for the PDR Application Analysis}
\label{app:pdr_experiment_details}

This appendix provides a detailed description of the experimental protocol used to assess the LLM-Direct baseline's ability to apply heuristics, evaluated on our diagnostic PDR-Bench dataset.

\subsection{Experimental Objective}
The primary objective of this experiment is to quantitatively assess the LLM's ability to apply externally provided heuristics, Priority Dispatching Rules (PDRs), to its decision-making process.
The PDR-Bench dataset is specifically designed for this purpose, as each instance has a single, empirically best-performing PDR.
The experiment tests the hypothesis that when the model is explicitly prompted with the best-performing rule for an instance, its performance should converge towards that of the rule itself.

\subsection{Experimental Conditions}
To test this hypothesis, we evaluated the LLM-Direct baseline under three distinct prompting conditions:
\begin{itemize}
    \item \textbf{LLM-Direct:} In this standard condition, the prompt contains no information about any PDRs.
    The model must rely solely on its pre-trained knowledge and the provided state information.
    \item \textbf{All-PDRs:} In this condition, the prompt is augmented with a list of all 24 PDR combinations.
    The model is instructed that it may use these rules for guidance but is not explicitly required to do so.
    \item \textbf{Single-Best-PDR:} In this condition, for each instance in PDR-Bench, the prompt is augmented with only the single, specific PDR known to yield the best makespan.
    This condition provides the model with the most direct and unambiguous guidance, thereby isolating its ability to follow a single, targeted instruction.
\end{itemize}

\subsection{Prompt Augmentation for Heuristics}
The PDRs are integrated into the prompt as a distinct, clearly demarcated section.
For each rule, the prompt provides its acronym (e.g., SPT), a full name, a precise selection criterion, and the strategic rationale for its use.
This structure is designed to give the model all the necessary information to understand and potentially apply the heuristic.
The following excerpt illustrates the format and instructional language used. 
\par\vspace{\baselineskip}
\begin{lstlisting}[numbers=none, xleftmargin=\fill, xrightmargin=\fill]
... (previous prompt sections: State, Ready Ops, etc.) ...

# Priority Scheduling Rules
Below are the ``Operations Priority Scheduling Rules'' and ``Machines Priority Scheduling Rules''. You can choose to use these rules for scheduling, but it is NOT required.

## Operations Priority Scheduling Rules
SPT (Shortest Processing Time):
    Criterion: Choose the candidate with the smallest `min_pt`.
    Rationale: Selecting the operation whose shortest possible processing time is minimal reduces average flow time and keeps machines busy with quick tasks first.

## Machines Priority Scheduling Rules
LIT (Least Idle Time):
    Criterion: Among candidate machines, choose the one with the smallest `free until time`.
    Rationale: Favors machines that free up next, keeping resources continuously occupied.

... (subsequent prompt sections: Candidate Actions, Instructions, etc.) ...
\end{lstlisting}
\par\vspace{\baselineskip}
\section{Benchmark Generation and Curation Methodology}
\label{app:dataset_curation}

The empirical evaluation in this study relies on two benchmark suites, GEN-Bench and PDR-Bench, which were generated for this work.
These datasets were not randomly selected; rather, they were systematically curated from a large, procedurally generated pool of DFJSP instances to serve specific analytical purposes.
This appendix details the formal methodology for their generation and subsequent curation.

\subsection{Instance Generation}
\label{subsec:instance_generation}

To create a diverse and challenging pool of problems, we developed a parameterized instance generator that allows for fine-grained control over both the static and dynamic characteristics of the DFJSP instances.
The key generation parameters are summarized in Table \ref{tab:generation_params}.

\begin{table}[htbp]
\centering
\caption{Parameter ranges for instance generation at Normal and Small scales.}
\label{tab:generation_params}
\resizebox{\columnwidth}{!}{%
\begin{tabular}{lcc}
\toprule
\textbf{Parameter} & \textbf{Normal Scale} & \textbf{Small Scale} \\
\midrule
\multicolumn{3}{l}{\textit{Static Characteristics}} \\
Number of Jobs & [15, 20] & [3, 5] \\
Operations per Job & [2, 4] & [2, 4] \\
Number of Machines & [3, 5] & [3, 5] \\
Candidate Machines per Op. & [1, 3] & [1, 3] \\
Processing Time Range & [1.0, 5.0] & [1.0, 3.0] \\
\midrule
\multicolumn{3}{l}{\textit{Dynamic Event Characteristics}} \\
Emergency Jobs per Instance & 1 & 1 \\
Machine Failure Probability & 0.5 & 0.3 \\
Job Cancellation Probability & 0.3 & 0.2 \\
\bottomrule
\end{tabular}
}
\end{table}

The generation process for each instance is as follows:
\begin{itemize}
    \item \textbf{Static Structure}: The generator first defines the static topology of the problem by sampling the number of jobs, machines, and operations per job from the uniform integer ranges specified in Table \ref{tab:generation_params}.
    For each operation, a set of candidate machines is randomly sampled.
    A base processing time is drawn from a uniform distribution, with an additional perturbation factor ($\pm 20\%$) applied to model machine-specific efficiency differences.
    \item \textbf{Dynamic Events}: A set of dynamic events is introduced to ensure a high degree of problem dynamism.
    Job arrival times are sampled from the first half of the simulation horizon to ensure significant system load during the early stages.
    Machine breakdown events are generated stochastically for each machine based on a specified failure probability.
    The duration of each repair is drawn from a uniform distribution $\mathcal{U}(1.0, 4.0)$.
    Similarly, job cancellations and the arrival of new, high-priority emergency jobs are introduced based on their respective probabilities.
    \item \textbf{Adaptive Horizon}: The total simulation horizon for each instance is not fixed but is calculated dynamically.
    It is based on a simple makespan lower bound estimate (total processing work of all initial jobs divided by the machine count), which is then multiplied by a buffer factor (1.2).
    This ensures that the horizon is sufficiently long to accommodate the completion of all tasks and dynamic events.
\end{itemize}
This generator was used to create an initial pool of 200 instances for each problem scale, which subsequently served as the input for our curation processes.

\subsection{PDR-Bench: The Diagnostic Heuristic Benchmark}
\label{subsec:pdr_bench}

\paragraph{Goal}
The primary objective of PDR-Bench is to generate a diagnostic dataset of problem instances where, for each instance, a single PDR demonstrably outperforms all others.
This design creates an unambiguous ground truth, enabling a clear, quantitative assessment of an LLM's ability to apply a specific, demonstrably superior heuristic when prompted.

\paragraph{Methodology}
The curation process begins with a large, diverse set of generated DFJSP instances.
For each instance $I$, we first simulate its scheduling process using every individual PDR from a predefined set of rules $\mathcal{R}$.
Let $M(I, r)$ denote the final makespan achieved on instance $I$ using rule $r \in \mathcal{R}$.
An instance $I$ is selected for inclusion in PDR-Bench if and only if there exists a single, uniquely dominant heuristic.
Formally, we first identify the best-performing rule $r^*$ for the instance:
\begin{equation}
    r^* = \arg\min_{r \in \mathcal{R}} M(I, r)
\end{equation}
The instance $I$ is then included in the benchmark if this minimum is unique.
That is, the makespan of the best-performing rule must be strictly better than that of any other rule in the set:
\begin{equation}
    \forall r' \in \mathcal{R}, r' \neq r^* \implies M(I, r') > M(I, r^*)
\end{equation}
This filtering process yields a collection of problem instances, each of which is constructed to have a single, empirically dominant heuristic.
This establishes a clear and objective ground truth for our subsequent analysis of the LLM's heuristic application capabilities.

\subsection{GEN-Bench: The Fair Evaluation Benchmark}
\label{subsec:gen_bench}

\paragraph{Goal}
The objective of GEN-Bench is to curate a general-purpose benchmark that is demonstrably fair and unbiased.
This is achieved by satisfying two competing criteria:
(1) each individual instance should be challenging and capable of discriminating between the performance of different rules (high discrimination), and
(2) the dataset as a whole should not exhibit a systemic bias towards any particular rule (high global balance).

\paragraph{Methodology}
The curation of GEN-Bench is framed as a multi-objective optimization problem, which we address using a greedy selection algorithm.
The process involves quantifying two key statistical properties: instance-level discrimination and subset-level global balance.

\paragraph{Instance-Level Discrimination}
For each candidate instance $I$, we measure its ability to discriminate between the performances of different rules.
Let $\mathcal{P}_I = \{M(I, r) \mid r \in \mathcal{R}\}$ be the set of makespan results for instance $I$ across all rules in the set $\mathcal{R}$.
We compute a composite Discrimination Score, $Disc(I)$, as a weighted sum of five normalized statistical dispersion metrics.
A higher $Disc(I)$ indicates that the instance is more effective at highlighting performance differences between scheduling strategies.
The components are defined as follows:

\subparagraph{1. Coefficient of Variation (CV).}
This metric provides a scale-invariant measure of dispersion.
Let $\mu(\mathcal{P}_I)$ and $\sigma(\mathcal{P}_I)$ be the mean and standard deviation of the makespans in $\mathcal{P}_I$.
\begin{equation}
    CV(\mathcal{P}_I) = \frac{\sigma(\mathcal{P}_I)}{\mu(\mathcal{P}_I)}
\end{equation}

\subparagraph{2. Relative Range.}
This captures the maximum performance spread relative to the mean.
\begin{equation}
    Range(\mathcal{P}_I) = \frac{\max(\mathcal{P}_I) - \min(\mathcal{P}_I)}{\mu(\mathcal{P}_I)}
\end{equation}

\subparagraph{3. Relative Interquartile Range (IQR).}
A robust measure of the spread of the central 50\% of the data~\cite{tukey1977exploratory}.
Let $Q1(\mathcal{P}_I)$ and $Q3(\mathcal{P}_I)$ be the first and third quartiles.
\begin{equation}
    IQR(\mathcal{P}_I) = \frac{Q3(\mathcal{P}_I) - Q1(\mathcal{P}_I)}{\text{median}(\mathcal{P}_I)}
\end{equation}

\subparagraph{4. Entropy of Performance Ranks.}
This quantifies the diversity of the performance ordering~\cite{jun2007normalized}.
Let $p_k$ be the proportion of rules achieving rank $k$.
The normalized Shannon entropy is:
\begin{equation}
    H(I) = -\frac{1}{\log(|\mathcal{R}|)} \sum_{k} p_k \log(p_k)
\end{equation}

\subparagraph{5. Average Pairwise Relative Difference.}
This assesses the average separation between all pairs of rule performances.
\begin{equation}
    d_{pair}(I) = \frac{1}{\binom{|\mathcal{R}|}{2}} \sum_{\{p_a, p_b\} \subseteq \mathcal{P}_I, a \neq b} \frac{|p_a - p_b|}{p_a + p_b}
\end{equation}

\subparagraph{Final Composite Score.}
The five metrics are normalized to a commensurable ``[0, 1]'' scale and then combined in a weighted sum to form the final discrimination score, $Disc(I)$.
We define a scaling function $\phi_c(x) = \min(x, c) / c$ to cap and normalize a value $x$ with a constant $c$.
The final score is calculated as:
\begin{equation}
\label{eq:disc_score_final}
\begin{aligned}
    Disc(I) = & \quad w_{cv} \cdot \phi_{2.0}(CV(\mathcal{P}_I)) \\
              & + w_{range} \cdot \phi_{3.0}(Range(\mathcal{P}_I)) \\
              & + w_{iqr} \cdot \phi_{1.5}(IQR(\mathcal{P}_I)) \\
              & + w_{H} \cdot H(I) \\
              & + w_{pair} \cdot \phi_{1.0}(d_{pair}(I))
\end{aligned}
\end{equation}
where the weights $(w_{cv}, w_{range}, w_{iqr}, w_{H}, w_{pair})$ are set to $(0.25, 0.20, 0.20, 0.15, 0.20)$. The entropy term $H(I)$ is already normalized.

\paragraph{Subset-Level Global Balance}
For any subset of instances $\mathcal{S}$, we measure its global balance to assess if the subset, as a whole, is biased towards any particular rule.
Let $\bar{M}(r, \mathcal{S})$ be the average performance of rule $r$ across all instances in $\mathcal{S}$.
Let $\bar{\mathcal{P}}_\mathcal{S} = \{\bar{M}(r, \mathcal{S}) \mid r \in \mathcal{R}\}$ be the set of these average performances.
The global balance is then defined as:
\begin{equation}
    Bal(\mathcal{S}) = 1 - \min(CV(\bar{\mathcal{P}}_\mathcal{S}), 1.0)
\end{equation}
A balance score close to 1 indicates that, on average, all PDRs perform similarly on the selected dataset, making it a fair and unbiased benchmark.

\paragraph{Greedy Selection Algorithm}
With these metrics defined, GEN-Bench is curated via a greedy algorithm.
Starting with an empty set $\mathcal{S}_0$, the algorithm iteratively adds the instance $I^*$ from the pool of remaining candidates that maximizes a weighted objective function.
At each step $k$, this function combines the average discrimination of the new set, its global balance, and the diversity of instance types:
\begin{equation}
\begin{aligned}
    I^* = \arg\max_{I \notin \mathcal{S}_k} \biggl[ 
    & w_{\text{disc}} \cdot \left( \frac{1}{|\mathcal{S}_k|+1} \sum_{I' \in \mathcal{S}_k \cup \{I\}} \text{Disc}(I') \right) \\
    +\, & w_{\text{bal}} \cdot \text{Bal}(\mathcal{S}_k \cup \{I\})\biggl]
\end{aligned}
\end{equation}

\section{Prompt Architecture and Design}
\label{app:promptdesign}

This appendix provides a detailed description of the architecture of the two core prompts used in this work: the \textbf{Decision Prompt} and the \textbf{Reflection Prompt}.
These prompts are programmatically constructed at runtime, assembling information from the simulation environment according to a predefined structure.

\subsection{The Decision Prompt}
The Decision Prompt is the primary interface for eliciting a scheduling decision from an LLM.
It is engineered to provide a structured, textual representation of the current shop-floor state, enabling the model to make a single, immediate scheduling choice.
This prompt's architecture is shared by both the LLM-Direct baseline and the Experience-Guided Decision-Making (EGDM) module within ReflecSched.
The key difference between the two applications is the inclusion of the ``Strategic Experience'' component, which is present only for the EGDM module.

The prompt is composed of several modular information blocks, assembled hierarchically to guide the model's reasoning.
\par\vspace{\baselineskip}
\begin{lstlisting}[numbers=none, xleftmargin=\fill, xrightmargin=\fill]
You are an expert scheduler in a dynamic factory. Your goal is to make smart, forward-looking decisions to keep the factory running smoothly and finish all jobs as early as possible (minimize makespan).

# Primary Objective
Choose the *single best* operation-machine pair to schedule right now. A good decision balances short-term gains with long-term risks.

# Key Information to Consider

1.  **Current Timestamp**: {snapshot['timestamp']}
2.  **Machine States**:
    - `status`: Is the machine available or broken?
    - `contention`: How many *future* operations need this machine? A high contention machine is a future bottleneck. **Avoid occupying a high-contention machine with a non-critical or flexible task.**
3.  **Ready Operations**:
    - `est`: When can this operation *actually* start?
    - `rem_work`: How much work is left for this job? High `rem_work` jobs might need to start sooner.
    - `flexibility`: How many machine options does this operation have? An operation with `flexibility: 1` is a critical constraint and may need to be prioritized to avoid it getting stuck.
    - `[EMERGENCY]`: These jobs MUST be scheduled before any non-emergency job.

{Machines States}

{Emergency Jobs}

{Strategic Experience}

{Ready Operations}

# Candidate Actions (only these are allowed)
```json
{actions_json}
```

# Task: Make a Decision

Think step-by-step. Your reasoning should balance these factors:

1.  **Urgency**: Handle `[EMERGENCY]` jobs first.
2.  **Constraints**: An operation with low `flexibility` (e.g., 1) is a constraint. Clearing it might unlock more options.
3.  **Bottlenecks**: Is the machine you are choosing a high-`contention` resource? If so, is this operation important enough to occupy it? Could a more flexible operation go to a less contended machine?
4.  **Flow**: Does scheduling a long-`rem_work` job now prevent it from becoming a problem later? Or is it better to clear a quick job to speed up the flow?

Based on your analysis, provide your final decision in JSON format.

```json
{{"job": <int>, "op": <int>, "machine": <int>}}
```
\end{lstlisting}
\par\vspace{\baselineskip}
\paragraph{Core Components of the Decision Prompt}
\begin{itemize}
    \item \textbf{Role and Objective:} The prompt begins by assigning the LLM the role of an ``expert scheduler'' and clearly states the primary objective (e.g., minimize makespan).
    
    \item \textbf{Glossary and Strategic Guide:} This section defines key operational metrics (e.g., machine contention, operation flexibility) to ensure consistent interpretation by the LLM and provides high-level advice on balancing competing objectives.
    
    \item \textbf{Machine States:} A summary of each machine's status, including its current availability (i.e., when it will be free) and a calculated contention level—a metric representing the number of future operations competing for that resource.
    
    \item \textbf{Emergency Jobs:} A list identifying any jobs currently designated with emergency status, which must be prioritized.

    \item \textbf{Ready Operations:} A catalog of all currently schedulable operations.
    For each operation, this section summarizes key attributes such as its earliest start time, shortest possible processing time, the remaining work in its parent job, and its scheduling flexibility (i.e., the number of candidate machines).
        
    \item \textbf{Candidate Actions:} A machine-readable, JSON-formatted list enumerating all feasible (job, operation, machine) scheduling actions.
    This serves as a strict output schema, constraining the LLM's final output to a valid and directly executable format.
    
    \item \textbf{Task Instructions:} A set of explicit instructions guiding the LLM's final reasoning process, prompting the model to balance the various informational components before committing to a final decision.
\end{itemize}

\paragraph{Conditional Component for EGDM}
\begin{itemize}
    \item \textbf{Strategic Experience:} (Included only for the EGDM module) This component inserts the core strategic guidance synthesized by the Hierarchical Reflection Module.
    This natural-language principle, enclosed in \texttt{<key\_insights>} tags, is the primary input for informing the LLM's non-myopic reasoning.
\end{itemize}

\subsection{The Reflection Prompt}
The Reflection Prompt serves a fundamentally different function from the Decision Prompt.
It does not request an immediate scheduling action.
Instead, it is engineered to instruct the LLM to act as a strategic analyst, tasked with synthesizing a new, more refined strategic principle based on a comparative analysis of simulated outcomes.
\par\vspace{\baselineskip}
\begin{lstlisting}[numbers=none, xleftmargin=\fill, xrightmargin=\fill]
You are a master Scheduling Strategist. Your mission is to analyze different simulated futures (rollout paths) to **REFINE and UPDATE** a generalizable strategic principle.

{The Existing Strategic Principle}

{The Originating Decision-Point State}

{Summarized Simulation Outcomes}

# Analysis and Synthesis Task
Your task is to integrate the **New Evidence** with the **Previous Experience**.

1.  **Analyze the Decision Paths**: Look at the "Best Path" and "Worst/Alternative Path". What is the key difference in their *initial decisions*? Why did starting with the better initial decision lead to a better overall makespan? Did it unblock a critical machine earlier? Did it prioritize a job with more work remaining? Look beyond just the first step and consider how it affected the sequence.

2.  **Synthesize a New Strategy**:
    *   **Evaluate New vs. Old**: How does the insight from your path analysis relate to the `Previous Experience`? Does it **CONFIRM** the old strategy (e.g., "The SPT rule worked again")? Does it **CONTRADICT** it (e.g., "SPT was bad here, prioritizing a job with more remaining work (MWKR) was better")? Or does it **ADD NUANCE** (e.g., "SPT is good, but only if it doesn't starve a critical machine that another job needs next").
    *   **Formulate the Updated Insight**: Create a single, robust strategic principle for the next decision. **Do not simply list old and new rules.** Synthesize them into one superior, more general rule.

# Output Requirements
Provide your analysis in the following XML-style tags. Be concise and focus on the final, actionable insights.

```xml
<comparison_summary>
(Your brief analysis comparing the decision paths, explaining the "why".)
</comparison_summary>

<key_insights>
(Your **NEW, REFINED, and SYNTHESIZED** strategic principle. This is the key output.)
</key_insights>
```
\end{lstlisting}
\par\vspace{\baselineskip}
\paragraph{Key Components of the Reflection Prompt}
\begin{itemize}
    \item \textbf{Role Assignment:} The LLM is instructed to act as a ``Scheduling Strategist.''
    Its stated task is to refine and update an existing scheduling policy, in contrast to the single-action task of the Decision Prompt.
    
    \item \textbf{The Existing Strategic Principle:} This block presents the current natural-language strategy (the ``Strategic Experience'') that was used to guide the initial heuristic-driven simulations, thereby providing the context for the refinement task.
    
    \item \textbf{The Originating Decision-Point State:} This component provides a full description of the system state at the moment the simulations were initiated.
    This information allows the LLM to ground its analysis of the simulation outcomes in the specific trade-offs and conditions that were present.
    
    \item \textbf{Summarized Simulation Outcomes:} This is the primary data for analysis.
    It presents a concise comparison of the best- and worst-performing simulated trajectories, highlighting their final makespans and, crucially, the divergent initial decision paths that produced these outcomes.
    
    \item \textbf{A Structured Reasoning Guide:} This section provides an explicit, two-stage set of instructions.
    First, it directs the LLM to analyze the causal relationship between the initial decision paths and the final outcomes.
    Second, it instructs the model to synthesize the insights from this analysis into a new, more robust strategic principle, potentially by refining or replacing the previous one.
    
    \item \textbf{A Strict Output Schema:} The prompt requires the LLM to structure its response using two specific XML-style tags: \texttt{<comparison\_summary>} for its analytical reasoning, and \texttt{<key\_insights>} for the final, synthesized strategic principle.
    This structured output is essential for the framework to programmatically extract and subsequently utilize the new ``Strategic Experience''.
\end{itemize}

{\color{blue}

\section{Data for Motivational Analysis Example}
\label{app:figure1_data}

This appendix provides the complete problem instance data corresponding to the motivating example discussed in Section \ref{sec:myopic_greed} and illustrated in Figure \ref{fig:problem_gantt}. Table \ref{tab:figure1_instance} details the candidate machines and corresponding processing times for each operation, derived directly from the schedule visualisations.

\begin{table}[htbp]
\color{blue}
\centering
\caption{Processing time data for the illustrative example in Figure \ref{fig:problem_gantt}. The pairs $(M_k, p)$ indicate that machine $k$ can process the operation in $p$ time units.}
\label{tab:figure1_instance}
\resizebox{\columnwidth}{!}{%
\begin{tabular}{cccc}
\toprule
\textbf{Job} & \textbf{Operation} & \textbf{Candidate Machines \& Processing Times} \\
\midrule
\multirow{3}{*}{Job 1} 
 & $O_{1,1}$ & $(M_2, 1.90), (M_3, 1.34)$ \\
 & $O_{1,2}$ & $(M_3, 1.98)$ \\
 & $O_{1,3}$ & $(M_2, 1.98)$ \\
\midrule
\multirow{4}{*}{Job 2} 
 & $O_{2,1}$ & $(M_1, 1.59), (M_2, 2.62)$ \\
 & $O_{2,2}$ & $(M_2, 1.71)$ \\
 & $O_{2,3}$ & $(M_1, 1.94), (M_2, 1.20)$ \\
 & $O_{2,4}$ & $(M_1, 0.96)$ \\
\midrule
\multirow{2}{*}{Job 3} 
 & $O_{3,1}$ & $(M_1, 1.08)$ \\
 & $O_{3,2}$ & $(M_3, 2.22)$ \\
\bottomrule
\end{tabular}
}
\end{table}
}

{\color{blue}

\section{Granular Performance Records for Per-Instance Comparative Analysis}
\label{app:full_result}

This appendix provides the complete record of raw performance results for every problem instance across the three benchmarking suites evaluated in this study, enabling a transparent comparison between ReflecSched and all baseline algorithms.

\begin{table*}[htbp]
\color{blue}
\centering
\caption{Comparative results of Makespan RPD (\%) across different problem scales on GEN-Bench. Relative Percentage Deviation (RPD) values are expressed as percentages. Bold values indicate the best performance for each instance across all evaluated methods.}
\label{tab:gen_bench_full_results}
\resizebox{\textwidth}{!}{
\begin{tabular}{l|ccccc|ccccccc}
\toprule
\multirow{2}{*}{\textbf{Instance}} & \multicolumn{5}{c|}{\textbf{Baseline Methods}} & \multicolumn{7}{c}{\textbf{ReflecSched (Proposed)}} \\
\cmidrule(lr){2-6} \cmidrule(lr){7-13}
& GP & HMPSAC & IDDQN & DAN & PPO-OC & Q3-8b & Q3-14b & Q3-32b & DS-V3 & DS-V3.2 & GPT-4o & GPT-5n \\
\midrule
Small-1  & 74.19 & 11.07 & 12.48 & 32.99 & 34.88 & 20.29 & 10.68 & 0.51 & 13.66 & \textbf{0.00} & 2.21 & \textbf{0.00} \\
Small-2  & 23.47 & 28.21 & 28.21 & 7.12 & 15.88 & 2.17 & 7.66 & 9.21 & 7.66 & \textbf{0.00} & 7.66 & 7.66 \\
Small-3  & 98.11 & 23.97 & 1.96 & 22.54 & 25.95 & \textbf{0.00} & 6.39 & 10.60 & 10.60 & 10.60 & 10.60 & 10.60 \\
Small-4  & 58.65 & 10.15 & 33.63 & \textbf{0.00} & 9.59 & 5.13 & 4.33 & 1.82 & 4.73 & 21.88 & 1.22 & 10.54 \\
Small-5  & 25.86 & 17.74 & 19.34 & 35.76 & 35.76 & 6.97 & \textbf{0.00} & 19.23 & 10.75 & 9.43 & 9.43 & 9.43 \\
Small-6  & 43.96 & 7.60 & 19.76 & 8.06 & 11.99 & 9.20 & \textbf{0.00} & 4.59 & 4.33 & 7.59 & 9.20 & 9.20 \\
Small-7  & 40.63 & \textbf{0.00} & 11.75 & 6.52 & 6.52 & 16.50 & 9.36 & 15.43 & 16.73 & 24.95 & 16.49 & 20.42 \\
Small-8  & 19.90 & 17.63 & 28.28 & 7.80 & 16.76 & 4.23 & 0.15 & \textbf{0.00} & 4.00 & 0.76 & 3.62 & 0.76 \\
Small-9  & 29.12 & 21.24 & 6.79 & 6.79 & 2.63 & 1.76 & 1.22 & 3.14 & 1.76 & \textbf{0.00} & 2.63 & 6.79 \\
Small-10 & 13.53 & \textbf{0.00} & \textbf{0.00} & \textbf{0.00} & 12.63 & 16.04 & 26.22 & 2.81 & 13.80 & 10.95 & 17.48 & 26.22 \\
Small-11 & 36.07 & 9.18 & 13.53 & \textbf{0.00} & 4.15 & 16.09 & 16.06 & 21.05 & 21.34 & 21.05 & 21.05 & 21.05 \\
Small-12 & 28.70 & 17.47 & 5.31 & 6.21 & \textbf{0.00} & 4.60 & 0.95 & 7.17 & 3.72 & 12.42 & 3.08 & 3.08 \\
Small-13 & 46.35 & \textbf{0.00} & 13.64 & 9.20 & 12.75 & 16.42 & 3.98 & 7.21 & 7.21 & 7.21 & 8.83 & 7.21 \\
Small-14 & 1.72 & 3.92 & 6.35 & \textbf{0.00} & 7.72 & 8.57 & 3.93 & 3.93 & 3.93 & 3.93 & 3.93 & 3.93 \\
Small-15 & 12.07 & \textbf{0.00} & 4.94 & 28.44 & 7.28 & 3.87 & 5.94 & 1.29 & \textbf{0.00} & \textbf{0.00} & 2.58 & \textbf{0.00} \\
Small-16 & 59.52 & \textbf{0.00} & 20.07 & 20.07 & 24.93 & 14.38 & 15.43 & 7.24 & 13.59 & 11.03 & 12.72 & 12.84 \\
Small-17 & 32.55 & 14.90 & 14.23 & 22.92 & 35.83 & 2.42 & \textbf{0.00} & 3.96 & 0.62 & 16.57 & 3.06 & 3.75 \\
Small-18 & 55.27 & 29.09 & \textbf{0.00} & 23.68 & 42.22 & 4.24 & 4.24 & 4.24 & 5.86 & 9.10 & 4.24 & 4.24 \\
\midrule
Normal-1 & 51.78 & 15.19 & 26.14 & 8.34 & 11.38 & 4.61 & \textbf{0.00} & 13.97 & 9.11 & 4.79 & 4.45 & 5.48 \\
Normal-2 & 60.73 & 16.79 & 5.13 & \textbf{0.00} & 6.77 & 4.14 & 7.29 & 4.37 & 3.35 & 4.73 & 1.98 & 16.72 \\
Normal-3 & 73.72 & 8.01 & \textbf{0.00} & 10.39 & 17.73 & 4.70 & 4.83 & 8.01 & 6.58 & 4.62 & 6.57 & 1.97 \\
Normal-4 & 46.75 & 3.89 & 2.36 & 4.63 & 1.38 & \textbf{0.00} & 5.25 & 7.07 & 7.82 & 6.18 & 8.77 & 6.18 \\
Normal-5 & 46.89 & 9.45 & \textbf{0.00} & 4.28 & 5.15 & 1.21 & 2.29 & 8.59 & 5.30 & 0.15 & 3.70 & 2.48 \\
Normal-6 & 55.35 & 1.26 & 3.14 & \textbf{0.00} & 5.02 & 0.91 & 0.87 & 4.38 & 1.25 & 14.83 & 3.87 & 16.16 \\
Normal-7 & 94.23 & 19.61 & 6.59 & 6.04 & 9.23 & 6.30 & 5.97 & 6.93 & 7.30 & \textbf{0.00} & 6.37 & 3.68 \\
Normal-8 & 95.64 & 13.98 & 7.58 & 4.58 & \textbf{0.00} & 3.62 & 3.58 & 4.84 & 0.62 & 2.14 & 4.27 & 8.80 \\
Normal-9 & 50.58 & 5.35 & 0.05 & 0.31 & 6.08 & 2.87 & \textbf{0.00} & 0.06 & 3.76 & 6.72 & 7.30 & 3.52 \\
Normal-10& 47.94 & 12.83 & 12.28 & 25.88 & 17.60 & 15.53 & 8.13 & 8.67 & 10.69 & 9.80 & 21.38 & \textbf{0.00} \\
Normal-11& 46.58 & 17.55 & 15.34 & 3.66 & 7.27 & 11.56 & 12.56 & 9.26 & 6.59 & 8.58 & 7.52 & \textbf{0.00} \\
Normal-12& 129.32 & 23.97 & 11.08 & 13.41 & 2.78 & 5.84 & 6.90 & 2.09 & 4.79 & 3.67 & \textbf{0.00} & 4.39 \\
Normal-13& 75.27 & \textbf{0.00} & 11.70 & 9.71 & 7.92 & 8.57 & 5.45 & 8.14 & 6.55 & 14.86 & 6.71 & 4.00 \\
Normal-14& 27.44 & 1.62 & 0.42 & 12.29 & 5.27 & 4.45 & \textbf{0.00} & 9.57 & 13.03 & 0.20 & 12.09 & 26.26 \\
Normal-15& 74.02 & 7.69 & 2.42 & 9.67 & 11.31 & 4.47 & 6.13 & 4.83 & 2.39 & 9.41 & 8.47 & \textbf{0.00} \\
Normal-16& 79.02 & 18.33 & 17.61 & \textbf{0.00} & 23.35 & 2.63 & 7.62 & 4.58 & 14.66 & 1.86 & 1.51 & 7.49 \\
Normal-17& 52.80 & \textbf{0.00} & 3.62 & 10.13 & 13.36 & 3.90 & 3.95 & 9.30 & 2.52 & 9.40 & 4.64 & 13.15 \\
Normal-18& 38.77 & 12.95 & 9.25 & 16.26 & 13.87 & 1.56 & 8.10 & 2.59 & 2.64 & 10.65 & \textbf{0.00} & 3.28 \\
Normal-19& 110.60 & 9.77 & 7.95 & \textbf{0.00} & 20.28 & 16.14 & 11.82 & 11.04 & 18.47 & 13.95 & 12.24 & 11.08 \\
Normal-20& 127.35 & 7.62 & 14.05 & 30.47 & 18.68 & 7.85 & 14.18 & 9.17 & 12.87 & \textbf{0.00} & 9.41 & 12.92 \\
\midrule
\rowcolor{gray!10} \textbf{Average} & 54.85 & 11.00 & 10.45 & 10.74 & 13.47 & 6.94 & \textbf{6.09} & 6.87 & 7.49 & 7.74 & 7.14 & 8.03 \\
\bottomrule
\end{tabular}
}
\end{table*}

\begin{table*}[htbp]
\color{blue}
\centering
\caption{Makespan RPD (\%) performance comparison on the established MK-Bench (Brandimarte-based). RPD values are expressed as percentages to underscore the performance variance. Bold values indicate the best performance for each instance across all evaluated methods.}
\label{tab:mk_bench_full_results}
\small
\setlength{\tabcolsep}{4.5pt} 
\begin{tabular}{l ccccc | ccccc}
\toprule
\multirow{2}{*}{\textbf{Instance}} & \multicolumn{5}{c}{\textbf{Baseline Methods}} & \multicolumn{5}{c}{\textbf{ReflecSched (Proposed)}} \\
\cmidrule(lr){2-6} \cmidrule(lr){7-11}
& GP & HMPSAC & IDDQN & DAN & PPO-OC & Q3-8b & Q3-14b & Q3-32b & DS-V3.2 & GPT-5n \\
\midrule
MK01 & 27.76 & 2.84 & 2.63 & 18.68 & \textbf{0.00} & 5.05 & 10.68 & 7.32 & 15.95 & 7.39 \\
MK02 & 16.93 & 12.67 & 9.40 & 9.46 & 28.15 & 17.32 & 30.32 & 34.32 & \textbf{0.00} & 12.67 \\
MK03 & 45.19 & 12.02 & 27.47 & 26.44 & 21.15 & 8.17 & 40.38 & 44.71 & \textbf{0.00} & 23.56 \\
MK04 & 123.64 & 3.34 & 10.16 & 4.02 & 3.76 & 1.74 & 7.11 & 0.70 & \textbf{0.00} & 10.16 \\
MK05 & 25.71 & 25.79 & 7.63 & 9.09 & 3.35 & 11.79 & \textbf{0.00} & 14.14 & 3.47 & 6.25 \\
MK06 & 147.97 & 22.42 & 17.22 & 28.05 & 29.73 & 12.49 & 15.66 & 9.33 & \textbf{0.00} & 9.33 \\
MK07 & 64.92 & 17.44 & 19.30 & 23.03 & 45.40 & 39.18 & \textbf{0.00} & 10.90 & 1.28 & 32.31 \\
MK08 & 7.72 & 6.10 & 6.25 & 23.46 & 7.29 & \textbf{0.00} & 9.74 & 9.43 & 20.13 & 5.96 \\
MK09 & 21.80 & 21.49 & 13.68 & 11.51 & 17.77 & 4.88 & 29.11 & \textbf{0.00} & 18.70 & 13.68 \\
MK10 & 76.36 & 19.06 & 4.71 & 24.35 & 2.49 & 2.36 & \textbf{0.00} & 7.46 & 8.80 & 10.61 \\
\midrule
\rowcolor{gray!10} \textbf{Average} & 55.80 & 14.32 & 11.85 & 17.81 & 15.91 & 10.30 & 14.30 & 13.83 & \textbf{6.83} & 13.19 \\
\bottomrule
\end{tabular}
\end{table*}

\begin{table*}[htbp]
\color{blue}
\centering
\caption{Makespan RPD (\%) performance comparison on JMS-Bench. RPD values are expressed as percentages to highlight subtle performance variations. Bold values indicate the best performance for each instance across all evaluated methods. All ReflecSched variants utilize the hierarchical reflection mechanism.}
\label{tab:jms_bench_full_results}
\small
\setlength{\tabcolsep}{4.5pt} 
\begin{tabular}{l ccccc | ccccc}
\toprule
\multirow{2}{*}{\textbf{Instance}} & \multicolumn{5}{c}{\textbf{Baseline Methods}} & \multicolumn{5}{c}{\textbf{ReflecSched (Proposed)}} \\
\cmidrule(lr){2-6} \cmidrule(lr){7-11}
& GP & HMPSAC & IDDQN & DAN & PPO-OC & Q3-8b & Q3-14b & Q3-32b & DS-V3.2 & GPT-5n \\
\midrule
instance\_01 & 25.58 & \textbf{0.00} & 6.47 & 20.51 & 3.67 & 9.30 & 14.88 & 3.67 & 14.88 & 2.55 \\
instance\_02 & 77.49 & 4.36 & 3.05 & 13.55 & 4.03 & 1.81 & 3.72 & \textbf{0.00} & 0.84 & 0.27 \\
instance\_03 & 30.69 & 0.70 & \textbf{0.00} & 3.53 & 7.72 & 11.66 & 5.31 & 9.95 & 18.27 & 10.50 \\
instance\_04 & 61.76 & 11.03 & 13.15 & 32.18 & 32.88 & 12.97 & \textbf{0.00} & 10.21 & 11.76 & 13.15 \\
instance\_05 & 62.60 & 19.04 & 5.01 & 20.33 & 11.79 & \textbf{0.00} & 9.35 & 11.79 & 8.88 & 22.90 \\
instance\_06 & 31.25 & 12.50 & 3.13 & 9.72 & \textbf{0.00} & 6.25 & 12.50 & 12.50 & 6.25 & 7.84 \\
instance\_07 & 7.20 & \textbf{0.00} & 1.70 & 18.20 & 15.50 & 9.95 & 2.75 & 3.38 & 3.93 & 0.80 \\
instance\_08 & 45.05 & 20.51 & 15.49 & \textbf{0.00} & 11.72 & 7.79 & 5.31 & 9.15 & 7.69 & 14.28 \\
instance\_09 & 78.09 & 10.69 & 13.38 & 18.73 & 12.90 & 4.96 & \textbf{0.00} & 10.63 & 5.29 & 10.36 \\
instance\_10 & 48.30 & 18.33 & 18.70 & 18.88 & 9.17 & \textbf{0.00} & 7.97 & 6.11 & 7.24 & 18.91 \\
\midrule
\rowcolor{gray!10} \textbf{Average} & 46.80 & 9.72 & 8.01 & 15.56 & 10.94 & 6.47 & \textbf{6.18} & 7.74 & 8.50 & 10.16 \\
\bottomrule
\end{tabular}
\end{table*}

\section{Hyperparameter Settings}

This appendix delineates the comprehensive hyperparameter configurations and architectural specifications utilized for the empirical evaluation of ReflecSched and the corresponding baselines.

\begin{table*}[htbp]
\centering
\color{blue}
\caption{Hyperparameter configurations for the baselines and the proposed ReflecSched.}
\label{tab:hyperparameters}
\small
\begin{tabular}{lll}
\toprule
\textbf{Algorithm} & \textbf{Parameter} & \textbf{Value} \\ \midrule
\multirow{5}{*}{\textbf{IDDQN}} & Training Episodes & 2,000 \\
 & Learning Rate / Discount Factor ($\gamma$) & 0.01 / 0.9 \\
 & Batch Size ($B$) / Buffer Size & $O_{sum}$ / $10 \times O_{sum}$ \\
 & Exploration / PER ($\alpha, \beta$) & $\epsilon$-greedy / 0.6, 0.4 \\
 & Target Network Update Interval & Every $O_{sum}$ steps \\ \midrule
\multirow{4}{*}{\textbf{PPO-OC / DAN}} & Learning Rate / Clipping Rate ($\epsilon$) & $3 \times 10^{-4}$ / 0.2 \\
 & Value ($c_v$) / Entropy ($c_e$) Loss Coef. & 0.5 / 0.01 \\
 & GAE ($\lambda$) & 0.95 (PPO), 0.98 (DAN) \\
 & PPO Update Epochs ($K$) & 10 (PPO), 4 (DAN) \\ \midrule
\multirow{4}{*}{\textbf{HMPSAC}} & Training Episodes & 1,000 \\
 & Learning Rate / Discount Factor ($\gamma$) & $3 \times 10^{-4}$ / 0.99 \\
 & Value ($c_v$) / Entropy ($c_e$) Loss Coef. & 0.5 / 0.01 \\
 & Max Decision Steps per Episode & 10,000 \\ \midrule
\multirow{11}{*}{\textbf{ReflecSched}} & Inference Temperature ($T_{inf}$) / Top-$p$ & 0.2 / 0.8 \\
 & Top-$k$ / Max Tokens & 20 / 8,192 \\
 & Reasoning Protocol & Reflective (Non-thinking) \\
 & Rollout Expansion Factor ($N_{roll}$) & 24 \\
 & Maximum Search Depth ($L_{max}$) & 6 \\
 & Search Iterations ($N_{iter}$) & 3 \\
 & Search Sampling Temperature ($T_{search}$) & 0.8 \\
 & Dynamic Gantt Information / Static Data & Enabled / Disabled \\ \bottomrule
\end{tabular}
\end{table*}

}

\bibliographystyle{elsarticle-num}
\bibliography{ref}

@article{pan2017scheduling,
  title={Scheduling cluster tools in semiconductor manufacturing: Recent advances and challenges},
  author={Pan, ChunRong and Zhou, MengChu and Qiao, Yan and Wu, NaiQi},
  journal={IEEE transactions on automation science and engineering},
  volume={15},
  number={2},
  pages={586--601},
  year={2017},
  publisher={IEEE}
}

@inproceedings{kwon2023efficient,
  title={Efficient Memory Management for Large Language Model Serving with PagedAttention},
  author={Woosuk Kwon and Zhuohan Li and Siyuan Zhuang and Ying Sheng and Lianmin Zheng and Cody Hao Yu and Joseph E. Gonzalez and Hao Zhang and Ion Stoica},
  booktitle={Proceedings of the ACM SIGOPS 29th Symposium on Operating Systems Principles},
  year={2023}
}

@article{huang2025leveraging,
  title={Leveraging large language models for efficient scheduling in Human--Robot collaborative flexible manufacturing systems},
  author={Huang, Jin and Teng, Yue and Liu, Qihao and Gao, Liang and Li, Xinyu and Zhang, Chunjiang and Xu, Guoqing},
  journal={npj Advanced Manufacturing},
  volume={2},
  number={1},
  pages={47},
  year={2025},
  publisher={Nature Publishing Group UK London}
}

@article{immordino2025explainable,
  title={Explainable AI for reinforcement learning based dynamic scheduling solutions in semiconductor manufacturing: A. Immordino et al.},
  author={Immordino, Alessandro and St{\"o}ckermann, Patrick and Hayen, Niels and Altenm{\"u}ller, Thomas and Susto, Gian Antonio and Gebser, Martin and Schekotihin, Konstantin and Seidel, Georg},
  journal={Journal of Intelligent Manufacturing},
  pages={1--17},
  year={2025},
  publisher={Springer}
}

@inproceedings{liu2002aps,
  author       = {William Liu and
                  Tay Jin Chua and
                  J. Larn and
                  Feng{-}Yu Wang and
                  Tian Xiang Cai and
                  Xiao{-}Feng Yin},
  title        = {APS, {ERP} and {MES} systems integration for semiconductor backend
                  assembly},
  booktitle    = {Seventh International Conference on Control, Automation, Robotics
                  and Vision, {ICARCV} 2002, Singapore, 2-5 December 2002, Proceedings},
  pages        = {1403--1408},
  publisher    = {{IEEE}},
  year         = {2002},
  url          = {https://doi.org/10.1109/ICARCV.2002.1234978},
  doi          = {10.1109/ICARCV.2002.1234978},
  bibsource    = {dblp computer science bibliography, https://dblp.org}
}

@article{brandimarte1993routing,
  title={Routing and scheduling in a flexible job shop by tabu search},
  author={Brandimarte, Paolo},
  journal={Annals of Operations research},
  volume={41},
  number={3},
  pages={157--183},
  year={1993},
  publisher={Springer}
}

@article{ahn2024novel,
  title={A novel mixed integer programming model with precedence relation-based decision variables for non-cyclic scheduling of cluster tools},
  author={Ahn, Jeongsun and Kim, Hyun-Jung},
  journal={IEEE Transactions on Automation Science and Engineering},
  volume={22},
  pages={2893--2908},
  year={2024},
  publisher={IEEE}
}

@misc{singh2025openaigpt5card,
      title={OpenAI GPT-5 System Card}, 
      author={Aaditya Singh and Adam Fry and Adam Perelman and Adam Tart and Adi Ganesh and Ahmed El-Kishky and Aidan McLaughlin and Aiden Low and AJ Ostrow and Akhila Ananthram  and others},
      year={2025},
      eprint={2601.03267},
      archivePrefix={arXiv},
      primaryClass={cs.CL},
      url={https://arxiv.org/abs/2601.03267}, 
}

@misc{deepseekai2025deepseekv32pushingfrontieropen,
      title={DeepSeek-V3.2: Pushing the Frontier of Open Large Language Models}, 
      author={DeepSeek-AI and Aixin Liu and Aoxue Mei and Bangcai Lin and Bing Xue and Bingxuan Wang and Bingzheng Xu and Bochao Wu and Bowei Zhang and Chaofan Lin and others},
      year={2025},
      eprint={2512.02556},
      archivePrefix={arXiv},
      primaryClass={cs.CL},
      url={https://arxiv.org/abs/2512.02556}, 
}

@article{xiong2020reducing,
  title={Reducing wafer delay time by robot idle time regulation for single-arm cluster tools},
  author={Xiong, WenQing and Pan, ChunRong and Qiao, Yan and Wu, NaiQi and Chen, MingXin and Hsieh, PinHui},
  journal={IEEE Transactions on Automation Science and Engineering},
  volume={18},
  number={4},
  pages={1653--1667},
  year={2020},
  publisher={IEEE}
}

@article{zhang2024dynamic,
  title={Dynamic flexible job-shop scheduling by multi-agent reinforcement learning with reward-shaping},
  author={Zhang, Lixiang and Yan, Yan and Yang, Chen and Hu, Yaoguang},
  journal={Advanced Engineering Informatics},
  volume={62},
  pages={102872},
  year={2024},
  publisher={Elsevier}
}

@article{wang2022multi,
  title={Multi-objective reinforcement learning framework for dynamic flexible job shop scheduling problem with uncertain events},
  author={Wang, Hao and Cheng, Junfu and Liu, Chang and Zhang, Yuanyuan and Hu, Shunfang and Chen, Liangyin},
  journal={Applied Soft Computing},
  volume={131},
  pages={109717},
  year={2022},
  publisher={Elsevier}
}

@article{liu2024dynamic,
  title={Dynamic job-shop scheduling using graph reinforcement learning with auxiliary strategy},
  author={Liu, Zhenyu and Mao, Haoyang and Sa, Guodong and Liu, Hui and Tan, Jianrong},
  journal={Journal of Manufacturing Systems},
  volume={73},
  pages={1--18},
  year={2024},
  publisher={Elsevier}
}

@article{ding2025data,
  title={Data-driven hierarchical multi-policy deep reinforcement learning framework for multi-objective multiplicity dynamic flexible job shop scheduling},
  author={Ding, Linshan and Guan, Zailin and Luo, Dan and Yue, Lei},
  journal={Journal of Manufacturing Systems},
  volume={80},
  pages={536--562},
  year={2025},
  publisher={Elsevier}
}

@article{li2025categorized,
  title={Categorized attention based hierarchical-agents reinforcement learning for multi-objective dynamic job shop scheduling problem with machine deterioration},
  author={Li, Yibing and Liang, Xueci and Guo, Jun and Li, Xixing and Wang, Lei and Du, Baigang},
  journal={Applied Soft Computing},
  volume={175},
  pages={113032},
  year={2025},
  publisher={Elsevier}
}

@article{chen2025optimizing,
  title={Optimizing dynamic flexible job shop scheduling using an evolutionary multi-task optimization framework and genetic programming},
  author={Chen, Xiaolong and Li, Junqing and Wang, Zunxun and Chen, Qingda and Gao, Kaizhou and Pan, Quanke},
  journal={IEEE Transactions on Evolutionary Computation},
  year={2025},
  publisher={IEEE}
}

@article{wu2025dynamic,
  title={Dynamic scheduling for flexible job shop under machine breakdown using improved Double Deep Q-network},
  author={Wu, Rui and Zheng, Jianxin and Li, Xixing and Tang, Hongtao and Wang, Xi Vincent and Li, Yibing},
  journal={Expert Systems with Applications},
  volume={288},
  pages={128280},
  year={2025},
  publisher={Elsevier}
}

@article{shi2025deep,
title = {A deep reinforcement learning method based on Hindsight experience replay for multi-objective dynamic job-shop scheduling problem},
journal = {Expert Systems with Applications},
volume = {284},
pages = {127989},
year = {2025},
issn = {0957-4174},
doi = {https://doi.org/10.1016/j.eswa.2025.127989},
url = {https://www.sciencedirect.com/science/article/pii/S0957417425016100},
author = {Zhiyuan Shi and Jinghua Si and Jian Zhang and Zhi Pang and Haojie Chen and Guofu Ding}
}

@article{yuan2025deep,
  title={Deep reinforcement learning based proximal policy optimization algorithm for dynamic job shop scheduling},
  author={Yuan, Minghai and Yu, Qi and Zhang, Lizhi and Lu, Songwei and Li, Zichen and Pei, Fengque},
  journal={Computers \& Operations Research},
  pages={107149},
  year={2025},
  publisher={Elsevier}
}

@article{yu2026adeepreinforcement,
title = {A deep reinforcement learning approach for dynamic job-shop scheduling problem considering time variable and new job arrivals},
journal = {Computers \& Operations Research},
volume = {185},
pages = {107263},
year = {2026},
issn = {0305-0548},
doi = {https://doi.org/10.1016/j.cor.2025.107263},
url = {https://www.sciencedirect.com/science/article/pii/S0305054825002928},
author = {Haoyang Yu and Wenbin Gu and Na Tang and Zhenyang Guo}
}

@article{zhang2026novel,
  title={A novel deep reinforcement learning framework based on digital twins for dynamic job shop scheduling problems},
  author={Zhang, Wenquan and Peng, Zhaoxian and Zhao, Fei and Feng, Bo and Mei, Xuesong},
  journal={Expert Systems with Applications},
  volume={296},
  pages={128708},
  year={2026},
  publisher={Elsevier}
}

@inproceedings{mei2016feature,
  title={Feature selection in evolving job shop dispatching rules with genetic programming},
  author={Mei, Yi and Zhang, Mengjie and Nyugen, Su},
  booktitle={Proceedings of the Genetic and Evolutionary Computation Conference 2016},
  pages={365--372},
  year={2016}
}

@article{wang2023flexible,
  title={Flexible job shop scheduling via dual attention network-based reinforcement learning},
  author={Wang, Runqing and Wang, Gang and Sun, Jian and Deng, Fang and Chen, Jie},
  journal={IEEE Transactions on Neural Networks and Learning Systems},
  volume={35},
  number={3},
  pages={3091--3102},
  year={2023},
  publisher={IEEE}
}

@article{cao2023inverse,
  title={Inverse model and adaptive neighborhood search based cooperative optimizer for energy-efficient distributed flexible job shop scheduling},
  author={Cao, Shijie and Li, Rui and Gong, Wenyin and Lu, Chao},
  journal={Swarm and Evolutionary Computation},
  volume={83},
  pages={101419},
  year={2023},
  publisher={Elsevier}
}

@article{yang2025qwen3technicalreport,
  title={Qwen3 technical report},
  author={Yang, An and Li, Anfeng and Yang, Baosong and Zhang, Beichen and Hui, Binyuan and Zheng, Bo and Yu, Bowen and Gao, Chang and Huang, Chengen and Lv, Chenxu and others},
  journal={arXiv preprint arXiv:2505.09388},
  year={2025}
}

@book{tukey1977exploratory,
  title={Exploratory data analysis},
  author={Tukey, John Wilder and others},
  volume={2},
  year={1977},
  publisher={Springer}
}

@article{jun2007normalized,
  title={Normalized entropy of rank distribution: a novel measure of heterogeneity of complex networks},
  author={Jun, Wu and Yue-Jin, Tan and Hong-Zhong, Deng and Da-Zhi, Zhu},
  journal={Chinese Physics},
  volume={16},
  number={6},
  pages={1576},
  year={2007},
  publisher={IOP Publishing}
}

@article{ren2024dynamic,
  title={Dynamic scheduling for flexible job shop based on MachineRank algorithm and reinforcement learning},
  author={Ren, Fujie and Liu, Haibin},
  journal={Scientific Reports},
  volume={14},
  number={1},
  pages={29741},
  year={2024},
  publisher={Nature Publishing Group UK London}
}

@article{baykasoglu2020greedy,
title = {Greedy randomized adaptive search for dynamic flexible job-shop scheduling},
journal = {Journal of Manufacturing Systems},
volume = {56},
pages = {425-451},
year = {2020},
issn = {0278-6125},
doi = {https://doi.org/10.1016/j.jmsy.2020.06.005},
url = {https://www.sciencedirect.com/science/article/pii/S0278612520300959},
author = {Adil Baykasoğlu and Fatma S. Madenoğlu and Alper Hamzadayı}
}

@inproceedings{hu2024hiagenthierarchicalworkingmemory,
  title={Hiagent: Hierarchical working memory management for solving long-horizon agent tasks with large language model},
  author={Hu, Mengkang and Chen, Tianxing and Chen, Qiguang and Mu, Yao and Shao, Wenqi and Luo, Ping},
  booktitle={Proceedings of the 63rd Annual Meeting of the Association for Computational Linguistics (Volume 1: Long Papers)},
  pages={32779--32798},
  year={2025}
}

@misc{packer2024memgptllmsoperatingsystems,
      title={MemGPT: Towards LLMs as Operating Systems}, 
      author={Charles Packer and Sarah Wooders and Kevin Lin and Vivian Fang and Shishir G. Patil and Ion Stoica and Joseph E. Gonzalez},
      year={2024},
      eprint={2310.08560},
      archivePrefix={arXiv},
      primaryClass={cs.AI},
      url={https://arxiv.org/abs/2310.08560}, 
}

@misc{sun2025hierarchicalmemoryhighefficiencylongterm,
      title={Hierarchical Memory for High-Efficiency Long-Term Reasoning in LLM Agents}, 
      author={Haoran Sun and Shaoning Zeng},
      year={2025},
      eprint={2507.22925},
      archivePrefix={arXiv},
      primaryClass={cs.CL},
      url={https://arxiv.org/abs/2507.22925}, 
}

@inproceedings{zhong2023memorybankenhancinglargelanguage,
  title={Memorybank: Enhancing large language models with long-term memory},
  author={Zhong, Wanjun and Guo, Lianghong and Gao, Qiqi and Ye, He and Wang, Yanlin},
  booktitle={Proceedings of the AAAI Conference on Artificial Intelligence},
  volume={38},
  number={17},
  pages={19724--19731},
  year={2024}
}

@article{shinn2023reflexion,
  title={Reflexion: Language agents with verbal reinforcement learning},
  author={Shinn, Noah and Cassano, Federico and Gopinath, Ashwin and Narasimhan, Karthik and Yao, Shunyu},
  journal={Advances in Neural Information Processing Systems},
  volume={36},
  pages={8634--8652},
  year={2023}
}

@article{yao2023tree,
  title={Tree of thoughts: Deliberate problem solving with large language models},
  author={Yao, Shunyu and Yu, Dian and Zhao, Jeffrey and Shafran, Izhak and Griffiths, Tom and Cao, Yuan and Narasimhan, Karthik},
  journal={Advances in neural information processing systems},
  volume={36},
  pages={11809--11822},
  year={2023}
}

@inproceedings{yao2023react,
  title={React: Synergizing reasoning and acting in language models},
  author={Yao, Shunyu and Zhao, Jeffrey and Yu, Dian and Du, Nan and Shafran, Izhak and Narasimhan, Karthik R and Cao, Yuan},
  booktitle={The eleventh international conference on learning representations},
  year={2022}
}

@article{zhang2020learning,
  title={Learning to dispatch for job shop scheduling via deep reinforcement learning},
  author={Zhang, Cong and Song, Wen and Cao, Zhiguang and Zhang, Jie and Tan, Puay Siew and Chi, Xu},
  journal={Advances in neural information processing systems},
  volume={33},
  pages={1621--1632},
  year={2020}
}

@article{wei2022chain,
  title={Chain-of-thought prompting elicits reasoning in large language models},
  author={Wei, Jason and Wang, Xuezhi and Schuurmans, Dale and Bosma, Maarten and Xia, Fei and Chi, Ed and Le, Quoc V and Zhou, Denny and others},
  journal={Advances in neural information processing systems},
  volume={35},
  pages={24824--24837},
  year={2022}
}

@inproceedings{wang2022self,
  author       = {Xuezhi Wang and
                  Jason Wei and
                  Dale Schuurmans and
                  Quoc V. Le and
                  Ed H. Chi and
                  Sharan Narang and
                  Aakanksha Chowdhery and
                  Denny Zhou},
  title        = {Self-Consistency Improves Chain of Thought Reasoning in Language Models},
  booktitle    = {The Eleventh International Conference on Learning Representations,
                  {ICLR} 2023, Kigali, Rwanda, May 1-5, 2023},
  publisher    = {OpenReview.net},
  year         = {2023},
  url          = {https://openreview.net/forum?id=1PL1NIMMrw},
  bibsource    = {dblp computer science bibliography, https://dblp.org}
}

@article{xu2025learn,
  title={Learn to optimise for job shop scheduling: a survey with comparison between genetic programming and reinforcement learning},
  author={Xu, Meng and Mei, Yi and Zhang, Fangfang and Zhang, Mengjie},
  journal={Artificial Intelligence Review},
  volume={58},
  number={6},
  pages={1--53},
  year={2025},
  publisher={Springer}
}

@article{ferreira2022effective,
  title={Effective and interpretable dispatching rules for dynamic job shops via guided empirical learning},
  author={Ferreira, Cristiane and Figueira, Gon{\c{c}}alo and Amorim, Pedro},
  journal={Omega},
  volume={111},
  pages={102643},
  year={2022},
  publisher={Elsevier}
}

@article{ngwu2025reinforcement,
  title={Reinforcement learning in dynamic job shop scheduling: a comprehensive review of AI-driven approaches in modern manufacturing},
  author={Ngwu, Chinyere and Liu, Ying and Wu, Rui},
  journal={Journal of Intelligent Manufacturing},
  pages={1--16},
  year={2025},
  publisher={Springer}
}

@article{li2025q,
  title={A Q-learning improved differential evolution algorithm for human-centric dynamic distributed flexible job shop scheduling problem},
  author={Li, Xixing and Guo, Ao and Yin, Xiyan and Tang, Hongtao and Wu, Rui and Zhao, Qingqing and Li, Yibing and Wang, XiVincent},
  journal={Journal of Manufacturing Systems},
  volume={80},
  pages={794--823},
  year={2025},
  publisher={Elsevier}
}

@article{baykasouglu2020greedy,
  title={Greedy randomized adaptive search for dynamic flexible job-shop scheduling},
  author={Baykaso{\u{g}}lu, Adil and Madeno{\u{g}}lu, Fatma S and Hamzaday{\i}, Alper},
  journal={Journal of Manufacturing Systems},
  volume={56},
  pages={425--451},
  year={2020},
  publisher={Elsevier}
}

@article{ren2022joint,
  title={Joint optimisation for dynamic flexible job-shop scheduling problem with transportation time and resource constraints},
  author={Ren, Weibo and Yan, Yan and Hu, Yaoguang and Guan, Yu},
  journal={International Journal of Production Research},
  volume={60},
  number={18},
  pages={5675--5696},
  year={2022},
  publisher={Taylor \& Francis}
}

@article{shahgholi2019heuristic,
  title={A heuristic model for dynamic flexible job shop scheduling problem considering variable processing times},
  author={Shahgholi Zadeh, Melissa and Katebi, Yalda and Doniavi, Ali},
  journal={International Journal of Production Research},
  volume={57},
  number={10},
  pages={3020--3035},
  year={2019},
  publisher={Taylor \& Francis}
}

@inproceedings{yang2024deep,
  title={A deep reinforcement learning based approach for dynamic job shop scheduling considering variable processing time},
  author={Yang, Shuai and Guo, Hongwei and Huang, Jiaqi and Han, Kexian},
  booktitle={Proceedings of the 2024 4th International Conference on Artificial Intelligence, Automation and High Performance Computing},
  pages={368--374},
  year={2024}
}

@article{liu2023deep,
  title={A deep multi-agent reinforcement learning approach to solve dynamic job shop scheduling problem},
  author={Liu, Renke and Piplani, Rajesh and Toro, Carlos},
  journal={Computers \& Operations Research},
  volume={159},
  pages={106294},
  year={2023},
  publisher={Elsevier}
}

@article{luo2021real,
  title={Real-time scheduling for dynamic partial-no-wait multiobjective flexible job shop by deep reinforcement learning},
  author={Luo, Shu and Zhang, Linxuan and Fan, Yushun},
  journal={IEEE Transactions on Automation Science and Engineering},
  volume={19},
  number={4},
  pages={3020--3038},
  year={2021},
  publisher={IEEE}
}

@article{abgaryan2025starjob,
  author       = {Henrik Abgaryan and
                  Tristan Cazenave and
                  Ararat Harutyunyan},
  title        = {Starjob: Dataset for LLM-Driven Job Shop Scheduling},
  journal      = {CoRR},
  volume       = {abs/2503.01877},
  year         = {2025},
  url          = {https://doi.org/10.48550/arXiv.2503.01877},
  doi          = {10.48550/ARXIV.2503.01877},
  eprinttype    = {arXiv},
  eprint       = {2503.01877},
  bibsource    = {dblp computer science bibliography, https://dblp.org}
}

@article{abgaryan2024llms,
  author       = {Henrik Abgaryan and
                  Ararat Harutyunyan and
                  Tristan Cazenave},
  title        = {LLMs can Schedule},
  journal      = {CoRR},
  volume       = {abs/2408.06993},
  year         = {2024},
  url          = {https://doi.org/10.48550/arXiv.2408.06993},
  doi          = {10.48550/ARXIV.2408.06993},
  eprinttype    = {arXiv},
  eprint       = {2408.06993},
  bibsource    = {dblp computer science bibliography, https://dblp.org}
}

@article{amarasinghe2023ai,
  author       = {Pivithuru Thejan Amarasinghe and
                  Su Nguyen and
                  Yuan Sun and
                  Damminda Alahakoon},
  title        = {AI-Copilot for Business Optimisation: {A} Framework and {A} Case Study
                  in Production Scheduling},
  journal      = {CoRR},
  volume       = {abs/2309.13218},
  year         = {2023},
  url          = {https://doi.org/10.48550/arXiv.2309.13218},
  doi          = {10.48550/ARXIV.2309.13218},
  eprinttype    = {arXiv},
  eprint       = {2309.13218},
  bibsource    = {dblp computer science bibliography, https://dblp.org}
}

@article{ahmaditeshnizi2024optimus,
  author       = {Ali AhmadiTeshnizi and
                  Wenzhi Gao and
                  Herman Brunborg and
                  Shayan Talaei and
                  Madeleine Udell},
  title        = {OptiMUS-0.3: Using Large Language Models to Model and Solve Optimization
                  Problems at Scale},
  journal      = {CoRR},
  volume       = {abs/2407.19633},
  year         = {2024},
  url          = {https://doi.org/10.48550/arXiv.2407.19633},
  doi          = {10.48550/ARXIV.2407.19633},
  eprinttype    = {arXiv},
  eprint       = {2407.19633},
  bibsource    = {dblp computer science bibliography, https://dblp.org}
}

@article{huang2024automatic,
  author       = {Jin Huang and
                  Xinyu Li and
                  Liang Gao and
                  Qihao Liu and
                  Yue Teng},
  title        = {Automatic programming via large language models with population self-evolution
                  for dynamic job shop scheduling problem},
  journal      = {CoRR},
  volume       = {abs/2410.22657},
  year         = {2024},
  url          = {https://doi.org/10.48550/arXiv.2410.22657},
  doi          = {10.48550/ARXIV.2410.22657},
  eprinttype    = {arXiv},
  eprint       = {2410.22657},
  bibsource    = {dblp computer science bibliography, https://dblp.org}
}

@book{bertsekas2012dynamic,
  title={Dynamic Programming and Optimal Control: Volume II; Approximate Dynamic Programming},
  author={Bertsekas, Dimitri},
  volume={4},
  year={2012},
  publisher={Athena Scientific}
}

@article{li2025learning,
  author       = {Sirui Li and
                  Wenbin Ouyang and
                  Yining Ma and
                  Cathy Wu},
  title        = {Learning-Guided Rolling Horizon Optimization for Long-Horizon Flexible
                  Job-Shop Scheduling},
  journal      = {CoRR},
  volume       = {abs/2502.15791},
  year         = {2025},
  url          = {https://doi.org/10.48550/arXiv.2502.15791},
  doi          = {10.48550/ARXIV.2502.15791},
  eprinttype    = {arXiv},
  eprint       = {2502.15791},
  bibsource    = {dblp computer science bibliography, https://dblp.org}
}

@article{cao2024novel,
  title={A novel memetic algorithm for energy-efficient distributed heterogeneous flexible job shop scheduling: Case studies in uavs delivery},
  author={Cao, Shijie and Yuan, Yuan},
  journal={IEEE Internet of Things Journal},
  year={2024},
  publisher={IEEE}
}

@article{liu2023lostmiddlelanguagemodels,
  author       = {Nelson F. Liu and
                  Kevin Lin and
                  John Hewitt and
                  Ashwin Paranjape and
                  Michele Bevilacqua and
                  Fabio Petroni and
                  Percy Liang},
  title        = {Lost in the Middle: How Language Models Use Long Contexts},
  journal      = {Trans. Assoc. Comput. Linguistics},
  volume       = {12},
  pages        = {157--173},
  year         = {2024},
  url          = {https://doi.org/10.1162/tacl\_a\_00638},
  doi          = {10.1162/TACL\_A\_00638},
  bibsource    = {dblp computer science bibliography, https://dblp.org}
}

@inproceedings{uzunoglu2024paradiseevaluatingimplicitplanning,
  author       = {Arda Uzunoglu and
                  G{\"{o}}zde G{\"{u}}l Sahin and
                  Abdulfattah Safa},
  editor       = {Lun{-}Wei Ku and
                  Andre Martins and
                  Vivek Srikumar},
  title        = {{PARADISE:} Evaluating Implicit Planning Skills of Language Models
                  with Procedural Warnings and Tips Dataset},
  booktitle    = {Findings of the Association for Computational Linguistics, {ACL} 2024,
                  Bangkok, Thailand and virtual meeting, August 11-16, 2024},
  pages        = {10085--10102},
  publisher    = {Association for Computational Linguistics},
  year         = {2024},
  url          = {https://doi.org/10.18653/v1/2024.findings-acl.599},
  doi          = {10.18653/V1/2024.FINDINGS-ACL.599},
  bibsource    = {dblp computer science bibliography, https://dblp.org}
}

@article{mu2024llmsfollowsimplerules,
  author       = {Norman Mu and
                  Sarah Li Chen and
                  Zifan Wang and
                  Sizhe Chen and
                  David Karamardian and
                  Lulwa Aljeraisy and
                  Dan Hendrycks and
                  David A. Wagner},
  title        = {Can LLMs Follow Simple Rules?},
  journal      = {CoRR},
  volume       = {abs/2311.04235},
  year         = {2023},
  url          = {https://doi.org/10.48550/arXiv.2311.04235},
  doi          = {10.48550/ARXIV.2311.04235},
  eprinttype    = {arXiv},
  eprint       = {2311.04235},
  bibsource    = {dblp computer science bibliography, https://dblp.org}
}

@article{schmied2025llmsgreedyagentseffects,
  author       = {Thomas Schmied and
                  J{\"{o}}rg Bornschein and
                  Jordi Grau{-}Moya and
                  Markus Wulfmeier and
                  Razvan Pascanu},
  title        = {LLMs are Greedy Agents: Effects of {RL} Fine-tuning on Decision-Making
                  Abilities},
  journal      = {CoRR},
  volume       = {abs/2504.16078},
  year         = {2025},
  url          = {https://doi.org/10.48550/arXiv.2504.16078},
  doi          = {10.48550/ARXIV.2504.16078},
  eprinttype    = {arXiv},
  eprint       = {2504.16078},
  bibsource    = {dblp computer science bibliography, https://dblp.org}
}

@article{baeumel2025lookaheadlimitationmultioperandaddition,
  author       = {Tanja Baeumel and
                  Josef van Genabith and
                  Simon Ostermann},
  title        = {The Lookahead Limitation: Why Multi-Operand Addition is Hard for LLMs},
  journal      = {CoRR},
  volume       = {abs/2502.19981},
  year         = {2025},
  url          = {https://doi.org/10.48550/arXiv.2502.19981},
  doi          = {10.48550/ARXIV.2502.19981},
  eprinttype    = {arXiv},
  eprint       = {2502.19981},
  bibsource    = {dblp computer science bibliography, https://dblp.org}
}

@inproceedings{zhang2024restmctsllmselftrainingprocess,
  author       = {Dan Zhang and
                  Sining Zhoubian and
                  Ziniu Hu and
                  Yisong Yue and
                  Yuxiao Dong and
                  Jie Tang},
  editor       = {Amir Globersons and
                  Lester Mackey and
                  Danielle Belgrave and
                  Angela Fan and
                  Ulrich Paquet and
                  Jakub M. Tomczak and
                  Cheng Zhang},
  title        = {ReST-MCTS*: {LLM} Self-Training via Process Reward Guided Tree Search},
  booktitle    = {Advances in Neural Information Processing Systems 38: Annual Conference
                  on Neural Information Processing Systems 2024, NeurIPS 2024, Vancouver,
                  BC, Canada, December 10 - 15, 2024},
  year         = {2024},
  url          = {http://papers.nips.cc/paper\_files/paper/2024/hash/76ec4dc30e9faaf0e4b6093eaa377218-Abstract-Conference.html},
  bibsource    = {dblp computer science bibliography, https://dblp.org}
}

@inproceedings{jiang2024longllmlinguaacceleratingenhancingllms,
  author       = {Huiqiang Jiang and
                  Qianhui Wu and
                  Xufang Luo and
                  Dongsheng Li and
                  Chin{-}Yew Lin and
                  Yuqing Yang and
                  Lili Qiu},
  editor       = {Lun{-}Wei Ku and
                  Andre Martins and
                  Vivek Srikumar},
  title        = {LongLLMLingua: Accelerating and Enhancing LLMs in Long Context Scenarios
                  via Prompt Compression},
  booktitle    = {Proceedings of the 62nd Annual Meeting of the Association for Computational
                  Linguistics (Volume 1: Long Papers), {ACL} 2024, Bangkok, Thailand,
                  August 11-16, 2024},
  pages        = {1658--1677},
  publisher    = {Association for Computational Linguistics},
  year         = {2024},
  url          = {https://doi.org/10.18653/v1/2024.acl-long.91},
  doi          = {10.18653/V1/2024.ACL-LONG.91},
  bibsource    = {dblp computer science bibliography, https://dblp.org}
}

@inproceedings{gui2025hypertreeplanningenhancingllm,
  author       = {Runquan Gui and
                  Zhihai Wang and
                  Jie Wang and
                  Chi Ma and
                  Huiling Zhen and
                  Mingxuan Yuan and
                  Jianye Hao and
                  Defu Lian and
                  Enhong Chen and
                  Feng Wu},
  title        = {HyperTree Planning: Enhancing {LLM} Reasoning via Hierarchical Thinking},
  booktitle    = {Forty-second International Conference on Machine Learning, {ICML}
                  2025, Vancouver, BC, Canada, July 13-19, 2025},
  publisher    = {OpenReview.net},
  year         = {2025},
  url          = {https://openreview.net/forum?id=45he3Ri6JP},
  bibsource    = {dblp computer science bibliography, https://dblp.org}
}

@inproceedings{light2025strategistselfimprovementllmdecision,
  author       = {Jonathan Light and
                  Min Cai and
                  Weiqin Chen and
                  Guanzhi Wang and
                  Xiusi Chen and
                  Wei Cheng and
                  Yisong Yue and
                  Ziniu Hu},
  title        = {Strategist: Self-improvement of {LLM} Decision Making via Bi-Level
                  Tree Search},
  booktitle    = {The Thirteenth International Conference on Learning Representations,
                  {ICLR} 2025, Singapore, April 24-28, 2025},
  publisher    = {OpenReview.net},
  year         = {2025},
  url          = {https://openreview.net/forum?id=gfI9v7AbFg},
  bibsource    = {dblp computer science bibliography, https://dblp.org}
}

@conference{Gordon1995StableFunction,
author = {Geoffrey Gordon},
title = {Stable Function Approximation in Dynamic Programming},
booktitle = {Proceedings of (ICML) International Conference on Machine Learning},
year = {1995},
month = {July},
pages = {261 - 268},
}

@article{Janez2006Statistical,
  author  = {Janez Dem{\v{s}}ar},
  title   = {Statistical Comparisons of Classifiers over Multiple Data Sets},
  journal = {Journal of Machine Learning Research},
  year    = {2006},
  volume  = {7},
  number  = {1},
  pages   = {1--30},
  url     = {http://jmlr.org/papers/v7/demsar06a.html}
}

@article{openai2024gpt4ocard,
  title={Gpt-4o system card},
  author={Hurst, Aaron and Lerer, Adam and Goucher, Adam P and Perelman, Adam and Ramesh, Aditya and Clark, Aidan and Ostrow, AJ and Welihinda, Akila and Hayes, Alan and Radford, Alec and others},
  journal={arXiv preprint arXiv:2410.21276},
  year={2024}
}

@article{deepseekai2025deepseekv3technicalreport,
  title={Deepseek-v3 technical report},
  author={Liu, Aixin and Feng, Bei and Xue, Bing and Wang, Bingxuan and Wu, Bochao and Lu, Chengda and Zhao, Chenggang and Deng, Chengqi and Zhang, Chenyu and Ruan, Chong and others},
  journal={arXiv preprint arXiv:2412.19437},
  year={2024}
}

\end{document}